\documentclass[10pt,journal,cspaper,compsoc]{IEEEtran}
\usepackage{graphicx} 
\usepackage[tight,sf,SF]{subfigure}

\usepackage{algorithm}
\usepackage{algorithmic}

\usepackage{units}
\usepackage{url}
\usepackage{graphicx}
\usepackage{mathtools}
\usepackage{dsfont}
\usepackage{amsmath, amssymb}
\usepackage{wrapfig}

\usepackage{hyperref}

\hypersetup{
    bookmarks=true,         
    unicode=false,          
    pdffitwindow=false,     
    pdfstartview={FitH},    
    pdftitle={Gaussian Processes for Data-Efficient Learning in
      Robotics and Control},    
     pdfauthor={Marc Deisenroth},     
    colorlinks=true,       
    linkcolor=black,          
    citecolor=black,        
    filecolor=black,      
    urlcolor=blue          
}

\usepackage[all]{hypcap}

\usepackage{color}

\newcommand\cut[1]{}

\newcommand{\twofig}{0.46\hsize}
\newcommand{\threefig}{0.31\hsize}
\newcommand{\fourfig}{0.23\hsize}
\newcommand{\sixfig}{0.16\hsize}
\newcommand{\R}{\mathds{R}}

\renewcommand{\vec}{\boldsymbol}
\newcommand{\mat}{\boldsymbol}
\newcommand{\E}{\mathds{E}}
\newcommand{\var}{\mathrm{var}}
\newcommand{\cov}{\mathrm{cov}}
\newcommand{\diag}{\mathrm{diag}}
\newcommand{\tr}{\mathrm{tr}}
\newcommand{\cost}{c}
\newcommand{\T}{^\top}
\newcommand{\inv}{^{-1}}
\newcommand{\prob}{{p}}
\renewcommand{\d}{\mathrm d}
\newcommand{\gauss}[2]{\mathcal N(#1,#2)}
\newcommand{\gaussx}[3]{\mathcal{N}\big(#1\,|\,#2,#3\big)}
\renewcommand{\d}{\operatorname{d}\!}
\newcommand{\polpar}{\theta}
\renewcommand{\sec}{Sec.}
\newcommand{\secs}{Secs.}

\newcommand{\eq}{}
\newcommand{\eqs}{}

\newcommand{\fig}{Fig.}
\newcommand{\Fig}{Fig.}
\newcommand{\alg}{Alg.}
\newcommand{\Alg}{Alg.}

\newcommand{\vaugx}{\tilde{\vec x}}

\newcommand{\idx}[1]{{(#1)}}
\newcommand{\squash}[0]{\sigma}
\newcommand{\figpath}[0]{./figures/}

\definecolor{orange}{rgb}{1,0.5,0}
\definecolor{cyan}{rgb}{0.6, 0.2, 0.6}
\definecolor{magenta}{rgb}{1,0,1}
\definecolor{green2}{rgb}{0.2, 0.5, 0.2}

\newcommand{\red}[1]{\textcolor{red}{#1}}
\newcommand{\blue}[1]{\textcolor{blue}{#1}}
\newcommand{\green}[1]{\textcolor{green2}{#1}}
\newcommand{\cyan}[1]{\textcolor{cyan}{#1}}
\newcommand{\new}[1]{{#1}}

\newcommand{\figspace}{\vspace{0mm}}

\usepackage{ifpdf}

\ifCLASSOPTIONcompsoc
  \usepackage[nocompress]{cite}
\else
  \usepackage{cite}
\fi

\ifCLASSINFOpdf
\else
\fi
\usepackage{array}

\usepackage{mdwmath}
\usepackage{mdwtab}

\usepackage{stfloats}

\begin{document}
%
\title{Gaussian Processes for Data-Efficient Learning in Robotics and Control}

\author{Marc Peter Deisenroth, Dieter Fox, and Carl Edward
  Rasmussen
  \IEEEcompsocitemizethanks{\IEEEcompsocthanksitem M.P.~Deisenroth is
    with the Department of Computing, Imperial College London, 180
    Queen's Gate, London SW7 2AZ, United Kingdom, and
    with the Department of Computer Science, TU Darmstadt, Germany.\protect
\IEEEcompsocthanksitem D.~Fox is with the Department of Computer Science
\& Engineering, University of Washington, Box 352350, Seattle, WA
98195-2350.
\IEEEcompsocthanksitem C.E.~Rasmussen is with the Department of
Engineering, University of Cambridge, Trumpington Street, Cambridge
CB2 1PZ, United Kingdom.}
\thanks{Manuscript received 15 Sept. 2012; revised 6 May 2013;
  accepted 20 Oct. 2013; published online 4 November 2013.\newline
Recommended for acceptance by R.P. Adams, E. Fox, E. Sudderth, and
Y.W.~Teh.\newline
For information on obtaining reprints of this article, please send
e-mail to: tpami@computer.org and reference IEEECS Log
Number\newline
TPAMISI-2012-09-0742.\newline
Digital Object Identifier no. 10.1109/TPAMI.2013.218}}

\markboth{IEEE TRANSACTIONS ON PATTERN ANALYSIS AND MACHINE
  INTELLIGENCE, \quad VOL. 37, \quad NO. 2, \quad FEBRUARY 2015}
{DEISENROTH \MakeLowercase{\textit{et al.}}: GAUSSIAN PROCESSES FOR DATA-EFFICIENT LEARNING IN
  ROBOTICS AND CONTROL}


\IEEEcompsoctitleabstractindextext{%
\begin{abstract}
  Autonomous learning has been a promising direction in control and
  robotics for more than a decade since data-driven learning allows to
  reduce the amount of engineering knowledge, which is otherwise
  required.
  However, autonomous reinforcement learning (RL) approaches typically
  require many interactions with the system to learn controllers,
  which is a practical limitation in real systems, such as robots,
  where many interactions can be impractical and time consuming. To
  address this problem, current learning approaches typically require
  task-specific knowledge in form of expert demonstrations, realistic
  simulators, pre-shaped policies, or specific knowledge about the
  underlying dynamics.
  In this article, we follow a different approach and speed up
  learning by extracting more information from data. In particular, we
  learn a probabilistic, non-parametric Gaussian process transition
  model of the system.
  By explicitly incorporating model uncertainty into long-term
  planning and controller learning our approach reduces the effects
  of model errors, a key problem in model-based learning.
  Compared to state-of-the art RL our model-based policy search method
  achieves an unprecedented speed of learning. We demonstrate its
  applicability to autonomous learning in real robot and control
  tasks. 


\end{abstract}

}

\maketitle

\IEEEdisplaynotcompsoctitleabstractindextext

\IEEEpeerreviewmaketitle

\section{Introduction}
\IEEEPARstart{O}{ne} of the main limitations of many current
reinforcement learning (RL) algorithms is that learning is
prohibitively slow, i.e., the required number of interactions with the
environment is impractically high. For example, many RL approaches in
problems with low-dimensional state spaces and fairly benign dynamics
require thousands of trials to learn. This \emph{data inefficiency}
makes learning \new{in real control\slash robotic systems} impractical
and prohibits RL approaches in more challenging scenarios.

Increasing the data efficiency \new{in RL} requires either
task-specific prior knowledge or extraction of more information from
available data. In this article, we assume that expert knowledge
(e.g., in terms of expert demonstrations~\cite{Schaal1997}, realistic
simulators, or explicit differential equations for the dynamics) is
unavaiable. Instead, we carefully model the observed dynamics using a
general flexible nonparametric approach.

\begin{figure*}[tb]
\centering
\subfigure{
  \includegraphics[width=\threefig]{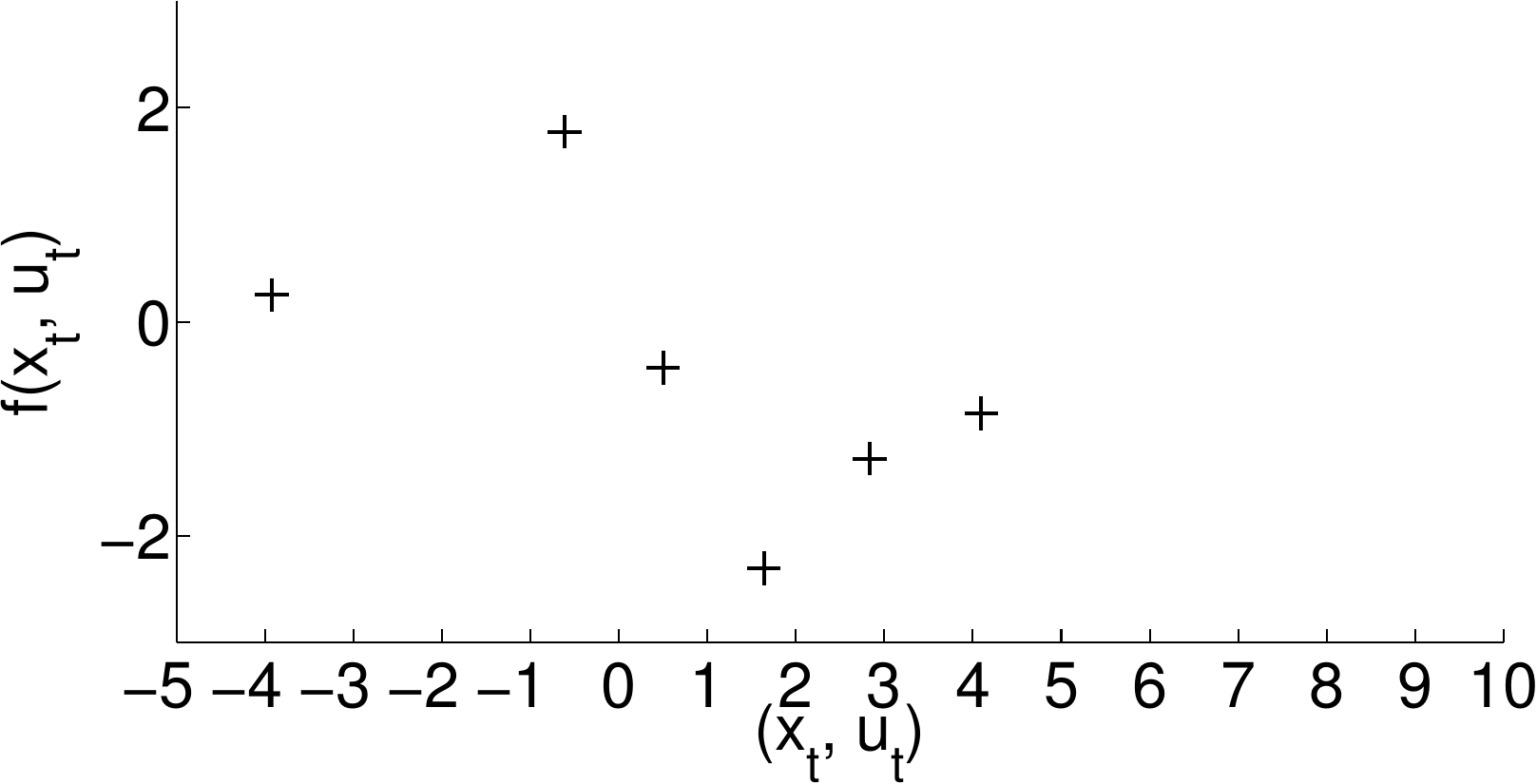}
}
\hfill
\subfigure{
  \includegraphics[width=\threefig]{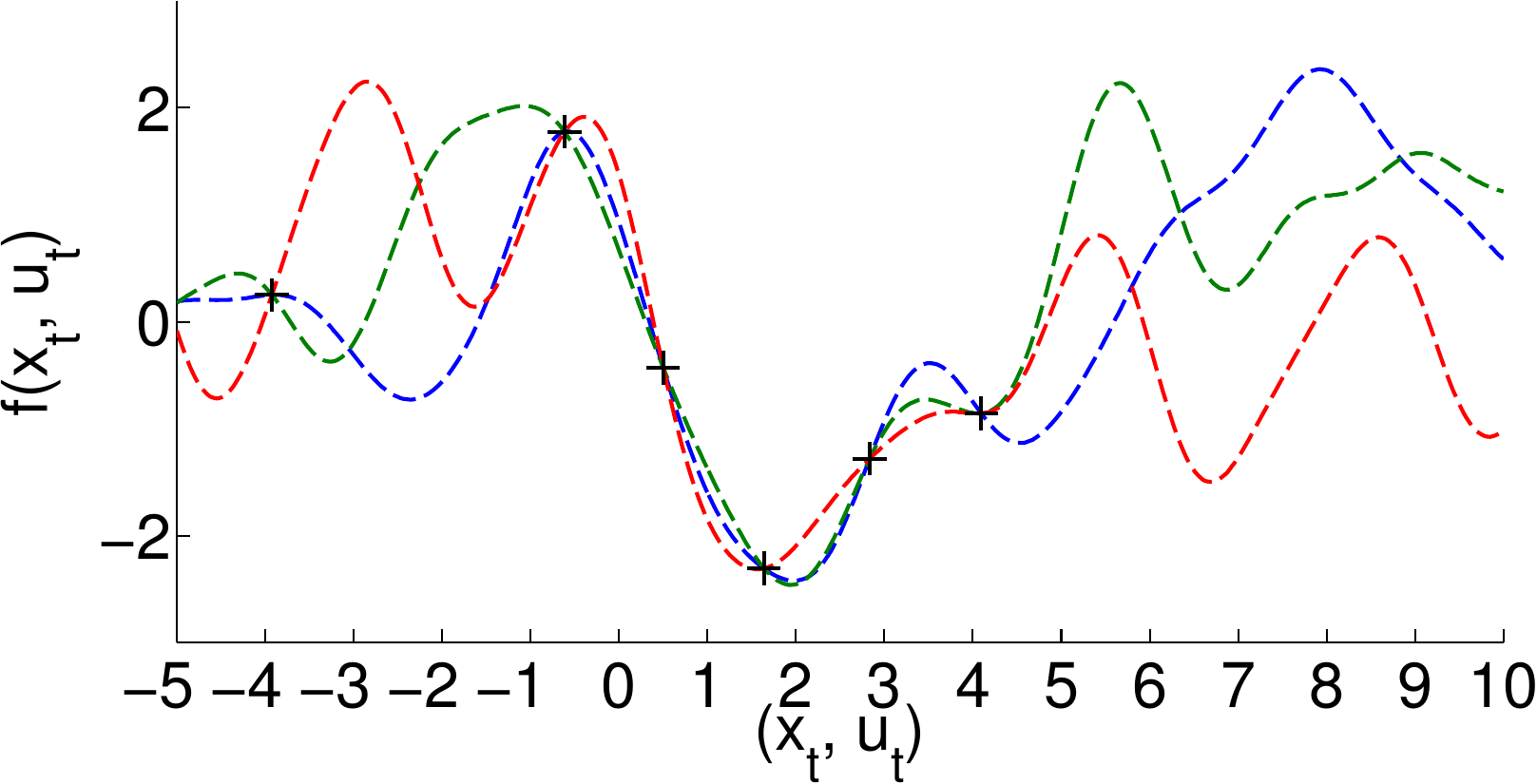}
}
\hfill
\subfigure{
  \includegraphics[width=\threefig]{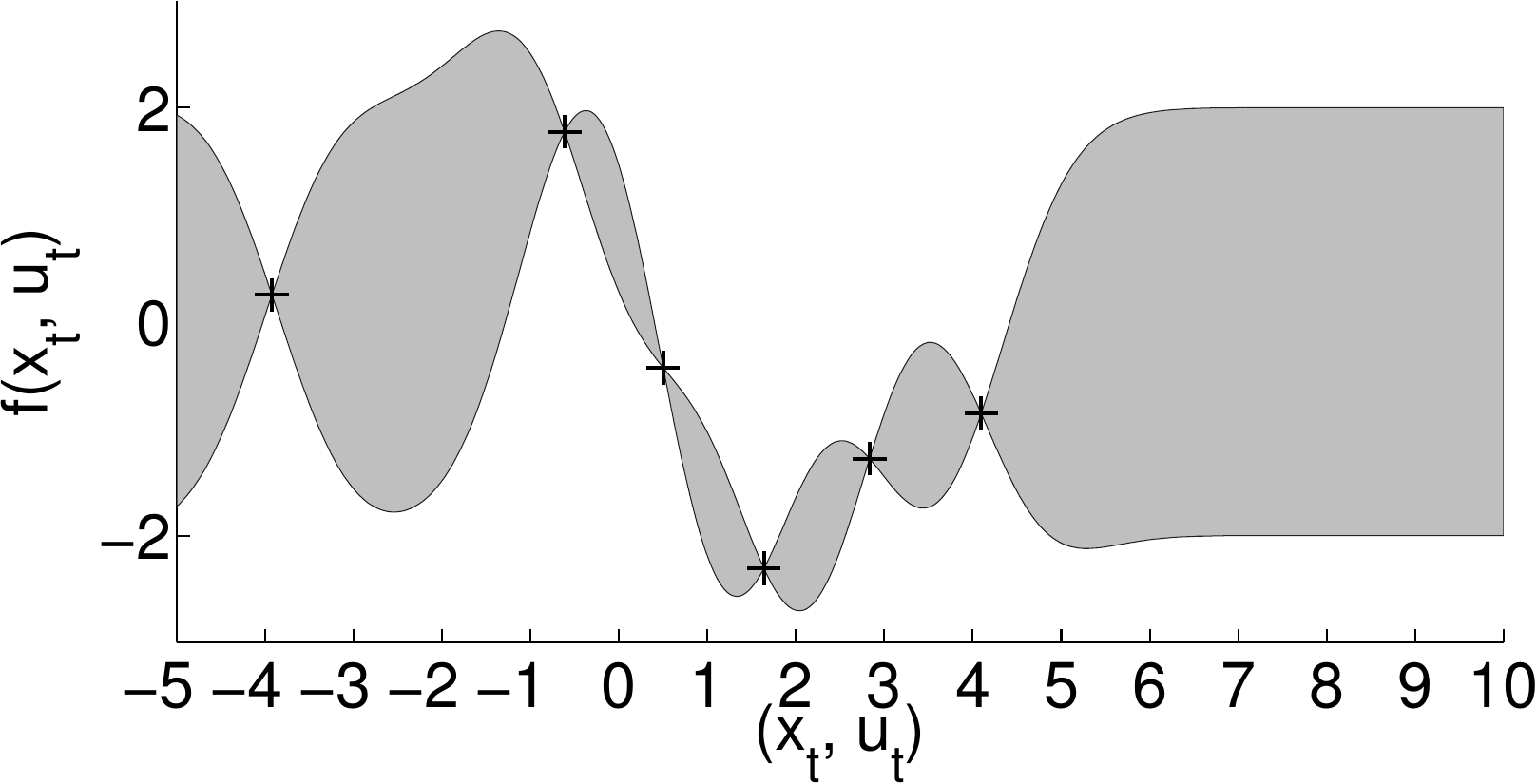}
}
\vspace{-0.2mm}
\caption{Effect of model errors. Left: Small data set of observed
  transitions from an idealized one-dimensional representations of
  states and actions $(x_t,u_t)$ to the next state
  $x_{t+1}=f(x_t,u_t)$. Center: Multiple plausible deterministic
  models. Right: Probabilistic model. The probabilistic model
  describes the uncertainty about the latent function by a probability
  distribution on the set of all plausible transition
  functions. Predictions with deterministic models are claimed with
  full confidence, while the probabilistic model expresses its
  predictive uncertainty by a probability distribution.}
\label{fig:model bias}
\figspace
\end{figure*}

Generally, model-based methods, i.e., methods which learn an explicit
dynamics model of the environment, are more promising to efficiently
extract valuable information from available data~\cite{Atkeson1997a}
than model-free methods, such as Q-learning~\cite{Watkins1989} or
TD-learning~\cite{Sutton1998}. The main reason why model-based methods
are not widely used in RL is that they can suffer severely from
\emph{model errors}, i.e., they inherently assume that the learned
model resembles the real environment sufficiently
accurately~\cite{Schneider1997,Schaal1997,Atkeson1997a}. Model errors
are especially an issue when only a few samples and no informative
prior knowledge about the task are available.
%
\Fig~\ref{fig:model bias} illustrates how model errors can affect
learning.
Given a small data set of observed transitions (left), multiple
transition functions plausibly could have generated them
(center). Choosing a single deterministic model has severe
consequences: Long-term predictions often leave the range of the
training data in which case the predictions become essentially
arbitrary. However, the deterministic model claims them with full
confidence! By contrast, a probabilistic model places a posterior
distribution on plausible transition functions (right) and expresses
the level of uncertainty about the model itself.

When learning models, considerable model uncertainty is present,
especially early on in learning. Thus, we require \emph{probabilistic}
models to express this uncertainty. Moreover, model uncertainty needs
to be incorporated into planning and policy evaluation. Based on these
ideas, we propose \textsc{pilco} (Probabilistic Inference for Learning
Control), a model-based policy search
method~\cite{Deisenroth2011c,Deisenroth2011b}. As a probabilistic
model we use nonparametric Gaussian processes
(GPs)~\cite{Rasmussen2006}.  \textsc{Pilco} uses computationally
efficient deterministic approximate inference for long-term
predictions and policy evaluation. Policy improvement is based on
\emph{analytic} policy gradients. Due to probabilistic modeling and
inference \textsc{pilco} achieves unprecedented learning efficiency in
continuous state-action domains and, hence, is directly applicable to
complex mechanical systems, such as robots.

In this article, we provide a detailed overview of the key ingredients
of the \textsc{pilco} learning framework. In particular, we assess the
quality of two different approximate inference methods in the context
of policy search. Moreover, we give a concrete example of the
importance of Bayesian modeling and inference for fast learning from
scratch. We demonstrate that \textsc{Pilco}'s unprecedented learning
speed makes it directly applicable to realistic control and robotic
hardware platforms.

This article is organized as follows: After discussing related work in
\sec~\ref{sec:relwork}, we describe the key ideas of the
\textsc{pilco} learning framework in \sec~\ref{sec:pilco}, i.e., the
dynamics model, policy evaluation, and gradient-based policy
improvement. In \sec~\ref{sec:approximate inference}, we detail two
approaches for long-term predictions for policy evaluation. In
\sec~\ref{sec:policy}, we describe how the policy is represented and
practically implemented. A particular cost function and its natural
exploration\slash exploitation trade-off are discussed in
\sec~\ref{sec:cost function}. Experimental results are provided in
\sec~\ref{sec:results}. In \sec~\ref{sec:discussion}, we discuss key
properties, limitations, and extensions of the \textsc{pilco}
framework before concluding in \sec~\ref{sec:conclusion}.


\section{Related Work}
\label{sec:relwork}
Controlling systems under parameter uncertainty has been investigated
for decades in robust and adaptive
control~\cite{McFarlane1989,Astrom2008}.
Typically, a certainty equivalence principle is applied, which treats
estimates of the model parameters as if they were the true
values~\cite{Wittenmark1995}.
Approaches to designing adaptive controllers that explicitly take
uncertainty about the model parameters into account are stochastic
adaptive control~\cite{Astrom2008} and dual
control~\cite{Feldbaum1961}. Dual control aims to reduce parameter
uncertainty by explicit probing, which is closely related to the
exploration problem in RL. Robust, adaptive, and dual control are most
often applied to linear systems~\cite{Wittenmark1995}; nonlinear
extensions exist in special cases~\cite{Fabri1998}. 

The specification of parametric models for a particular control
problem is often challenging and requires intricate knowledge about
the system. Sometimes, a rough model estimate with uncertain
parameters is sufficient to solve challenging control problems. For
instance, in~\cite{Abbeel2006}, this approach was applied together
with locally optimal controllers and temporal bias terms for handling
model errors.  The key idea was to ground policy evaluations using
real-life trials, but not the approximate model.
%

All above-mentioned approaches to finding controllers require more or
less accurate \emph{parametric} models. These models are problem
specific and have to be manually specified, i.e., they are not suited
for learning models for a broad range of tasks. Nonparametric
regression methods, however, are promising to automatically extract
the important features of the latent dynamics from
data. In~\cite{Schneider1997, Bagnell2001} locally weighted Bayesian
regression was used as a nonparametric method for learning these
models. To deal with model uncertainty, in~\cite{Bagnell2001} model
parameters were sampled from the parameter posterior, which accounts
for temporal correlation. In~\cite{Schneider1997}, model uncertainty
was treated as noise. The approach to controller learning was based on
stochastic dynamic programming in discretized spaces, where the model
errors at each time step were assumed independent.


\textsc{Pilco} builds upon the idea of treating model uncertainty as
noise~\cite{Schneider1997}. However, unlike~\cite{Schneider1997},
\textsc{pilco} is a policy search method and does not require state
space discretization.  Instead closed-form Bayesian averaging over
infinitely many plausible dynamics models is possible by using
nonparametric GPs.

Nonparametric GP dynamics models in RL were previously proposed
in~\cite{Rasmussen2004,Ko2007,Deisenroth2009}, where the GP training
data were obtained from ``motor babbling''. Unlike \textsc{pilco},
these approaches model global value functions to derive policies,
requiring accurate value function models. To reduce the effect of
model errors in the value functions, many data points are necessary as
value functions are often discontinuous, rendering value-function
based methods in high-dimensional state spaces often statistically and
computationally impractical. Therefore,~\cite{Engel2003,Rasmussen2004,
  Wilson2010, Deisenroth2009} propose to learn GP value function
models to address the issue of model errors in the value
function. However, these methods can usually only be applied to
low-dimensional RL problems.
As a policy search method, \textsc{pilco} does not require an explicit
global value function model but rather searches directly in policy
space. However, unlike value-function based methods, \textsc{pilco} is
currently limited to episodic set-ups.

\section{Model-based Policy Search}
\label{sec:pilco}
In this article, we consider dynamical systems 
\begin{align}
  \vec x_{t+1} = f(\vec x_{t},\vec u_{t})+\vec w\,,\quad \vec
  w\sim\gauss{\vec 0}{\mat\Sigma_w}\,,
\label{eq:system equation}
\end{align}
with continuous-valued states $\vec x\in\R^D$ and controls $\vec
u\in\R^F$, i.i.d. Gaussian system noise $\vec w$, and unknown
transition dynamics $f$. The policy search objective is to find a
\emph{policy\slash controller} $\pi:\vec x\mapsto \pi(\vec
x,\vec\theta)=\vec u$, which minimizes the \emph{expected long-term
  cost}
\begin{align}\label{eq:expected return}
  J^\pi(\vec\polpar) = \sum\nolimits_{t = 0}^T\E_{\vec x_t}[\cost(\vec
  x_t)]\,,\quad \vec x_0\sim\gauss{\vec\mu_0}{\mat\Sigma_0}\,,
\end{align}
of following $\pi$ for $T$ steps, where $\cost(\vec x_t)$ is the cost
of being in state $\vec x$ at time $t$. We assume that $\pi$ is a
function parametrized by $\vec\polpar$.\footnote{\new{
    In our experiments in \sec~\ref{sec:results}, we use a)
    nonlinear parametrizations by means of RBF networks, where the
    parameters $\vec\polpar$ are the weights and the features, or b)
    linear-affine parametrizations, where the parameters $\vec\theta$
    are the weight matrix and a bias term.}}

To find a policy $\pi^*$, which minimizes \eq\eqref{eq:expected
  return}, \textsc{pilco} builds upon three components: 1) a
probabilistic GP dynamics model (\sec~\ref{sec:gp}), 2) deterministic
approximate inference for long-term predictions and policy evaluation
(\sec~\ref{sec:policy evaluation}), 3) analytic computation of the
policy gradients $\d J^\pi(\vec\theta)/\d\vec\theta$ for policy
improvement (\sec~\ref{sec:controller learning}). The GP model
internally represents the dynamics in \eq\eqref{eq:system equation}
and is subsequently employed for long-term predictions $p(\vec
x_1|\pi), \dotsc,p(\vec x_T|\pi)$, given a policy $\pi$. These
predictions are obtained through approximate inference and used to
evaluate the expected long-term cost $J^\pi(\vec\theta)$ in
\eq\eqref{eq:expected return}. The policy $\pi$ is improved based on
gradient information $\d
J^\pi(\vec\theta)/\d\vec\theta$. \Alg~\ref{alg:pilco} summarizes the
\textsc{pilco} learning framework.
%
\begin{algorithm}[tb]
   \caption{\textsc{pilco}}
   \label{alg:pilco}
\begin{algorithmic}[1]
   \STATE {\bfseries init:} Sample controller parameters
   $\vec\polpar\sim\gauss{\vec 0}{\mat I}$. Apply random control
   signals and record data. 
   \REPEAT
   \STATE Learn probabilistic (GP) dynamics model, see
  \sec~\ref{sec:gp}, using all data
  \REPEAT 
  \STATE  Approximate inference for policy evaluation, see
  \sec~\ref{sec:policy evaluation}: get $J^\pi(\vec\polpar)$,
  Eq.~\eqs(\ref{eq:mu_t})--(\ref{eq:E(c)})
  \label{enum:approx inf}
  \STATE Gradient-based policy improvement, see \sec~\ref{sec:controller
    learning}: get $ \d J^\pi(\vec\polpar)/\d\vec\polpar$, Eq.~\eqs(\ref{eq:chain
    rule})--(\ref{eq:last eq in gradient chain})
  \STATE Update parameters $\vec\polpar$ (e.g., CG or L-BFGS).
  \UNTIL{convergence; {\bf return} $\vec\polpar^*$}
  \STATE Set $\pi^*\leftarrow \pi(\vec\polpar^*)$
  \STATE Apply $\pi^*$ to system and record
  data
\label{alg:pilco:apply}
   \UNTIL{task learned}
\end{algorithmic}
\end{algorithm}

\subsection{Model Learning}
\label{sec:gp}
\textsc{Pilco}'s probabilistic dynamics model is implemented as a GP,
where we use tuples $(\vec x_{t}, \vec u_{t})\in\R^{D+F}$ as training
inputs and differences $\vec\Delta_t = \vec x_{t+1} - \vec
x_{t}\in\R^D$ as training targets.\footnote{\new{Using differences as
    training targets encodes an implicit prior mean function $m(\vec
    x) = \vec x$. This means that when leaving the training data, the
    GP predictions do not fall back to 0 but they remain constant.}}  A GP is completely specified by a
mean function $m(\,\cdot\,)$ and a positive semidefinite covariance
function/kernel $k(\,\cdot\,,\,\cdot\,)$.  In this paper, we consider
a prior mean function $m\equiv 0$ and the covariance function
\begin{align}
k(\tilde{\vec x}_p,\tilde{\vec x}_q) &\!=\!
  \sigma_f^2\exp\big(\!-\!\tfrac{1}{2}(\tilde{\vec x}_p\! -\! \tilde{\vec
    x}_q)\T\mat\Lambda\inv(\tilde{\vec x}_p\!-\!\tilde{\vec
    x}_q)\big) \!+\! \delta_{pq}\sigma_w^2\label{eq:SE kernel}
\end{align}
with $\tilde{\vec x}\coloneqq [\vec x\T \vec u\T]\T$. We defined
\mbox{$\mat\Lambda\coloneqq\diag([\ell_1^2,\dotsc,\ell_{D+F}^2])$} in
\eq\eqref{eq:SE kernel}, which depends on the characteristic
length-scales $\ell_i$, and $\sigma_f^2$ is the variance of the latent
transition function $f$. Given $n$ training inputs $\tilde{\mat
  X}=[\tilde{\vec x}_1, \dotsc, \tilde{\vec x}_n]$ and corresponding
training targets $\vec y=[\Delta_1, \dots , \Delta_n]\T$, the
posterior GP hyper-parameters (length-scales $\ell_i$, signal variance
$\sigma_f^2$, and noise variance $\sigma_w^2$) are learned by evidence
maximization~\cite{MacKay2003, Rasmussen2006}.

The posterior GP is a one-step prediction model, and the predicted
successor state $\vec x_{t+1}$ is Gaussian distributed
\begin{align}
  &\prob(\vec x_{t+1}|\vec x_{t},\vec u_{t}) = \gaussx{\vec
    x_{t+1}}{\vec\mu_{t+1}}{\mat\Sigma_{t+1}}\label{eq:one-step prediction distr}\\
  &\vec\mu_{t+1} = \vec x_{t} + \E_f[\vec\Delta_t]\,,\quad 
  \mat\Sigma_{t+1} = \var_f[\vec\Delta_t] \,,
\label{eq:one-step prediction mean and cov}
\end{align}
where the mean and variance of the GP prediction are
\begin{align}
\E_f[\vec\Delta_t]  &= m_f(\tilde{\vec x}_{t})=\vec k_*\T(\mat K +
\sigma_w^2\mat I)^{-1}\vec
y= \vec k_*\T\vec\beta\label{predicted mean}\,,\\
\var_f[\vec\Delta_t] &= k_{**} -\vec k_*\T(\mat K +
\sigma_w^2\mat I)^{-1} \vec k_*\,,
\label{predicted variance}
\end{align}
respectively, with $\vec k_*\coloneqq k(\tilde{\mat X},\tilde{\vec
  x}_{t})$, $k_{**}\coloneqq k(\tilde{\vec x}_{t},\tilde{\vec
  x}_{t})$, and \mbox{$\vec\beta\coloneqq(\mat K + \sigma_w^2\mat
  I)^{-1}\vec y$}, where $\mat K$ is the kernel matrix with entries
$K_{ij}=k(\tilde{\vec x}_i, \tilde{\vec x}_j)$.

For multivariate targets, we train conditionally independent GPs for
each target dimension, i.e., the GPs are independent for given test
inputs. For uncertain inputs, the target dimensions
covary~\cite{Quinonero-Candela2003a}, see also
\sec~\ref{sec:approximate inference}.

\subsection{Policy Evaluation}
\label{sec:policy evaluation}
To evaluate and minimize $J^\pi$ in \eq(\ref{eq:expected return})
\textsc{pilco} uses long-term predictions of the state evolution. In
particular, we determine the marginal $t$-step-ahead predictive
distributions $\prob(\vec x_1|\pi),\dotsc,\prob(\vec x_T|\pi)$ from
the initial state distribution $\prob(\vec x_0)$, $t=1,\dotsc,T$. To
obtain these long-term predictions, we cascade one-step predictions,
see \eqs(\ref{eq:one-step prediction distr})--(\ref{eq:one-step
  prediction mean and cov}), which requires mapping uncertain test
inputs through the GP dynamics model. In the following, we assume that
these test inputs are Gaussian distributed. For notational
convenience, we omit the explicit conditioning on the policy $\pi$ in
the following and assume that episodes start from $\vec
x_0\sim\prob(\vec x_0)=\gaussx{\vec x_0}{\vec\mu_0}{\mat\Sigma_0}$.

For predicting $\vec x_{t+1}$ from $\prob(\vec x_{t})$, we require a
joint distribution $\prob(\tilde{\vec x}_{t}) = \prob(\vec x_{t},\vec
u_{t})$, see \eq\eqref{eq:system equation}. The control $\vec u_{t} =
\pi(\vec x_{t},\vec\polpar)$ is a function of the state, and we
approximate the desired joint distribution $p(\tilde{\vec x}_{t}) =
\prob(\vec x_{t}, \vec u_{t})$ by a Gaussian. Details are provided in
\sec~\ref{sec:succState}.

\begin{figure}[tb]
  \centering 
  \includegraphics[width = 0.8\hsize]{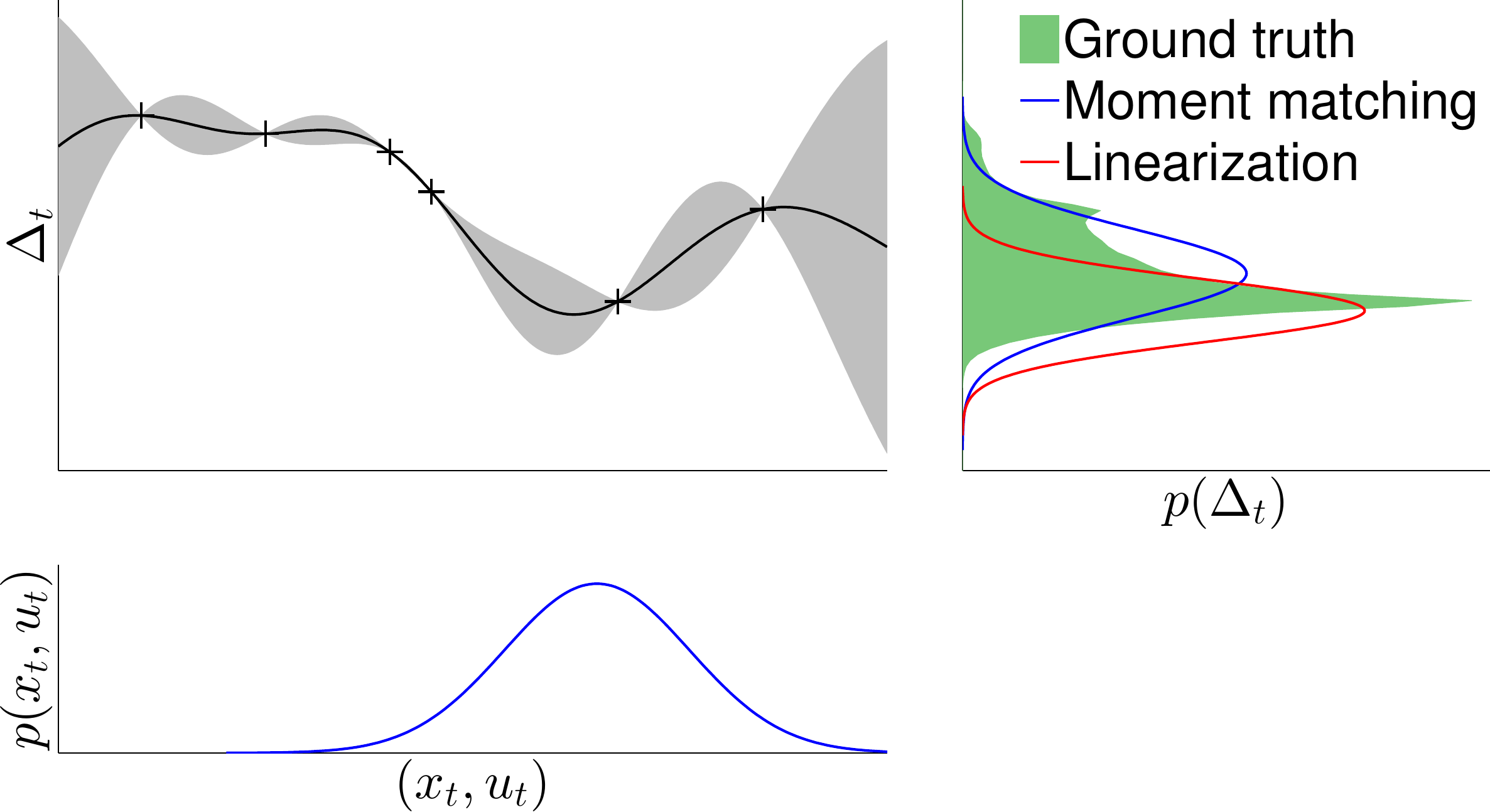}
\caption{GP prediction at an uncertain input. The input distribution
  $\prob(\vec x_{t},\vec u_{t})$ is assumed Gaussian (lower left
  panel). When propagating it through the GP model (upper left panel),
  we obtain the shaded distribution $\prob(\vec\Delta_t)$, upper right
  panel. We approximate $\prob(\vec\Delta_t)$ by a Gaussian (upper
  right panel), which is computed by means of either moment matching
  (blue) or linearization of the posterior GP mean (red).  Using
  linearization for approximate inference can lead to predictive
  distributions that are too tight.}
\label{fig:EKF vs ADF}
\figspace
\end{figure}

From now on, we assume a joint Gaussian distribution distribution 
$\prob(\tilde{\vec x}_{t}) = \gaussx{\tilde{\vec
    x}_{t}}{\tilde{\vec\mu}_{t}}{\tilde{\mat\Sigma}_{t}}$ at
time $t$. To compute
\begin{align}
  \prob(\vec\Delta_t) = \iint \prob(f(\tilde{\vec x}_{t})|\tilde{\vec
    x}_{t}) \prob(\tilde{\vec x}_{t})\d f\d\tilde{\vec x}_{t}\,,
\label{pred. with uncert. inputs}
\end{align}
we integrate out both the random variable $\tilde{\vec x}_{t}$ and
the random function $f$, the latter one according to the posterior GP
distribution.
Computing the exact predictive distribution in \eq(\ref{pred. with
  uncert. inputs}) is analytically intractable as illustrated in
\fig~\ref{fig:EKF vs ADF}. Hence, we approximate $\prob(\vec\Delta_t)$
by a Gaussian.
%

Assume the mean $\vec\mu_{\vec\Delta}$ and the covariance
$\mat\Sigma_{\vec\Delta}$ of the predictive distribution
$\prob(\vec\Delta_t)$ are known\footnote{We will detail their computations in
\secs~\ref{sec:moment matching}--\ref{sec:linearization}.}. Then, a
Gaussian approximation to the desired predictive distribution
$\prob(\vec x_{t+1})$ is given as $\gaussx{\vec
  x_{t+1}}{\vec\mu_{t+1}}{\mat\Sigma_{t+1}}$ with
\begin{align}
  &\vec\mu_{t+1} = \vec\mu_{t} + \vec\mu_{\vec\Delta}\,,\label{eq:mu_t}\\
  &\mat\Sigma_{t+1} = \mat\Sigma_{t} + \mat\Sigma_{\vec\Delta}
  + \cov[\vec x_{t},
  \vec\Delta_t] + \cov[\vec\Delta_t,\vec x_{t}]\,.\label{eq:Sigma_t}
\end{align}
%
Note that both $\vec\mu_{\vec\Delta}$ and $\mat\Sigma_{\vec\Delta}$
are functions of the mean $\vec\mu_u$ and the covariance
$\mat\Sigma_u$ of the control signal.


To evaluate the expected long-term cost $J^\pi$ in \eq(\ref{eq:expected
  return}), it remains to compute the expected values
\begin{align}
  \E_{\vec x_t}[c(\vec x_t)] = \int c(\vec x_t)\gaussx{\vec
    x_t}{\vec\mu_t}{\mat\Sigma_t}\d\vec x_t\,,
  \label{eq:E(c)}
\end{align}
$t = 1,\dotsc,T$, of the cost $c$ with respect to the predictive state
distributions. We choose the cost $c$ such that the integral in
\eq(\ref{eq:E(c)}) and, thus, $J^\pi$ in \eq\eqref{eq:expected return}
can computed analytically. Examples of such cost functions include
polynomials and mixtures of Gaussians.

\subsection{Analytic Gradients for Policy Improvement}
\label{sec:controller learning}
To find policy parameters $\vec\polpar$, which minimize
$J^\pi(\vec\polpar)$ in \eq\eqref{eq:expected return}, we use gradient
information $\d J^\pi(\vec\polpar)/\d\vec\polpar$. We require that the
expected cost in \eq\eqref{eq:E(c)} is differentiable with respect to
the moments of the state distribution. Moreover, we assume that the
moments of the control distribution $\vec\mu_u$ and $\mat\Sigma_u$ can
be computed analytically and are differentiable with respect to the
policy parameters $\vec\polpar$.

In the following, we describe how to analytically compute these
gradients for a gradient-based policy search. We obtain the gradient
$\d J^\pi/\d\vec\polpar$ by repeated application of the chain-rule:
First, we move the gradient into the sum in \eq(\ref{eq:expected
  return}), and with $\mathcal E_t\coloneqq \E_{\vec x_t}[c(\vec
x_t)]$ we obtain
\begin{align}
  \frac{\d J^\pi(\vec \polpar)}{\d\vec\polpar} &=
  \sum\nolimits_{t=1}^T \frac{\d\mathcal E_t}{\d\vec\polpar}\,, \nonumber \\
  \frac{\d\mathcal E_t}{\d\vec\polpar}&= \frac{\d\mathcal
    E_t}{\d\prob(\vec x_t)}\frac{\d\prob(\vec x_t)}{\d\vec\polpar}
  \coloneqq \frac{\partial\mathcal E_t
  }{\partial\vec\mu_t}\green{\frac{\d\vec\mu_t}{\d\vec\polpar}} +
  \frac{\partial\mathcal E_t
  }{\partial\mat\Sigma_t}\frac{\d\mat\Sigma_t}{\d\vec\polpar}\,,
  \label{eq:chain rule}
\end{align} 
where we used the shorthand notation $\d \mathcal E_t/\d\prob(\vec
x_t) = \{\d\mathcal E_t/\d\vec\mu_t,\d\mathcal E_t/\d\mat\Sigma_t\}$
for taking the derivative of $\mathcal E_t$ with respect to both the
mean and covariance of $\prob(\vec x_t)=\gaussx{\vec
  x_t}{\vec\mu_t}{\mat\Sigma_t}$. Second, as we will show in
\sec~\ref{sec:approximate inference}, the predicted mean $\vec\mu_t$
and covariance $\mat\Sigma_t$ depend on the moments of $\prob(\vec
x_{t-1})$ and the controller parameters $\vec\polpar$. By applying the
chain-rule to \eq(\ref{eq:chain rule}), we obtain then
\begin{align}
  \frac{\d\prob(\vec x_t)}{\d\vec\polpar} &=
  \blue{\frac{\partial\prob(\vec x_t)}{\partial\prob(\vec
      x_{t-1})}}\frac{\d\prob(\vec x_{t-1})}{\d\vec\polpar} +
  \frac{\partial\prob(\vec
    x_t)}{\partial\vec\polpar}\,,\label{eq:dp(x)dpsi}\\
  \blue{\frac{\partial\prob(\vec x_t)}{\partial\prob(\vec
      x_{t-1})}}&=\left\{
    \red{\frac{\partial\vec\mu_t}{\partial\prob(\vec x_{t-1})}} ,
    \frac{\partial\mat\Sigma_t}{\partial\prob(\vec
      x_{t-1})}\right\}\,.
\label{eq:dp(x_t)dp(x_t-1)}
\end{align}
From here onward, we focus on \green{$\d\vec\mu_t/\d\vec\polpar$}, see
\eq(\ref{eq:chain rule}), but computing
$\d\mat\Sigma_t/\d\vec\polpar$ in \eq(\ref{eq:chain rule}) is
similar. For \green{$\d\vec\mu_t/\d\vec\polpar$}, we compute the
derivative
\begin{align}
  \green{\frac{\d\vec\mu_t}{\d\vec\polpar}} &= \red{\frac{\partial
    \vec\mu_t}{\partial\vec\mu_{t-1}}}\frac{\d
    \vec\mu_{t-1}}{\d\vec\polpar} + \red{\frac{\partial
    \vec\mu_t}{\partial\mat\Sigma_{t-1}}}\frac{\d
    \mat\Sigma_{t-1}}{\d\vec\polpar} +
   \cyan{\frac{\partial\vec\mu_t}{\partial\vec\polpar}}\,. 
\end{align}
Since $\d\prob(\vec x_{t-1})/\d\vec\polpar$ in
\eq(\ref{eq:dp(x)dpsi}) is known from time step $t-1$ and
\red{$\partial\vec\mu_t/\partial\prob(\vec x_{t-1})$} is computed by
applying the chain-rule to \eqs(\ref{eq: pred. mean uncertain
  input})--(\ref{eq:nu_i}), we conclude with
\begin{align}
 \cyan{\frac{\partial\vec\mu_t}{\partial\vec\polpar}}\!&=\!
  \frac{\partial\vec\mu_{\vec\Delta}}{\partial\prob(\vec
    u_{t-1})}\frac{\partial\prob(\vec u_{t-1})}{\partial\vec\polpar}
\!=\!\frac{\partial\vec\mu_{\vec\Delta}}{\partial\vec\mu_u}
  \frac{\partial\vec\mu_u}{\partial\vec\polpar}\!+\!
  \frac{\partial\vec\mu_{\vec\Delta}}{\partial\mat\Sigma_u}
  \frac{\partial\mat\Sigma_u}{\partial\vec\polpar}\,.
\label{eq:last eq in gradient chain}
\end{align}
The partial derivatives of $\vec\mu_u$ and $\mat\Sigma_u$, i.e., the
mean and covariance of $\prob(\vec u_t)$, used in \eq(\ref{eq:last eq
  in gradient chain}) depend on the policy representation. The
individual partial derivatives in \eqs(\ref{eq:chain
  rule})--(\ref{eq:last eq in gradient chain}) depend on the
approximate inference method used for propagating state distributions
through time. For example, with moment matching or linearization of
the posterior GP (see \sec~\ref{sec:approximate inference} for
details) the desired gradients can be computed analytically by
repeated application of the chain-rule. The Appendix derives the
gradients for the moment-matching approximation.

A gradient-based optimization method using \emph{estimates} of the
gradient of $J^\pi(\vec\theta)$ such as finite differences or more
efficient sampling-based methods (see~\cite{Peters2008b} for an
overview) requires many function evaluations, which can be
computationally expensive. However, since in our case policy
evaluation can be performed analytically, we profit from analytic
expressions for the gradients, which allows for standard
gradient-based non-convex optimization methods, such as CG or BFGS,
to determine optimized policy parameters $\vec\polpar^*$.


%

\section{Long-Term Predictions}
\label{sec:approximate inference}
Long-term predictions $p(\vec x_1),\dotsc,p(\vec x_T)$ for a given
policy parametrization are essential for policy evaluation and
improvement as described in \secs~\ref{sec:policy evaluation}
and~\ref{sec:controller learning}, respectively. These long-term
predictions are computed iteratively: At each time step,
\textsc{pilco} approximates the predictive state distribution $p(\vec
x_{t+1})$ by a Gaussian, see
\eqs\eqref{eq:mu_t}--\eqref{eq:Sigma_t}. For this approximation, we
need to predict with GPs when the input is given by a probability
distribution $p(\tilde{\vec x}_t)$, see \eq\eqref{pred. with
  uncert. inputs}.
In this section, we detail the computations of the mean
$\vec\mu_{\vec\Delta}$ and covariance matrix $\mat\Sigma_{\vec\Delta}$
of the GP predictive distribution, see \eq(\ref{pred. with
  uncert. inputs}), as well as the cross-covariances $\cov[\tilde{\vec
  x}_{t},\vec\Delta_t] = \cov\big[[\vec x_{t}\T,\vec
u_{t}\T]\T,\vec\Delta_t\big]$, which are required in
\eqs\eqref{eq:mu_t}--\eqref{eq:Sigma_t}.  
We present two approximations to predicting with GPs at uncertain
inputs: Moment matching~\cite{Deisenroth2011c,Quinonero-Candela2003a}
and linearization of the posterior GP mean
function~\cite{Ko2008}. While moment matching computes the first two
moments of the predictive distribution exactly, their approximation by
explicit linearization of the posterior GP is computationally
advantageous.

\subsection{Moment Matching}
\label{sec:moment matching}
Following the law of iterated expectations, for target dimensions
$a=1,\dotsc,D,$ we obtain the \emph{predictive mean}
\begin{align}
  \mu_{\vec\Delta}^a&=\E_{\tilde{\vec x}_{t}}[\E_{f_a}[f_a(\tilde{\vec
    x}_{t})|\tilde{\vec x}_{t}]]= \E_{\tilde{\vec
      x}_{t}}[m_{f_a}(\tilde{\vec x}_{t})]\nonumber\\ 
  &= \int m_{f_a}(\tilde{\vec x}_{t})\gaussx{\tilde{\vec
      x}_{t}}{\tilde{\vec\mu}_{t}}{\tilde{\mat\Sigma}_{t}}\d
  \tilde{\vec x}_{t}=\vec\beta_a\T\vec q_a\,,
  \label{eq: pred. mean uncertain input}\\
  \vec\beta_a &= (\mat K_a + \sigma_{w_a}^2)\inv\vec y_a\,,
\label{eq:beta}
\end{align}
with $\vec q_a = [q_{a_1}, \ldots, q_{a_n}]\T$. The entries of $\vec
q_a \in\R^n$ are computed using standard results from multiplying and
integrating over Gaussians and are given by
\begin{align}
  q_{a_i} &= \int k_a(\tilde{\vec x}_i,\tilde{\vec
    x}_{t})\gaussx{\tilde{\vec
      x}_{t}}{\tilde{\vec\mu}_{t}}{\tilde{\mat\Sigma}_{t}}
  \d\tilde{\vec x}_{t}\label{eq:q_i}\\
  & = \sigma_{f_a}^2|\tilde{\mat\Sigma}_{t}\mat\Lambda_a\inv + \mat
  I|^{-\frac{1}{2}}\exp\big(-\tfrac{1}{2}\vec\nu_i\T(\tilde{\mat\Sigma}_{t}
  + \mat\Lambda_a)\inv\vec\nu_i\big) \nonumber\,,
\end{align}
where  we define
\begin{align}
  \vec\nu_i\coloneqq (\tilde{\vec x}_i - \tilde{\vec\mu}_{t})
\label{eq:nu_i}
\end{align}
as the difference between the training input $\tilde{\vec x}_i$ and
the mean of the test input distribution $\prob(\vec x_{t},\vec
u_{t})$.

Computing the \emph{predictive covariance matrix}
$\mat\Sigma_{\vec\Delta}\in\R^{D\times D}$ requires us to distinguish
between diagonal elements $\sigma_{aa}^2$ and off-diagonal elements
$\sigma_{ab}^2$, $a\neq b$: Using the law of total
\mbox{(co-)}variance, we obtain for target dimensions $a,b=1,\dotsc,D$
\begin{align}
  \sigma_{aa}^2 &= \E_{\tilde{\vec x}_{t}}\big[\var_f
  [\Delta_a|\tilde{\vec x}_{t}]\big]+ \E_{f,\tilde{\vec
    x}_{t}}[\Delta_a^2]-(\vec\mu_{\vec\Delta}^a)^2
\label{eq:diagonal entry def.}\,,\\
\sigma_{ab}^2 &= \E_{f,\tilde{\vec
    x}_{t}}[\Delta_a\Delta_b]\!-\!\vec\mu_{\vec\Delta}^a\vec\mu_{\vec\Delta}^b\,,\quad
a\neq b\,,
\label{eq:off-diagonal entry def}
\end{align}
respectively, where $\mu_{\vec\Delta}^a$ is known from \eq(\ref{eq:
  pred. mean uncertain input}). The off-diagonal terms $\sigma_{ab}^2$
do not contain the additional term $\E_{\tilde{\vec x}_{t}}[\cov_f
[\Delta_a,\Delta_b|\tilde{\vec x}_{t}]]$ because of the conditional
independence assumption of the GP models: Different target dimensions
do not covary for given $\tilde{\vec x}_{t}$.

We start the computation of the covariance matrix with the terms that
are common to both the diagonal and the off-diagonal entries: With
$\prob(\tilde{\vec x}_{t}) = \gaussx{\tilde{\vec
    x}_{t}}{\tilde{\vec\mu}_{t}}{\tilde{\mat\Sigma}_{t}}$ and
the law of iterated expectations, we obtain
\begin{align}
  \E_{f,\tilde{\vec x}_{t}} [\Delta_a\Delta_b]&=\E_{\tilde{\vec
      x}_{t}}\big[\E_f[\Delta_a|\tilde{\vec
    x}_{t}]\,\E_f[\Delta_b|\tilde{\vec x}_{t}]\big]\nonumber\\
  &\stackrel{(\ref{predicted mean})}{=}\int m_f^a(\tilde{\vec x}_{t})
  m_f^b(\tilde{\vec x}_{t})\prob(\tilde{\vec x}_{t})\d\tilde{\vec
    x}_{t}
\end{align}
because of the conditional independence of $\Delta_a$ and $\Delta_b$
given $\tilde{\vec x}_{t}$. Using the definition of the GP mean
function in \eq(\ref{predicted mean}), we obtain
\begin{align}
  &\E_{f,\tilde{\vec x}_{t}}[\Delta_a\Delta_b] =\vec\beta_a\T\mat
  Q\vec\beta_b\,,
\label{eq:off-diagonal intermediate result}
\\
&\mat Q\coloneqq\int k_a(\tilde{\vec x}_{t},\tilde{\mat X})\T \, k_b(\tilde{\vec
  x}_{t},\tilde{\mat X})\prob(\tilde{\vec x}_{t})\d\tilde{\vec x}_{t}\,.
\label{eq:definition Q matrix as an integral}
\end{align}
Using standard results from Gaussian multiplications and integration,
we obtain the entries $Q_{ij}$ of $\mat Q\in\R^{n\times n}$
\begin{align}
  Q_{ij}
  &=|\mat R|^{-\tfrac{1}{2}}k_a(\tilde{\vec
    x}_i,\tilde{\vec\mu}_{t})k_b(\tilde{\vec x}_j,
  \tilde{\vec\mu}_{t})\exp\big(\tfrac{1}{2}\vec z_{ij}\T\mat T\inv
  \vec z_{ij}\big)
\label{eq:Q-matrix entries}
\end{align}
where we define
\begin{align*}
  \mat R&\coloneqq
  \tilde{\mat\Sigma}_{t}(\mat\Lambda_a\inv+\mat\Lambda_b\inv)+\mat
  I\,,\quad \mat T \coloneqq \mat\Lambda_a\inv + \mat\Lambda_b\inv +\tilde{\mat\Sigma}_{t}\inv\,,\\
  \vec
  z_{ij}&\coloneqq\mat\Lambda_a\inv\vec\nu_i+\mat\Lambda_b\inv\vec\nu_j\,,
\end{align*}
with $\vec \nu_i$ defined in \eq(\ref{eq:nu_i}).  Hence, the
off-diagonal entries of $\mat\Sigma_{\vec\Delta}$ are fully determined by
\eqs(\ref{eq: pred. mean uncertain input})--(\ref{eq:nu_i}),
(\ref{eq:off-diagonal entry def}), and (\ref{eq:off-diagonal intermediate
  result})--(\ref{eq:Q-matrix entries}).

From \eq(\ref{eq:diagonal entry def.}), we see that the diagonal
entries contain the additional term
\begin{align}
\hspace{-2mm} \E_{\tilde{\vec x}_{t}}\big[\var_f [\Delta_a|\tilde{\vec
    x}_{t}]\big]\!&=\! \sigma_{f_a}^2 - \tr\big((\mat K_a\! +\!
  \sigma_{w_a}^2\mat I)\inv\mat Q\big) + \sigma_{w_a}^2
\label{eq:expected signal variance}
\end{align}
with $\mat Q$ given in \eq(\ref{eq:Q-matrix entries}) and
$\sigma_{w_a}^2$ being the system noise variance of the $a$th target
dimension. This term is the expected variance of the function, see
\eq\eqref{predicted variance}, under the distribution  $p(\tilde{\vec
  x}_{t})$.

To obtain the \emph{cross-covariances} $\cov[\vec x_{t}, \vec\Delta_t]$ in
\eq\eqref{eq:Sigma_t}, we compute the cross-covariance
$\cov[\tilde{\vec x}_t,\vec\Delta_t]$ between an uncertain state-action
pair $\tilde{\vec
  x}_{t}\sim\gauss{\tilde{\vec\mu}_{t}}{\tilde{\mat\Sigma}_{t}}$
and the corresponding predicted state difference $\vec x_{t+1}-\vec
x_{t} = \vec\Delta_t\sim\gauss{\vec\mu_{\vec\Delta}}{\mat\Sigma_{\vec\Delta}}$. This
cross-covariance is given by
\begin{align}
  \cov[\tilde{\vec x}_{t},\vec\Delta_t] &=
     \E_{\tilde{\vec
      x}_{t},f}[\tilde{\vec x}_{t}
  \vec\Delta_t\T]\!-\!\tilde{\vec\mu}_{t}\vec\mu_{\vec\Delta}\T\,,
\end{align}
where the components of $\vec\mu_{\vec\Delta}$ are given in \eq\eqref{eq:
  pred. mean uncertain input}, and $\tilde{\vec\mu}_{t}$ is the known
mean of the input distribution of the state-action pair at time step
$t$.

Using the law of iterated expectation, for each state dimension
$a=1,\dotsc,D$, we compute $\E_{\tilde{\vec x}_{t},f}[\tilde{\vec
  x}_{t}\, \Delta_t^a]$ as
\begin{align}
  \E_{\tilde{\vec x}_{t},f}[\tilde{\vec x}_{t}\, \Delta_t^a]
  &=\E_{\tilde{\vec x}_{t}}[\tilde{\vec
    x}_{t}\,\E_{f}[\Delta_t^a|\tilde{\vec x}_{t}]] =\!\! \int\!\tilde{\vec
    x}_{t}
  m_f^a(\tilde{\vec x}_{t})\prob(\tilde{\vec x}_{t})\d\tilde{\vec
    x}_{t}\nonumber\\ 
  &\stackrel{\eqref{predicted mean}}{=}\int \tilde{\vec x}_{t}
  \Big(\sum_{i=1}^n\beta_{a_i}\, k_f^a(\tilde{\vec x}_{t},\tilde{\vec
    x}_i)\Big) \prob(\tilde{\vec x}_{t})\d\tilde{\vec x}_{t}\,,
\end{align}
where the (posterior) GP mean function $m_f(\tilde{\vec x}_{t})$ was
represented as a finite kernel expansion. Note that $\tilde{\vec x}_i$
are the state-action pairs, which were used to train the dynamics GP
model. By pulling the constant $\beta_{a_i}$ out of the integral and
changing the order of summation and integration, we obtain
\begin{align}
 & \E_{\tilde{\vec x}_{t},f}[\tilde{\vec x}_{t}\, \Delta_t^a] \nonumber\\
&=
  \sum_{i=1}^n\beta_{a_i} \int\tilde{\vec
    x}_{t}\,\underbrace{c_1\,\mathcal N(\tilde{\vec
      x}_{t}|\tilde{\vec x}_i,\mat\Lambda_a)}_{=k_f^a(\tilde{\vec
      x}_{t},\tilde{\vec x}_i)}\underbrace{\mathcal N(\tilde{\vec
      x}_{t}|\tilde{\vec\mu}_{t},\tilde{\mat\Sigma}_{t})}_{\prob(\tilde{\vec
      x}_{t})}\d\tilde{\vec x}_{t}\,,\label{eq:prod 2 gaussians}
\end{align}
where we define $\new{c_1\coloneqq
  \sigma_{f_a}^{2}(2\pi)^{\tfrac{D+F}{2}}|\mat\Lambda_a|^{\tfrac{1}{2}}}$
with $\tilde{\vec x}\in\R^{D+F}$, such that $k_f^a(\tilde{\vec
  x}_{t},\tilde{\vec x}_i)=c_1\gaussx{\tilde{\vec x}_{t}}{\tilde{\vec
    x}_i}{\mat\Lambda_a}$ is an unnormalized Gaussian probability
distribution in $\tilde{\vec x}_{t}$, where $\tilde{\vec x}_i$,
$i=1,\dotsc,n$, are the GP training inputs. The product of the two
Gaussians in \eq\eqref{eq:prod 2 gaussians} yields a new
(unnormalized) Gaussian $c_2\inv\gaussx{\tilde{\vec
    x}_{t}}{\vec\psi_i}{\mat\Psi}$ with
\begin{align}
  c_2\inv&=(2\pi)^{-\tfrac{D+F}{2}}|\mat\Lambda_a+\tilde{\mat\Sigma}_{t}|^{-\tfrac{1}{2}}\nonumber\\
  &\quad\times\exp\big(-\tfrac{1}{2}(\tilde{\vec x}_i
  -\tilde{\vec\mu}_{t})\T(\mat\Lambda_a+\tilde{\mat\Sigma}_{t})\inv(\tilde{\vec
    x}_i
  -\tilde{\vec\mu}_{t})\big)\,,\nonumber\\
  \mat\Psi &=
  (\mat\Lambda_a\inv+\tilde{\mat\Sigma}_{t}\inv)\inv\,,\quad
  \vec\psi_i = \mat\Psi(\mat\Lambda_a\inv\tilde{\vec x}_i +
  \tilde{\mat\Sigma}_{t}\inv\tilde{\vec\mu}_{t})\,. \nonumber
\end{align}
By pulling all remaining variables, which are independent of $\tilde{\vec
  x}_{t}$, out of the integral in \eq(\ref{eq:prod 2 gaussians}), the
integral determines the expected value of the product of the two
Gaussians, $\vec\psi_i$. Hence, we obtain
\begin{align}
  \E_{\tilde{\vec x}_{t},f}[\tilde{\vec x}_{t}\, \Delta_t^a] &\!=\!
  \sum\nolimits_{i=1}^n c_1c_2\inv\beta_{a_i}\vec\psi_i\,,\,\, a=1,\dotsc,D\,,\nonumber\\
  \cov_{\tilde{\vec x}_{t},f}[\tilde{\vec x}_{t},\Delta_t^a] &\!=\!
  \sum\nolimits_{i=1}^nc_1c_2\inv
  \beta_{a_i}\vec\psi_i\!-\!\tilde{\vec\mu}_{t}\mu_{\vec\Delta}^a\,,\,\,
\label{eq:cross-cov. preliminary result}
\end{align}
for all predictive dimensions $a = 1,\dotsc, E$.  With $c_1c_2\inv =
q_{a_i}$, see \eq\eqref{eq:q_i}, and
$\vec\psi_i=\tilde{\mat\Sigma}_{t}(\tilde{\mat\Sigma}_{t} +
\mat\Lambda_a)\inv\tilde{\vec x}_i +
\mat\Lambda(\tilde{\mat\Sigma}_{t} +
\mat\Lambda_a)\inv\tilde{\vec\mu}_{t}$ we simplify
\eq(\ref{eq:cross-cov. preliminary result}) and obtain
\begin{align}
  \cov&_{\tilde{\vec x}_{t},f}[\tilde{\vec
    x}_{t},\Delta_t^a]
  = \sum\limits_{i=1}^n\beta_{a_i}
  q_{a_i}\tilde{\mat\Sigma}_{t}(\tilde{\mat\Sigma}_{t}\! +\!
  \mat\Lambda_a)\inv(\tilde{\vec
    x}_i\!-\!\tilde{\vec\mu}_{t})\,,
\label{eq:crosscov}
\end{align}
\new{$a=1,\dotsc, E$.  The desired covariance $\cov[\vec x_t,\Delta_t]$
is a $D\times E$ submatrix of the $(D+F)\times E$ cross-covariance
computed in to \eq\eqref{eq:crosscov}.}

A visualization of the approximation of the predictive distribution by
means of exact moment matching is given in \fig~\ref{fig:EKF vs ADF}.

\subsection{Linearization of the Posterior GP Mean Function}
\label{sec:linearization}
An alternative way of approximating the predictive distribution
$\prob(\vec\Delta_t)$ by a Gaussian for $\tilde{\vec x}_t
\sim\gaussx{\tilde{\vec x}_t}{\tilde{\vec
    \mu}_t}{\tilde{\mat\Sigma}_t}$ is to linearize the posterior GP
mean function.  \Fig~\ref{fig:EKF vs ADF} visualizes the approximation by means
of linearizing the posterior GP mean function.

The \emph{predicted mean} is obtained by evaluating the posterior GP
mean in \eq\eqref{eq:one-step prediction mean and cov} at the mean
$\tilde{\vec\mu}_t$ of the input distribution, i.e., \new{
\begin{align}
  \vec\mu_{\vec\Delta}^a &= \E_f[f_a(\tilde{\vec\mu}_{t})] =
  m_{f_a}(\tilde{\vec\mu}_t) = \vec\beta_a\T k_a(\tilde{\mat
    X},\tilde{\vec\mu}_t)\,,
\label{eq:r_a}
\end{align}
}
$a = 1,\dotsc, E$, where $\vec \beta_a$ is given in
\eq\eqref{eq:beta}.
%

To compute the GP \emph{predictive covariance matrix}
$\mat\Sigma_{\vec\Delta}$, we explicitly linearize the posterior GP
mean function around $\tilde{\vec\mu}_t$. By applying standard results
for mapping Gaussian distributions through linear models, the
predictive covariance is given by
\begin{align}
  \hspace{-2mm} \mat\Sigma_{\vec\Delta} &= \mat
  V\tilde{\mat\Sigma}_{t}\mat V\T + \mat\Sigma_w\,,
\label{eq:EKF pred covariance}
\\
\mat V &=
\frac{\partial\vec\mu_{\vec\Delta}}{\partial\tilde{\vec\mu}_{t}}=
\vec\beta\T\frac{\partial k(\tilde{\mat
    X},\tilde{\vec\mu}_t)}{\partial\tilde{\vec\mu}_{t}}\,.
\label{eq:EKF V-matrix}
\end{align}
In \eq\eqref{eq:EKF pred covariance}, $\mat\Sigma_w$ is a diagonal
matrix whose entries are the noise variances $\sigma_{w_a}^2$ plus the
model uncertainties $\var_f[\Delta_t^a|\tilde{\vec\mu}_t]$ evaluated
at $\tilde{\vec\mu}_{t}$, see \eq\eqref{predicted variance}. This
means, model uncertainty no longer depends on the density of the data
points. Instead it is assumed to be constant.
Note that the moments computed in \eqs\eqref{eq:r_a}--\eqref{eq:EKF
  pred covariance} are not exact.

The \emph{cross-covariance} $\cov[\vaugx_{t}, \vec\Delta_t]$ is
given by $\tilde{\mat\Sigma}_{t}\mat V$, where $\mat V$ is defined
in \eq(\ref{eq:EKF V-matrix}).

\section{Policy}
\label{sec:policy}
In the following, we describe the desired properties of the policy
within the \textsc{pilco} learning framework. First, to compute the
long-term predictions $\prob(\vec x_1),\dotsc,\prob(\vec x_T)$ for
policy evaluation, the policy must allow us to compute a distribution
over controls $\prob(\vec u)= \prob(\pi(\vec x))$ for a given
(Gaussian) state distribution $\prob(\vec x)$. Second, in a realistic
real-world application, the amplitudes of the control signals are
bounded. Ideally, the learning system takes these constraints
explicitly into account.  In the following, we detail how
\textsc{pilco} implements these desiderata.

\subsection{Predictive Distribution over Controls}
During the long-term predictions, the states are given by a
probability distribution $\prob(\vec x_t)$, $t=0,\dotsc,T$. The
probability distribution of the state $\vec x_t$ induces a predictive
distribution $\prob(\vec u_t) = \prob(\pi(\vec x_t))$ over controls,
even when the policy is deterministic. We approximate the distribution
over controls using moment matching, which is in many interesting cases analytically tractable.


\subsection{Constrained Control Signals}

\begin{figure}[tb]
\centering
\subfigure[Preliminary policy $\tilde\pi$ as a function of the state.]{
\includegraphics[width = \twofig]{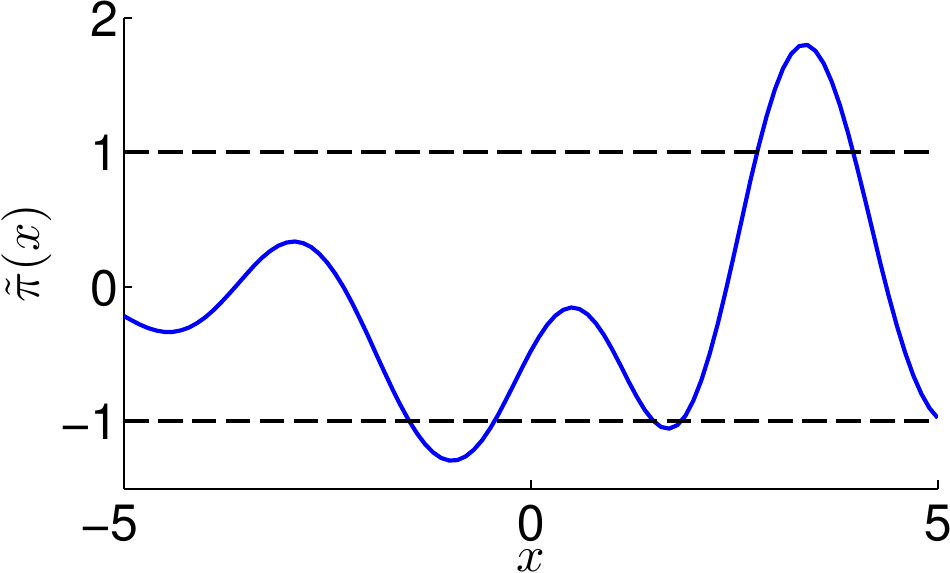}
\label{fig:unsquashed policy}
}
\hfill
\subfigure[Policy $\pi=\squash(\tilde\pi(x))$ as a function of the state.]{
\includegraphics[width =
\twofig]{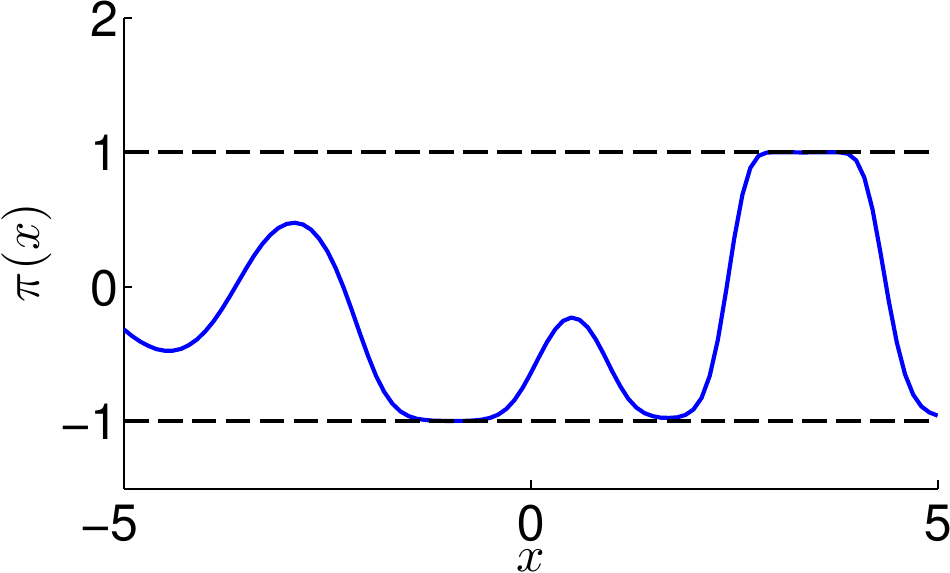}
\label{fig:squashed policy}
}
\caption[Constraining the control signal.]{Constraining the control
  signal. Panel~\subref{fig:unsquashed policy} shows an example of an
  unconstrained preliminary policy $\tilde\pi$ as a function of the
  state $x$. Panel~\subref{fig:squashed policy} shows the constrained
  policy $\pi(x) = \squash(\tilde\pi(x))$ as a function of the state $x$.}
\label{fig:constraining the policy}
\figspace
\end{figure}

In practical applications, force or torque limits are present and must
be accounted for during planning.  Suppose the control limits are such
that $\vec u\in[-\vec u_{\max}, \vec u_{\max}]$. Let us consider a
\emph{preliminary policy} $\tilde\pi$ with an unconstrained amplitude.
To account for the control limits coherently during simulation, we
squash the preliminary policy $\tilde\pi$ through a bounded and
differentiable squashing function, which limits the amplitude of the
final policy $\pi$. As a squashing function, we use
\begin{align}
\squash(x) =
\tfrac{9}{8}\sin(x) + \tfrac{1}{8}\sin(3x)\quad \in [-1,1]\,,
\label{eq:squashing function}
\end{align}
which is the third-order Fourier series expansion of a trapezoidal
wave, normalized to the interval \mbox{$[-1,1]$}. The squashing
function in \eq\eqref{eq:squashing function} is computationally
convenient as we can analytically compute predictive moments for
Gaussian distributed states. 
%
Subsequently, we multiply the squashed policy by $\vec
u_{\max}$ and obtain the final policy
\begin{equation}\label{eq:squashed policy}
  \pi(\vec x) = \vec u_{\max}\squash(\tilde\pi(\vec x))\in[-\vec
  u_{\max},\vec u_{\max}]\,,
\end{equation}
an illustration of which is shown in \fig~\ref{fig:constraining the policy}.
Although the squashing function in \eq\eqref{eq:squashing function}
is periodic, it is almost always used within a half wave if the
preliminary policy $\tilde\pi$ is initialized to produce function
values that do not exceed the domain of a single period. Therefore,
the periodicity does not matter in practice.
%

%
%
%
To compute a distribution over constrained control signals, we execute
the following steps:
\begin{align}
  \prob(\vec x_t) \mapsto \prob(\tilde{\pi}(\vec x_t)) \mapsto
  \prob(\vec u_{\max}\squash(\tilde\pi(\vec x_t))) = \prob(\vec u_t)\,.
\label{eq:distribution over controls (final)}
\end{align}
First, we map the Gaussian state distribution $\prob(\vec x_t)$
through the preliminary (unconstrained) policy $\tilde\pi$. Thus, we
require a preliminary policy $\tilde\pi$ that allows for closed-form
computation of the moments of the distribution over controls
$\prob(\tilde\pi(\vec x_t))$.  Second, we squash the approximate
Gaussian distribution $\prob(\tilde\pi(\vec x))$ according to
\eq(\ref{eq:squashed policy}) and compute exactly the mean and
variance of $\prob(\tilde\pi(\vec x))$. Details are given in the 
Appendix. We approximate $\prob(\tilde\pi(\vec
x))$ by a Gaussian with these moments,
yielding the distribution $\prob(\vec u)$ over controls in
\eq\eqref{eq:distribution over controls (final)}.

\subsection{Representations of the Preliminary Policy}
\label{sec:policy representation}
In the following, we present two representations of the preliminary
policy $\tilde\pi$, which allow for closed-form computations of the
mean and covariance of $\prob(\tilde\pi(\vec x))$ when the state $\vec
x$ is Gaussian distributed. We consider both a linear and a nonlinear
representations of $\tilde\pi$.

\subsubsection{Linear Policy}
The linear preliminary policy is given by
\begin{equation}\label{eq:policy linear}
\tilde\pi(\vec x_*) = \mat A\vec
x_* + \vec b\,,
\end{equation}
where $\mat A$ is a parameter matrix of weights and $\vec b$ is an
offset vector. In each control dimension $d$, the
policy in \eq(\ref{eq:policy linear}) is a linear combination of the states
(the weights are given by the $d$th row in $\mat A$) plus an offset
$b_d$.

The predictive distribution $\prob(\tilde\pi(\vec x_*))$ for a state
distribution $\vec x_*\sim\gauss{\vec\mu_*}{\mat\Sigma_*}$ is an exact
Gaussian with mean and covariance
\begin{align}
\E_{\vec x_*}[\tilde\pi(\vec x_*)] &= \mat A\vec\mu_* + \vec b\,,\quad
\cov_{\vec x_*}[\tilde\pi(\vec x_*)] = \mat A\mat\Sigma_*\mat A\T\,,
\end{align}
respectively. A drawback of the linear policy is that it is not
flexible. However, a linear controller can often be used for
stabilization around an equilibrium.




\subsubsection{Nonlinear Policy: Deterministic Gaussian Process}
\label{sec:detGP}

In the nonlinear case, we represent the preliminary policy $\tilde\pi$
by 
\begin{align}
\hspace{-2mm}  \tilde\pi(\vec x_*) \!=\! \sum_{i=1}^N k(\vec m_i,\vec x_*)(\mat
  K+\sigma_\pi^2\mat I)\inv\vec t = k(\mat M,\vec x_*)\T\vec\alpha\,,
\label{eq:detGP}
\end{align}
where $\vec x_*$ is a test input, $\vec\alpha = (\mat K+0.01\mat
I)\inv\vec t$, where $\vec t$ plays the role of a GP's training
targets. In \eq\eqref{eq:detGP}, $\mat M=[\vec m_1,\dotsc,\vec m_N]$
are the centers of the (axis-aligned) Gaussian basis functions
\begin{align}
  k(\vec x_p, \vec x_q) = \exp\big(-\tfrac{1}{2}(\vec x_p-\vec
  x_q)\T\mat\Lambda\inv(\vec x_p -\vec x_q)\big)\,.
\label{eq:kernel policy}
\end{align}
We call the policy representation in \eq\eqref{eq:detGP} a
\emph{deterministic GP} with a fixed number of $N$ basis
functions. Here, ``deterministic'' means that there is no uncertainty
about the underlying function, that is,
$\var_{\tilde\pi}[\tilde\pi(\vec x)]=0$. Therefore, the deterministic
GP is a degenerate model, which is functionally equivalent to a
regularized RBF network. The deterministic GP is functionally
equivalent to the posterior GP mean function in \eq\eqref{predicted
  mean}, where we set the signal variance to 1, see
\eq\eqref{eq:kernel policy}, and the noise variance to $0.01$.  As
the preliminary policy will be squashed through $\squash$ in
\eq\eqref{eq:squashing function} whose relevant support is the
interval $[-\tfrac{\pi}{2},\tfrac{\pi}{2}]$, a signal variance of 1 is
about right. Setting additionally the noise standard deviation to 0.1
corresponds to fixing the signal-to-noise ratio of the policy to 10
and, hence, the regularization.

For a Gaussian distributed state $\vec x_*\sim\gauss{\vec
  \mu_*}{\mat\Sigma_*}$, the \emph{predictive mean} of $\tilde\pi(\vec
x_*)$ as defined in \eq\eqref{eq:detGP} is given as
\begin{align}
  \E_{\vec x_*}[\tilde\pi(\vec x_*)] &= \vec \alpha_a\T \E_{\vec
    x_*}[k(\mat M, \vec x_*)]\nonumber\\
  &= \vec\alpha_a\T\int k(\mat M, \vec x_*)\prob(\vec x_*)\d\vec x_*
  =\vec\alpha_a\T\vec r_a\,,
\label{eq:pred mean prel pol}
\end{align}
where for $i=1,\dotsc,N$ and all policy dimensions $a = 1,\dotsc, F$
\begin{align}
r_{a_i} &= |\mat\Sigma_*\mat\Lambda_a\inv + \mat
I|^{-\tfrac{1}{2}}\nonumber\\
&\quad\times\exp(-\tfrac{1}{2}(\vec\mu_*
-\vec m_i)\T(\mat\Sigma_* + \mat\Lambda_a)\inv(\vec\mu_*
-\vec m_i))\,.\nonumber
\end{align}
The diagonal matrix $\mat\Lambda_a$ contains the squared length-scales
$\ell_i$, $i=1,\dotsc,D$. The predicted mean in
\eq\eqref{eq:pred mean prel pol} is equivalent to the standard
predicted GP mean in \eq\eqref{eq: pred. mean uncertain input}.

For $a,b=1,\dotsc,F$, the entries of the \emph{predictive covariance
  matrix} are computed according to
\begin{align*}
\cov_{\vec x_*}&[\tilde\pi_a(\vec x_*),\tilde\pi_b(\vec
x_*)]\\
&=\E_{\vec x_*}[\tilde\pi_a(\vec
x_*)\tilde\pi_b(\vec x_*)] - \E_{\vec x_*}[\tilde\pi_a(\vec
x_*)]\E_{\vec x_*}[\tilde\pi_b(\vec
x_*)]\,,
\end{align*}
where $\E_{\vec x_*}[\tilde\pi_{\{a,b\}}(\vec x_*)]$ is given in
\eq(\ref{eq:pred mean prel pol}). Hence, we focus on the term
$\E_{\vec x_*}[\tilde\pi_a(\vec x_*)\tilde\pi_b(\vec x_*)]$, which for
$a,b=1,\dotsc,F$ is given by
\begin{align*}
\E_{\vec x_*}[\tilde\pi_a(\vec x_*)\tilde\pi_b(\vec x_*)]&=\vec \alpha_a\T\E_{\vec x_*}[k_a(\mat M, \vec x_*)k_b(\mat M, \vec x_*)\T]\vec \alpha_b\\
&=\vec\alpha_a\T\mat Q\vec\alpha_b\,.
\end{align*}
For $i,j = 1,\dotsc, N$, we compute the entries of $\mat Q$ as
\begin{align*}
Q_{ij} &= \int k_a(\vec m_i, \vec x_*)k_b(\vec m_j, \vec x_*)
\prob(\vec x_*)\d\vec x_*\\
&=k_a(\vec m_i, \vec \mu_*)k_b(\vec m_j, \vec \mu_*)|\mat R|^{-\tfrac{1}{2}}\exp(\tfrac{1}{2}\vec
z_{ij}\T\mat T\inv\vec z_{ij})\,,\\
\mat R &= \mat\Sigma_*(\mat\Lambda_a\inv + \mat\Lambda_b\inv) + \mat
I\,, \quad \mat T = \mat\Lambda_a\inv + \mat\Lambda_b\inv + \mat\Sigma_*\inv\,,\\
\vec z_{ij} &= \mat\Lambda_a\inv(\vec\mu_*-\vec m_i) +
\mat\Lambda_b\inv(\vec\mu_*-\vec m_j)\,.
\end{align*}
Combining this result with \eq(\ref{eq:pred mean prel pol}) fully
determines the predictive covariance matrix of the preliminary policy.

Unlike the predictive covariance of a probabilistic GP, see
\eqs\eqref{eq:diagonal entry def.}--\eqref{eq:off-diagonal entry
  def}, the predictive covariance matrix of the deterministic GP does
not \new{comprise} any model uncertainty in its diagonal entries.


\subsection{Policy Parameters}
In the following, we describe the policy parameters for both the
linear and the nonlinear policy\footnote{For notational convenience,
  with a (non)linear policy we mean the (non)linear preliminary policy
  $\tilde\pi$ mapped through the squashing function $\squash$ and
  subsequently multiplied by $\vec u_{\max}$.}.  

\subsubsection{Linear Policy}
The linear policy in \eq~\eqref{eq:policy linear} possesses $D+1$
parameters per control dimension: For control dimension $d$ there are
$D$ weights in the $d$th row of the matrix $\mat A$. One additional
parameter originates from the offset parameter $b_d$.

\subsubsection{Nonlinear Policy}

The parameters of the deterministic GP in \eq(\ref{eq:detGP}) are the
locations $\mat M$ of the centers ($DN$ parameters), the (shared)
length-scales of the Gaussian basis functions ($D$ length-scale
parameters per target dimension), and the $N$ targets $\vec t$ per
target dimension. In the case of multivariate controls, the basis
function centers $\mat M$ are shared.

\subsection{Computing the Successor State Distribution}
\label{sec:succState}
Alg.~\ref{alg:succState} summarizes the computational steps required
to compute the successor state distribution $\prob(\vec x_{t+1})$ from
$\prob(\vec x_t)$.
%
\begin{algorithm}[tb]
   \caption{Computing the Successor State Distribution}
   \label{alg:succState}
\begin{algorithmic}[1]
   \STATE {\bfseries init:} $\vec
   x_t\sim\gauss{\vec\mu_t}{\mat\Sigma_t}$
\STATE Control distribution $\prob(\vec u_t) = \prob(\vec
u_{\max}\squash(\tilde\pi(\vec x_t,\vec\theta)))$ 
\STATE Joint state-control distribution $\prob(\vaugx_t) =
\prob(\vec x_t,\vec u_t)$
\STATE Predictive GP distribution of change in state
$\prob(\vec\Delta_{t})$
\STATE Distribution of successor state $\prob(\vec x_{t+1})$
\end{algorithmic}
\end{algorithm}
The computation of a distribution over controls $\prob(\vec u_{t})$
from the state distribution $\prob(\vec x_{t})$ requires two steps:
First, for a Gaussian state distribution $\prob(\vec x_{t})$ at time
$t$ a Gaussian approximation of the distribution $\prob(\tilde\pi(\vec
x_{t}))$ of the preliminary policy is computed analytically. Second,
the preliminary policy is squashed through $\squash$ and an
approximate Gaussian distribution of $\prob(\vec
u_{\max}\squash(\tilde\pi(\vec x_{t})))$ is computed analytically in
\eq(\ref{eq:distribution over controls (final)}) using results from
the Appendix. Third, we analytically compute a Gaussian
approximation to the joint distribution $\prob(\vec x_{t},\vec
u_{t})=\prob(\vec x_{t},\pi(\vec x_{t}))$. For this, we compute (a)
a Gaussian approximation to the joint distribution $\prob(\vec
x_{t},\tilde\pi(\vec x_{t}))$, which is exact if $\tilde\pi$ is
linear, and (b) an approximate fully joint Gaussian
distribution $\prob(\vec x_{t},\tilde\pi(\vec x_{t}),\vec u_t)$. We
obtain cross-covariance information between the state $\vec x_{t}$ and
the control signal $\vec u_{t} = \vec u_{\max}\squash(\tilde\pi(\vec
x_{t}))$ via
\begin{align*}
  \cov[\vec x_{t},\vec u_{t}]\!=\!\cov[\vec
  x_{t},\tilde\pi(\vec x_{t})]\cov[\tilde\pi(\vec
  x_{t}),\tilde\pi(\vec x_{t})]\inv\cov[\tilde\pi(\vec
  x_{t}),\vec u_{t}]\,, 
\end{align*}
where we exploit the conditional independence of $\vec x_t$ and $\vec
u_t$ given $\tilde\pi(\vec x_t)$.  Then, we integrate $\tilde\pi(\vec
x_{t})$ out to obtain the desired joint distribution $\prob(\vec
x_{t},\vec u_{t})$.  This leads to an approximate Gaussian joint
probability distribution $\prob(\vec x_{t},\vec u_{t})=\prob(\vec
x_{t},\pi(\vec x_{t}))=\prob(\vaugx_t)$. Fourth, with the approximate
Gaussian input distribution $\prob(\vaugx_t)$, the distribution
$\prob(\vec \Delta_t)$ of the change in state is computed using the
results from \sec~\ref{sec:approximate inference}.
Finally, the mean and covariance of a Gaussian approximation of the
successor state distribution $\prob(\vec x_{t+1})$ are given by
\eq\eqref{eq:mu_t} and~\eqref{eq:Sigma_t}, respectively.

%
All required computations can be performed analytically because of the
choice of the Gaussian covariance function for the GP dynamics model,
see \eq\eqref{eq:SE kernel}, the representations of the preliminary
policy $\tilde\pi$, see \sec~\ref{sec:policy representation}, and the
choice of the squashing function, see \eq(\ref{eq:squashing
  function}).

\section{Cost Function}\label{sec:cost function}
In our learning set-up, we use a cost function that solely penalizes
the Euclidean distance $d$ of the current state to the target state.
Using only distance penalties is often sufficient to solve a task:
Reaching a target $\vec x_{\text{target}}$ with high speed naturally
leads to overshooting and, thus, to high long-term costs. In
particular, we use the \new{generalized binary} saturating cost
\begin{equation}\label{eq:immediate cost}
  \cost(\vec x)=1-\exp\big(-\tfrac{1}{2\sigma_c^2}\,d(\vec x,\vec
  x_{\text{target}})^2\big) \in[0,1]\,,
\end{equation}
which is locally quadratic but saturates at unity for large deviations
$d$ from the desired target $\vec x_{\text{target}}$.
In \eq(\ref{eq:immediate cost}), the geometric distance from the
state $\vec x$ to the target state is denoted by $d$, and the
parameter $\sigma_c$ controls the width of the cost function.\footnote{In the
context of sensorimotor control, the saturating cost function in
\eq(\ref{eq:immediate cost}) resembles the cost function in human
reasoning as experimentally validated by~\cite{Kording2004}.}

In classical control, typically a quadratic cost is assumed. However,
a quadratic cost tends to focus attention on the worst deviation from
the target state along a predicted trajectory. In the early stages of
learning the predictive uncertainty is large and, therefore, the
policy gradients, which are described in \sec~\ref{sec:controller
  learning} become less useful. Therefore, we use the saturating cost
in \eq\eqref{eq:immediate cost} as a default within the
\textsc{pilco} learning framework.

The immediate cost in \eq(\ref{eq:immediate cost}) is an unnormalized
Gaussian with mean $\vec x_{\text{target}}$ and variance $\sigma_c^2$,
subtracted from unity.
Therefore, the expected immediate cost can be computed analytically
according to
\begin{align}
  &\E_{\vec x}[\cost(\vec x)]=\int c(\vec x)\prob(\vec x)\d\vec x \label{eq:E(c(x))}\\
&=1-\int
  \exp\big(-\tfrac{1}{2}(\vec x-\vec x_{\text{target}})\T \mat
  T\inv(\vec x-\vec x_{\text{target}})\big)\prob(\vec x)\d\vec x\,, 
 \nonumber
\end{align}
where $\mat T\inv$
is the precision matrix of the unnormalized Gaussian in
\eq(\ref{eq:E(c(x))}). 
If the state $\vec x$ has the same representation as the target
vector, $\mat T\inv$ is a diagonal matrix with entries either unity or
zero, scaled by $1/\sigma_c^2$.  Hence, for $\vec
x\sim\gauss{\vec\mu}{\mat\Sigma}$ we obtain the expected immediate
cost
\begin{align}
  \E_{\vec x}[\cost(\vec x)] &=1-| \mat I+\mat\Sigma\mat
  T\inv|^{-1/2}\nonumber\\
&\quad\times\exp(-\tfrac{1}{2}(\vec\mu-\vec
  x_{\text{target}})\T\tilde{\mat S}_1(\vec\mu-\vec
  x_{\text{target}}))\,,\label{eq:expected cost equation}\\
  \tilde{\mat S}_1&\coloneqq \mat
T\inv(\mat I+\mat\Sigma\mat T\inv)\inv\label{eq:S1-matrix}\,.
\end{align}
The partial derivatives
$
\tfrac{\partial}{\partial\vec\mu_t}\E_{\vec x_t}[\cost(\vec x_t)],\,
\tfrac{\partial}{\partial\mat\Sigma_t}\E_{\vec x_t}[\cost(\vec x_t)]
$
of the immediate cost with respect to the mean and the covariance of
the state distribution $\prob(\vec x_t)=\gauss{\vec\mu_t}{\mat\Sigma_t}$,
which are required to compute the policy gradients analytically, are
given by
\begin{align}
  \frac{\partial \E_{\vec x_t}[\cost(\vec
  x_t)]}{\partial\vec\mu_t}&=-\E_{\vec x_t}[\cost(\vec
  x_t)]\,(\vec\mu_t-\vec x_{\text{target}})\T\tilde{\mat S}_1\,,
  \label{eq:E[c]/dm sat cost}\\
  \frac{\partial \E_{\vec x_t}[\cost(\vec
  x_t)]}{\partial\mat\Sigma_t}&=\tfrac{1}{2}\E_{\vec x_t}[\cost(\vec
x_t)]\label{eq:E[c]/dS sat cost}\\
&\quad\times\big(\tilde{\mat
    S}_1(\vec\mu_t-\vec x_{\text{target}})(\vec\mu_t-\vec
  x_{\text{target}})\T-\mat I\big)\tilde{\mat S}_1\,,
  \nonumber
\end{align}
respectively, where $\tilde{\mat S}_1$ is given in
\eq(\ref{eq:S1-matrix}). 

\subsection{Exploration and Exploitation}\label{sec:cost exploration}
\begin{figure}[tb]
  \centering \subfigure[When the mean of the state is far
  away from the target, uncertain states (red, dashed-dotted) are
  preferred to more certain states with a more peaked distribution
  (black, dashed). This leads to initial exploration.]{
\includegraphics[ width = \twofig]{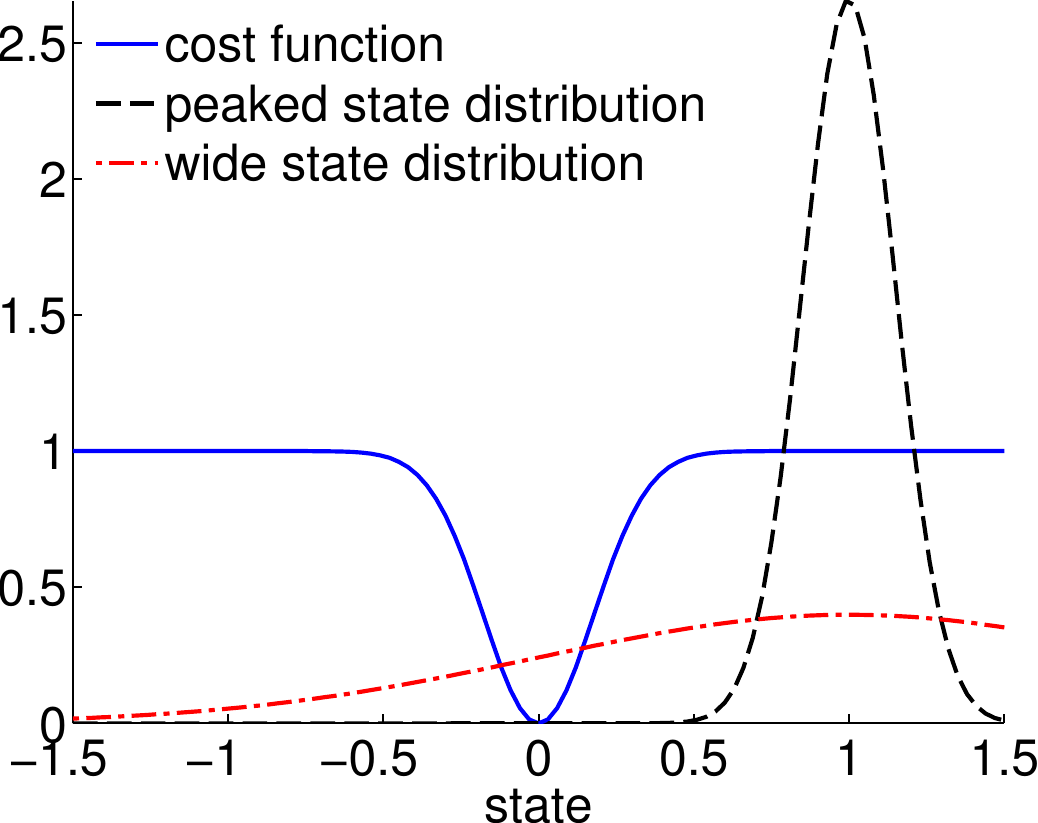}
\label{fig:exploration1}}
\hfill
\subfigure[When the mean of the state is close to the target,
peaked state distributions (black, dashed) cause less expected cost
and, thus, are preferable to more uncertain states (red,
dashed-dotted), leading to exploitation close to the target.]{
\includegraphics[ width = \twofig]{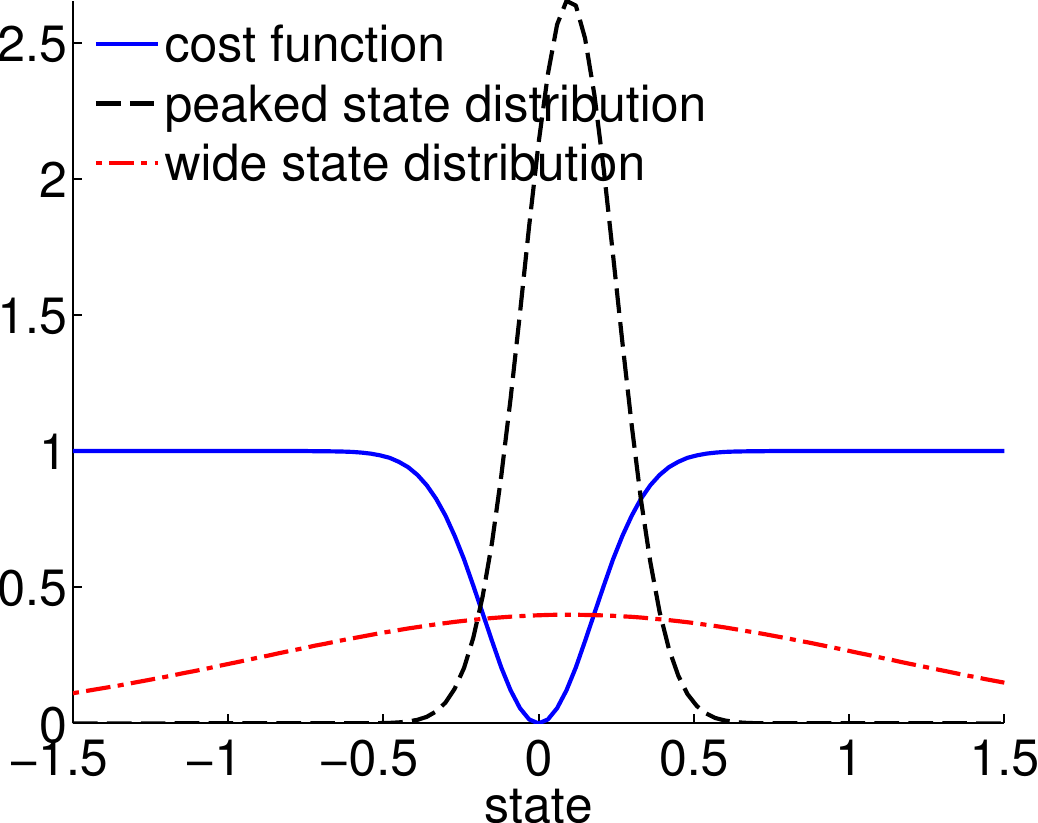}
\label{fig:exploration2}}
\caption{Automatic exploration and exploitation with the saturating
  cost function (blue, solid). The $x$-axes describe the state space.
  The target state is the origin.}
\label{fig:exploration cost}
\figspace
\end{figure}

The saturating cost function in \eq(\ref{eq:immediate cost}) allows
for a natural exploration when the policy aims to minimize the
expected long-term cost in \eq\eqref{eq:expected return}.  This
property is illustrated in \fig~\ref{fig:exploration cost} for a
single time step where we assume a Gaussian state distribution
$\prob(\vec x_t)$.
If the mean of $\prob(\vec x_t)$ is \emph{far away from the target}
$\vec x_{\text{target}}$, a wide state distribution is more likely to
have substantial tails in some low-cost region than a more peaked
distribution as shown in \fig~\ref{fig:exploration1}. In the early
stages of learning, the predictive state uncertainty is largely due to
propagating model uncertainties forward. If we predict a state
distribution in a high-cost region, the saturating cost then leads to
automatic \emph{exploration} by favoring uncertain states, i.e.,
states in regions far from the target with a poor dynamics model. When
visiting these regions during interaction with the physical system,
subsequent model learning reduces the model uncertainty locally. In
the subsequent policy evaluation, \textsc{pilco} will predict a
tighter state distribution in the situations described in
\fig~\ref{fig:exploration cost}.

If the mean of the state distribution is \emph{close to the target} as
in \fig~\ref{fig:exploration2}, wide distributions are likely to have
substantial tails in high-cost regions. By contrast, the mass of a
peaked distribution is more concentrated in low-cost regions. In this
case, the policy prefers peaked distributions close to the target,
leading to \emph{exploitation}.

To summarize, combining a probabilistic dynamics model, Bayesian
inference, and a saturating cost leads to automatic exploration as
long as the predictions are far from the target---even for a policy,
which greedily minimizes the expected cost. Once close to the target,
the policy does not substantially deviate from a confident trajectory
that leads the system close to the target.\footnote{Code is available
  at \url{http://mloss.org/software/view/508/}.}

\section{Experimental Results}
\label{sec:results}

In this section, we assess \textsc{pilco}'s key properties and show
that \textsc{pilco} scales to high-dimensional control
problems. Moreover, we demonstrate the hardware applicability of our
learning framework on two real systems.  In all cases, \textsc{pilco}
followed the steps outlined in \alg~\ref{alg:pilco}. To reduce the
computational burden, we used the sparse GP method
of~\cite{Snelson2006} after 300 collected data points.

\subsection{Evaluation of Key Properties}
In the following, we assess the quality of the approximate inference
method used for long-term predictions in terms of computational demand
and learning speed. Moreover, we shed some light on the quality of the
Gaussian approximations of the predictive state distributions and the
importance of Bayesian averaging. For these assessments, we applied
\textsc{pilco} to two nonlinear control tasks, which are introduced in
the following.

\subsubsection{Task Descriptions}
We considered two simulated tasks (double-pendulum swing-up, cart-pole
swing-up) to evaluate important properties of the \textsc{pilco}
policy search framework: learning speed, quality of approximate
inference, importance of Bayesian averaging, and hardware
applicability. In the following we briefly introduce the experimental
set-ups.

\paragraph{\bf Double-Pendulum Swing-Up with Two Actuators}
\label{sec:double pendulum}

\begin{figure}
\centering
\includegraphics[height = 4.5cm]{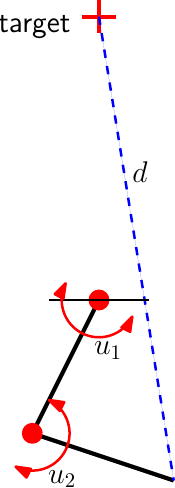}
\caption{Double pendulum with two actuators applying torques $u_1$
and $u_2$. The cost function penalizes the distance $d$ to the target.}
\label{fig:double pendulum}
\end{figure}

The double pendulum system is a two-link robot arm with two actuators,
see \fig~\ref{fig:double pendulum}. The state $\vec x$ is given by
the angles $\theta_1,\theta_2$ and the corresponding angular
velocities $\dot\theta_1,\dot\theta_2$ of the inner and outer link,
respectively, measured from being upright. Each link was of length
$\unit[1]{m}$ and mass $\unit[0.5]{kg}$. Both torques $u_1$ and $u_2$
were constrained to $[-3,3]\,\unit{Nm}$. The control signal could be
changed every $\unit[100]{ms}$. In the meantime it was constant
(zero-order-hold control).  The objective was to learn a controller
that swings the double pendulum up from an initial distribution
$\prob(\vec x_0)$ around $\vec\mu_0=[\pi, \pi, 0, 0]\T$ and balances
it in the inverted position with $\theta_1=0=\theta_2$. The prediction
horizon was $\unit[2.5]{s}$.

The task is challenging since its solution requires the interplay of
two correlated control signals. The challenge is to automatically
learn this interplay from experience.
%
To solve the double pendulum swing-up task, a nonlinear policy is
required. Thus, we parametrized the preliminary policy as a
deterministic GP, see \eq(\ref{eq:detGP}), with 100 basis functions
resulting in 812 policy parameters. We chose the saturating immediate
cost in \eq(\ref{eq:immediate cost}), where the Euclidean distance
between the upright position and the tip of the outer link was
penalized. We chose the cost width $\sigma_c=0.5$, which means that
the tip of the outer pendulum had to cross horizontal to achieve an
immediate cost smaller than unity.


\paragraph{\textbf{Cart-Pole Swing-Up}}
\label{sec:cartpole results}
The cart-pole system consists of a cart running on a track and a
freely swinging pendulum attached to the cart. The state of the system
is the position $x$ of the cart, the velocity $\dot x$ of the cart,
the angle $\theta$ of the pendulum measured from hanging downward, and
the angular velocity $\dot\theta$. A horizontal force
$u\in[-10,10]\,\unit{N}$ could be applied to the cart. The objective
was to learn a controller to swing the pendulum up from around
$\vec\mu_0 = [x_0,\dot x_0, \theta_0,\dot\theta_0]\T = [0,0,0,0]\T$
and to balance it in the inverted position in the middle of the track,
i.e., around $\vec x_{\text{target}} = [0, *, \pi, *]\T$. Since a
linear controller is not capable of solving the task~\cite{Raiko2009},
\textsc{pilco} learned a nonlinear state-feedback controller based on
a deterministic GP with 50 basis functions (see Sec.~\ref{sec:detGP}),
resulting in 305 policy parameters to be learned.

In our simulation, we set the masses of the cart and the pendulum to
$\unit[0.5]{kg}$ each, the length of the pendulum to $\unit[0.5]{m}$,
and the coefficient of friction between cart and ground to
$\unit[0.1]{Ns/m}$. The prediction horizon was set to
$\unit[2.5]{s}$. The control signal could be changed every
$\unit[100]{ms}$. In the meantime, it was constant (zero-order-hold
control).  
The only knowledge employed about the system was the length of the
pendulum to find appropriate orders of magnitude to set the sampling
frequency ($\unit[10]{Hz}$) and the standard deviation of the cost
function ($\sigma_c=\unit[0.25]{m}$), requiring the tip of the
pendulum to move above horizontal not to incur full cost.

\subsubsection{Approximate Inference Assessment}
In the following, we evaluate the quality of the presented approximate
inference methods for policy evaluation (moment matching as described
in \sec~\ref{sec:moment matching}) and linearization of the posterior
GP mean as described in \sec~\ref{sec:linearization}) with respect to
computational demand (\sec~\ref{sec:computational demand}) and
learning speed~(\sec~\ref{sec:learning speed}).

\paragraph{\bf Computational Demand}
\label{sec:computational demand}
For a single time step, the computational complexity of \emph{moment
  matching} is $\mathcal O(n^2E^2D)$, where $n$ is the number of GP
training points, $D$ is the input dimensionality, and $E$ the
dimension of the prediction. The most expensive computations are the
entries of $\mat Q\in\R^{n\times n}$, which are given in
\eq\eqref{eq:Q-matrix entries}. Each entry $Q_{ij}$ requires
evaluating a kernel, which is essentially a $D$-dimensional scalar
product. The values $\vec z_{ij}$ are cheap to compute and $\mat R$
needs to be computed only once. We end up with $\mathcal O(n^2E^2D)$
since $\mat Q$ needs to be computed for all entries of the $E\times E$
predictive covariance matrix.

For a single time step, the computational complexity of
\emph{linearizing the posterior GP mean function} is $\mathcal
O(n^2DE)$. The most expensive operation is the determination of
$\mat\Sigma_w$ in \eq\eqref{eq:EKF pred covariance}, i.e., the model
uncertainty at the mean of the input distribution, which scales in
$\mathcal O(n^2 D)$. This computation is performed for all $E$
predictive dimensions, resulting in a computational complexity of
$\mathcal O(n^2DE)$.

%
\begin{figure}
  \centering \subfigure[Linearizing the mean function.]{
\includegraphics[width=\twofig]{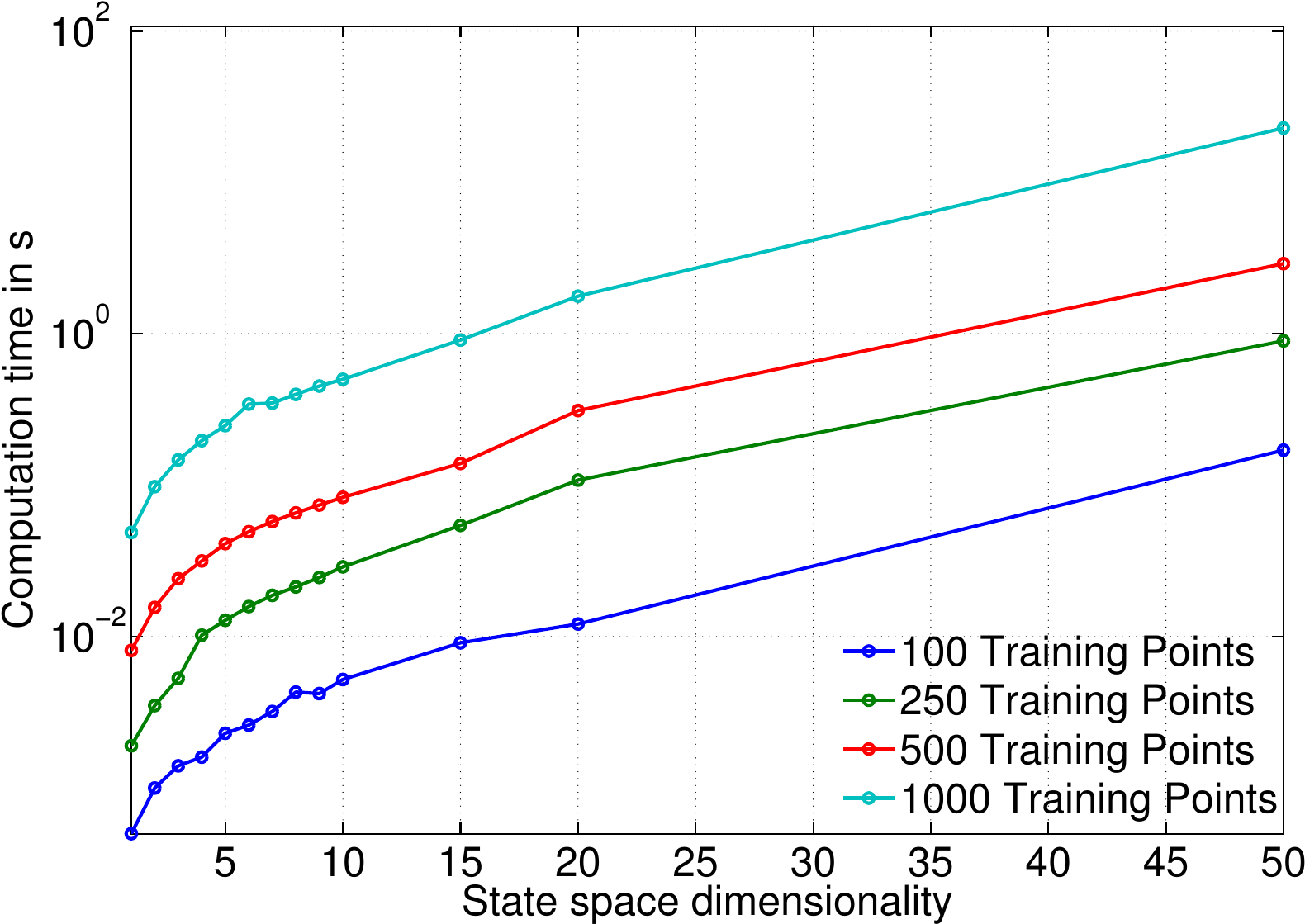}
\label{fig:dim vs comptime lin}
}
\hfill
\subfigure[Moment matching.]{
\includegraphics[width=\twofig]{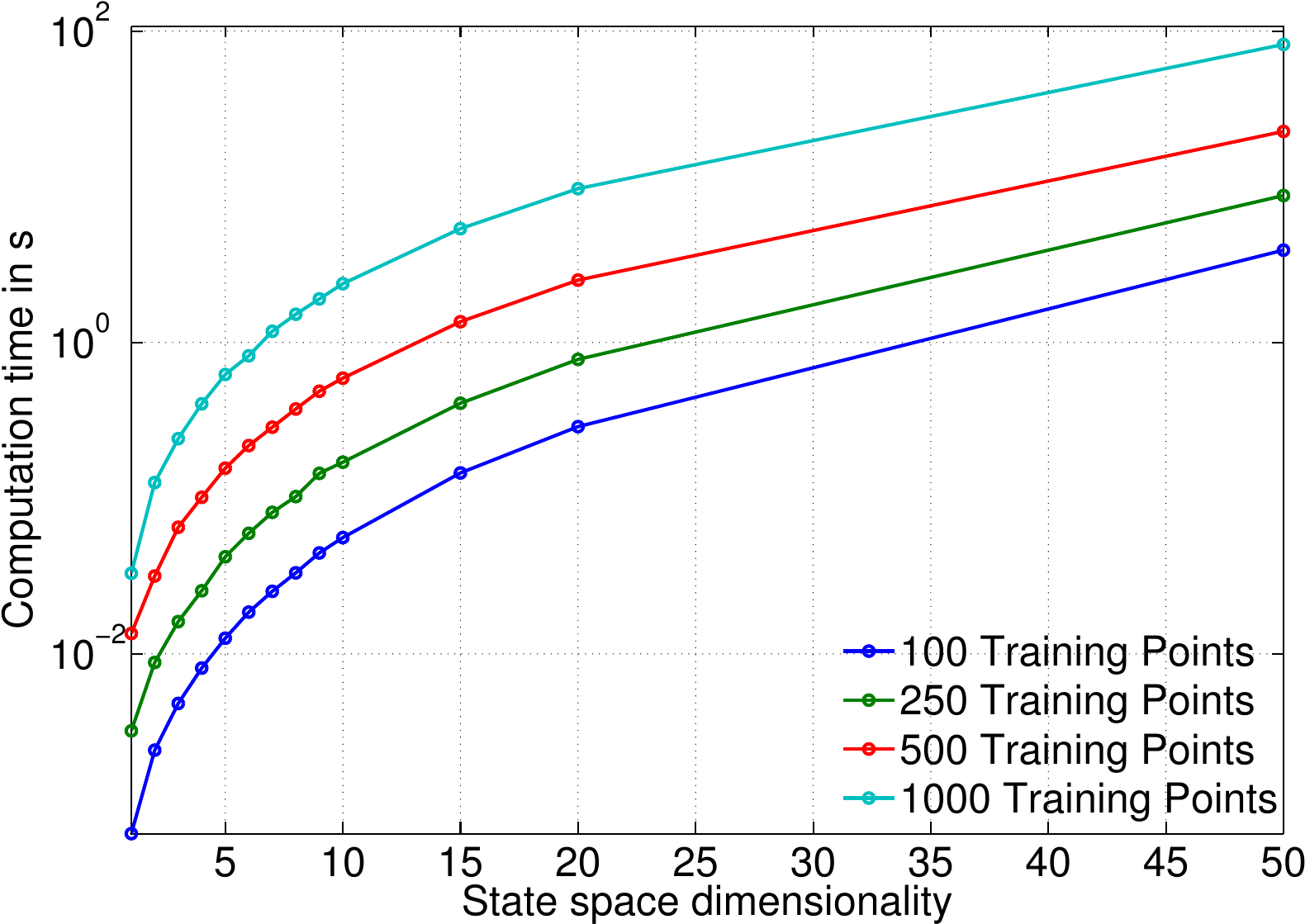}
\label{fig:dim vs comptime MM}
}
\caption{Empirical computational demand for approximate inference and
  derivative computation with GPs for a single time step, shown on a
  log scale. \subref{fig:dim vs comptime lin}: Linearization of the
  posterior GP mean. \subref{fig:dim vs comptime MM}: Exact moment
  matching.}
\label{fig:dim vs comptime}
\figspace
\end{figure}
Fig.~\ref{fig:dim vs comptime} illustrates the empirical computational
effort for both linearization of the posterior GP mean and exact
moment matching. We randomly generated GP models in $D=1, 2, 3, 4, 5,
6, 7, 8, 9, 10, 15, 20, 50$ dimensions and GP training set sizes of
$n=100, 250, 500, 1000$ data points. We set the predictive dimension
$E=D$.
The CPU time (single core) for computing a predictive state
distribution and the required derivatives are shown as a function of
the dimensionality of the state. Four graphs are shown for set-ups
with 100, 250, 500, and 1000 GP training points,
respectively. \Fig~\ref{fig:dim vs comptime lin} shows the graphs for
approximate inference based on linearization of the posterior GP mean,
and \fig~\ref{fig:dim vs comptime MM} shows the corresponding graphs
for exact moment matching on a logarithmic scale. Computations based
on linearization were consistently faster by a factor of 5--10.

\paragraph{\bf Learning Speed}
\label{sec:learning speed}
For eight different random initial trajectories and controller
initializations, \textsc{pilco} followed \alg~\ref{alg:pilco} to learn
policies. In the cart-pole swing-up task, \textsc{pilco} learned for
15 episodes, which corresponds to a total of $\unit[37.5]{s}$ of
data. In the double-pendulum swing-up task, \textsc{pilco} learned for
30 episodes, corresponding to a total of $\unit[75]{s}$ of data.
To evaluate the learning progress, we applied the learned controllers
after each policy search (see line~\ref{alg:pilco:apply} in
\alg~\ref{alg:pilco}) 20 times for $\unit[2.5]{s}$, starting from 20
different initial states $\vec x_0\sim\prob(\vec x_0)$. The learned
controller was considered successful when the tip of the pendulum was
close to the target location from $\unit[2]{s}$ to $\unit[2.5]{s}$,
i.e., at the end of the rollout.

\begin{itemize}
\item 
\textbf{Cart-Pole Swing-Up.}
\Fig~\ref{fig:learning curve cart pole MM vs lin} shows
\textsc{pilco}'s average learning success for the cart-pole swing-up
task as a function of the total experience. We evaluated both
approximate inference methods for policy evaluation, moment matching
and linearization of the posterior GP mean
function. \Fig~\ref{fig:learning curve cart pole MM vs lin} shows that
learning using the computationally more demanding moment matching is
more reliable than using the computationally more advantageous
linearization. On average, after $\unit[15]{s}$--$\unit[20]{s}$ of
experience, \textsc{pilco} reliably, i.e., in $\approx 95\%$ of the
test runs, solved the cart-pole swing-up task, whereas the
linearization resulted in a success rate of about $83\%$.

\Fig~\ref{fig:cp comparison} relates \textsc{pilco}'s learning speed
(blue bar) to other RL methods (black bars), which solved the
cart-pole swing-up task from scratch, i.e., without human
demonstrations or known dynamics
models~\cite{Coulom2002,Kimura1999,Doya2000,Wawrzynski2004,Raiko2009}.
\begin{figure}[tb]
  \centering \subfigure[Average learning curves with 95\% standard
  errors: moment matching (MM) and posterior GP linearization (Lin).]{
\includegraphics[width =
\twofig]{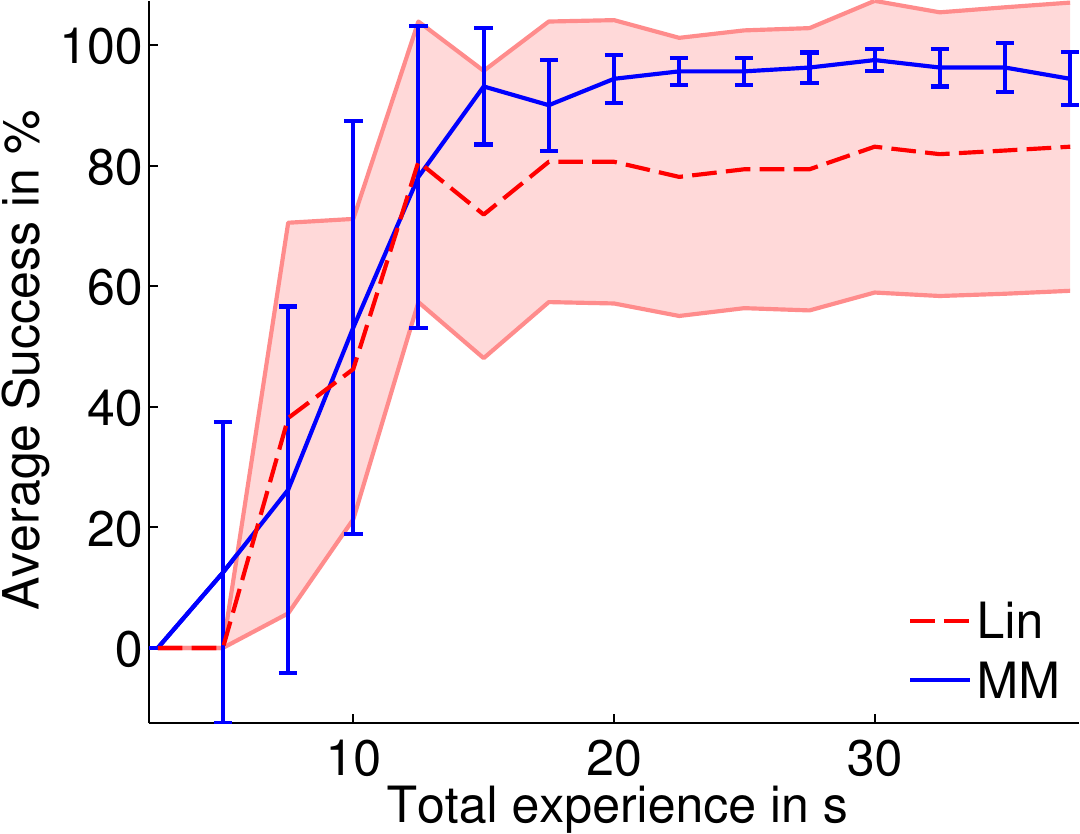}
\label{fig:learning curve cart pole MM vs lin}
}
\hfill
\subfigure[Required interaction time of different RL algorithms for
  learning the cart-pole swing-up from scratch, shown on a log
  scale.]{
\includegraphics[width=\twofig]{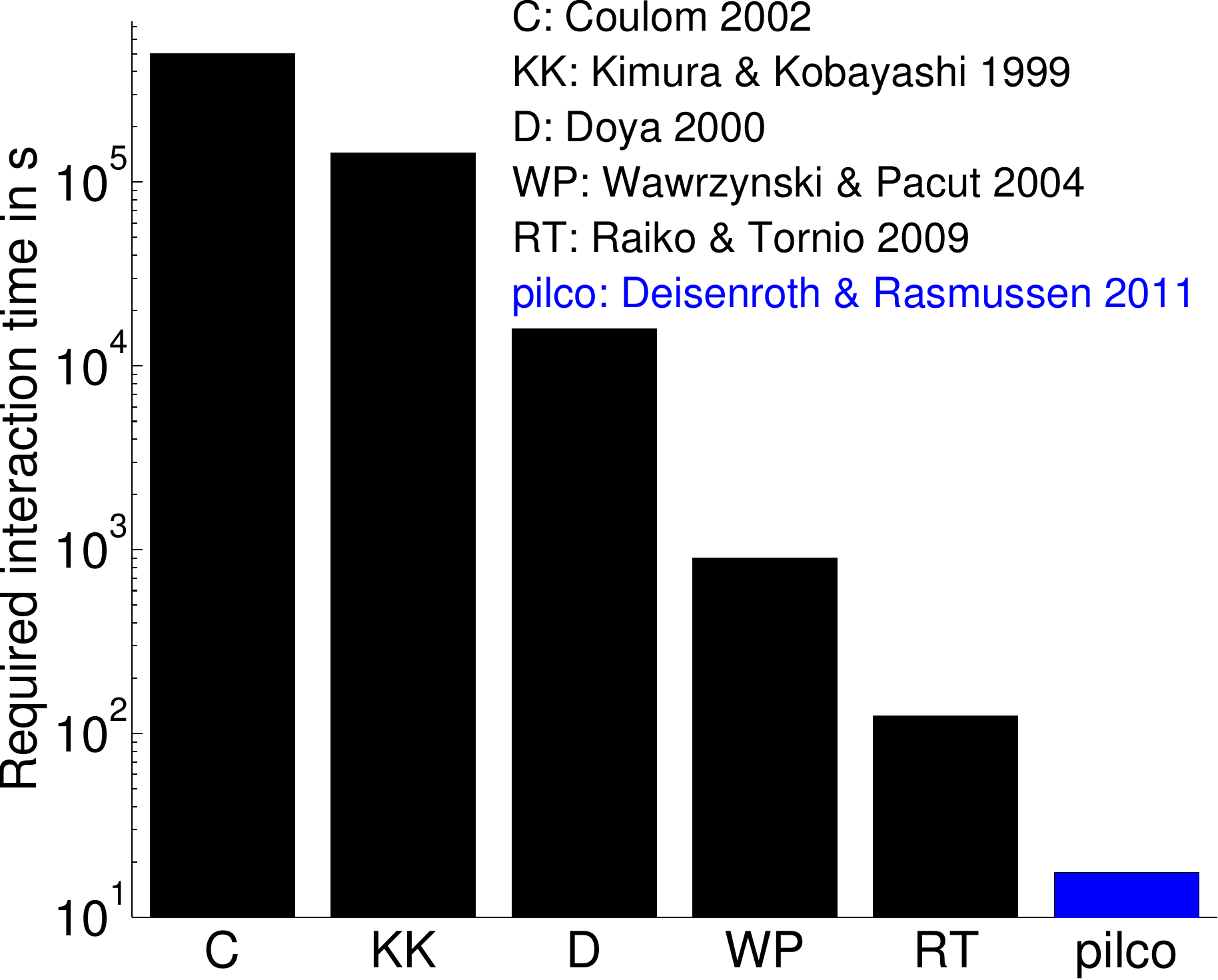}
\label{fig:cp comparison}
}
\caption{Results for the cart-pole swing-up task. \subref{fig:learning
    curve cart pole MM vs lin} Learning curves for moment matching and
linearization (simulation task), \subref{fig:cp comparison} required
interaction time for solving the cart-pole swing-up task compared with
other algorithms.}
\label{fig:cp learning speeds}
\vspace{-2mm}
\end{figure}
\nocite{Kimura1999,Doya2000,Coulom2002,Wawrzynski2004,Raiko2009}
Dynamics models were only learned in~\cite{Doya2000,Raiko2009}, using
RBF networks and multi-layered perceptrons, respectively. In all cases
without state-space discretization, cost functions similar to ours
(see \eq\eqref{eq:immediate cost}) were used. \Fig~\ref{fig:cp
  comparison} stresses \textsc{pilco}'s data efficiency:
\textsc{Pilco} outperforms any other currently existing RL algorithm
by at least one order of magnitude.
\item
\textbf{Double-Pendulum Swing-Up with Two Actuators.}
\begin{figure}
\centering
\includegraphics[height = 4cm]{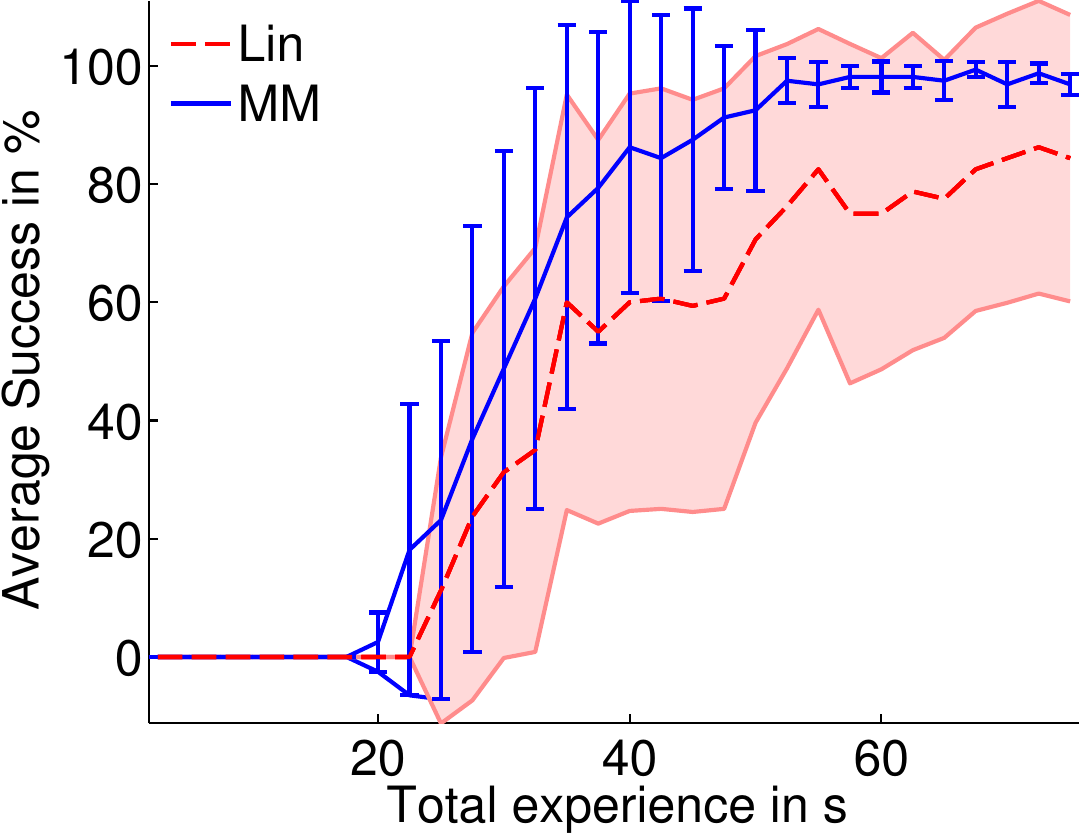}
\caption{Average success as a function of the total data used for
  learning (double pendulum swing-up). The blue error bars show the
  95\% confidence bounds of the standard error for the moment matching
  (MM) approximation, the red area represents the corresponding
  confidence bounds of success when using approximate inference by
  means of linearizing the posterior GP mean (Lin).}
\label{fig:double pendulum learning curves fullLin}
\figspace
\end{figure}
\Fig~\ref{fig:double pendulum learning curves fullLin} shows the
learning curves for the double-pendulum swing-up task when using
either moment matching or mean function linearization for approximate
inference during policy evaluation. \Fig~\ref{fig:double pendulum
  learning curves fullLin} shows that \textsc{pilco} learns faster
(learning already kicks in after $\unit[20]{s}$ of data) and overall
more successfully with moment matching. Policy evaluation based on
linearization of the posterior GP mean function achieved about $80\%$
success on average, whereas moment matching on average solved the task
reliably after about $\unit[50]{s}$ of data with a success rate
$\approx 95\%$.
\end{itemize}
\textbf{Summary.}  We have seen that both approximate inference
methods have pros and cons: Moment matching requires more
computational resources than linearization, but learns faster and more
reliably. The reason why linearization did not reliably succeed in
learning the tasks is that it gets relatively easily stuck in local
minima, which is largely a result of underestimating predictive
variances, an example of which is given in \fig~\ref{fig:EKF vs
  ADF}. Propagating too confident predictions over a longer horizon
often worsens the problem.  Hence, in the following, we focus solely
on the moment matching approximation.

\subsubsection{Quality of the Gaussian Approximation}
\label{sec:quality Gauss}
\textsc{Pilco} strongly relies on the quality of approximate
inference, which is used for long-term predictions and policy
evaluation, see \sec~\ref{sec:approximate inference}. We already saw
differences between linearization and moment matching; however, both
methods approximate predictive distributions by a Gaussian. Although
we ultimately cannot answer whether this approximation is good under
all circumstances, we will shed some light on this issue.

\begin{figure}
\centering
\subfigure[Early stage of learning.]{
\includegraphics[width =
\twofig]{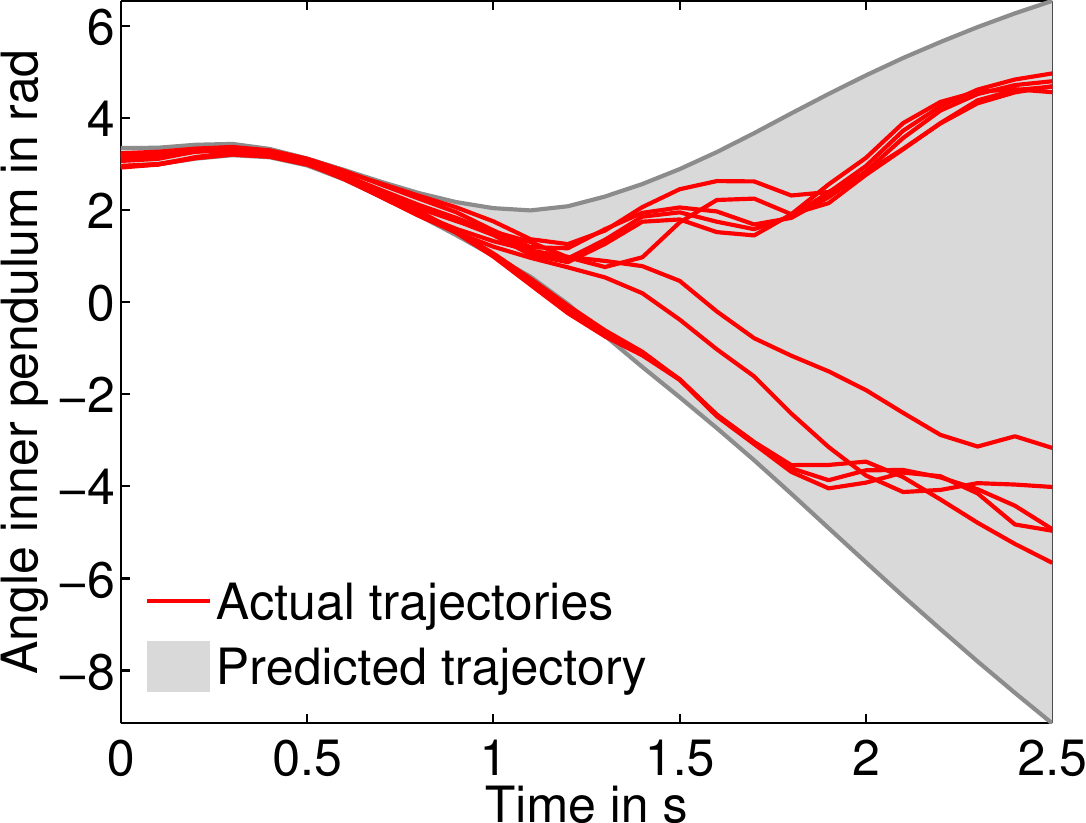}
\label{fig:quality Gaussian early}
}
\hfill
\subfigure[After successful learning.]{
\includegraphics[width =
\twofig]{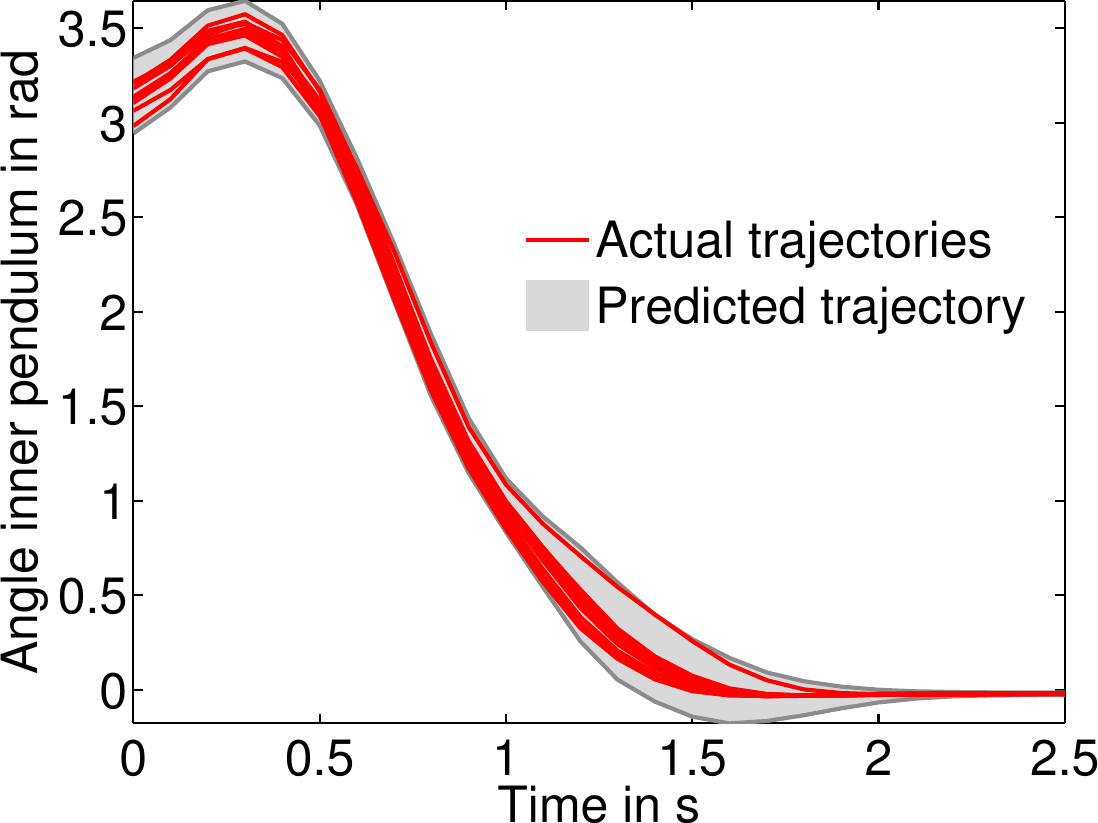}
\label{fig:quality Gaussian success}
}
\caption{Long-term predictive (Gaussian) distributions during planning
  (shaded) and sample rollouts (red). \subref{fig:quality Gaussian
    early} In the early stages of learning, the Gaussian approximation
  is a \new{suboptimal} choice. \subref{fig:quality Gaussian success}
  \textsc{Pilco} learned a controller such that the Gaussian
  approximations of the predictive states are good. Note the different
  scales in \subref{fig:quality Gaussian early} and
  \subref{fig:quality Gaussian success}.}
\label{fig:quality Gaussian approximation}
\figspace
\end{figure}
\Fig~\ref{fig:quality Gaussian approximation} shows a typical example
of the angle of the inner pendulum of the double pendulum system
where, in the early stages of learning, the Gaussian approximation to
the multi-step ahead predictive distribution is not ideal. The
trajectory distribution of a set of rollouts (red) is
multimodal. \textsc{Pilco} deals with this inappropriate modeling by
learning a controller that forces the actual trajectories into a
unimodal distribution such that a Gaussian approximation is
appropriate, \fig~\ref{fig:quality Gaussian success}.

We explain this behavior as follows: Assuming that \textsc{pilco}
found different paths that lead to a target, a wide Gaussian
distribution is required to capture the variability of the bimodal
distribution. However, when computing the expected cost using a
quadratic or saturating cost, for example, uncertainty in the
predicted state leads to higher expected cost, assuming that the mean
is close to the target. Therefore, \textsc{pilco} uses its ability to
choose control policies to push the marginally multimodal trajectory
distribution into a single mode---from the perspective of minimizing
expected cost with limited expressive power, this approach is
desirable. Effectively, learning good controllers and models goes hand
in hand with good Gaussian approximations.

\subsubsection{Importance of Bayesian Averaging}
Model-based RL greatly profits from the flexibility of nonparametric
models as motivated in \sec~\ref{sec:relwork}. In the following, we
have a closer look at whether Bayesian models are strictly necessary
as well.  In particular, we evaluated whether Bayesian averaging is
necessary for successfully learning from scratch. To do so, we
considered the cart-pole swing-up task with two different dynamics
models: first, the standard nonparametric Bayesian GP model, second, a
nonparametric deterministic GP model, i.e., a GP where we considered
only the posterior mean, but discarded the posterior model uncertainty
when doing long-term predictions. We already described a similar kind
of function representation to learn a deterministic policy, see
\sec~\ref{sec:detGP}. The difference to the policy is that in this
section the deterministic GP is still nonparametric (new basis
functions are added if we get more data), whereas the number of basis
functions in the policy is fixed. However, the deterministic GP is no
longer probabilistic because of the loss of model uncertainty, which
also results in a degenerate model. Note that we still propagate
uncertainties resulting from the initial state distribution $p(\vec
x_0)$ forward.

\begin{table}[tb]
\caption{Average learning success with learned nonparametric (NP)
  transition models (cart-pole swing-up).}
\label{tab:success rates}
\centering
\begin{tabular}{c|cc}
& Bayesian NP model & Deterministic NP model\\
\hline
Learning success & \textbf{94.52\%} & 0\%
\end{tabular}
\end{table}
Tab.~\ref{tab:success rates} shows the average learning success of
swinging the pendulum up and balancing it in the inverted position in
the middle of the track.  We used moment matching for approximate
inference, see \sec~\ref{sec:approximate
  inference}. Tab.~\ref{tab:success rates} shows that learning is only
successful when model uncertainties are taken into account during
long-term planning and control learning, which strongly suggests
Bayesian nonparametric models in model-based RL. 

The reason why model uncertainties must be appropriately taken into
account is the following: In the early stages of learning, the learned
dynamics model is based on a relatively small data set. States close
to the target are unlikely to be observed when applying random
controls. Therefore, the model must extrapolate from the current set
of observed states. This requires to predict function values in
regions with large posterior model uncertainty. Depending on the
choice of the deterministic function (we chose the MAP estimate), the
predictions (point estimates) are very different. Iteratively
predicting state distributions ends up in predicting trajectories,
which are essentially arbitrary and not close to the target state
either, resulting in vanishing policy gradients.

\subsection{Scaling to Higher Dimensions: Unicycling}
\begin{figure}[tb]
\centering
\subfigure[Robotic unicycle.]{
\includegraphics[height = 3.7cm]{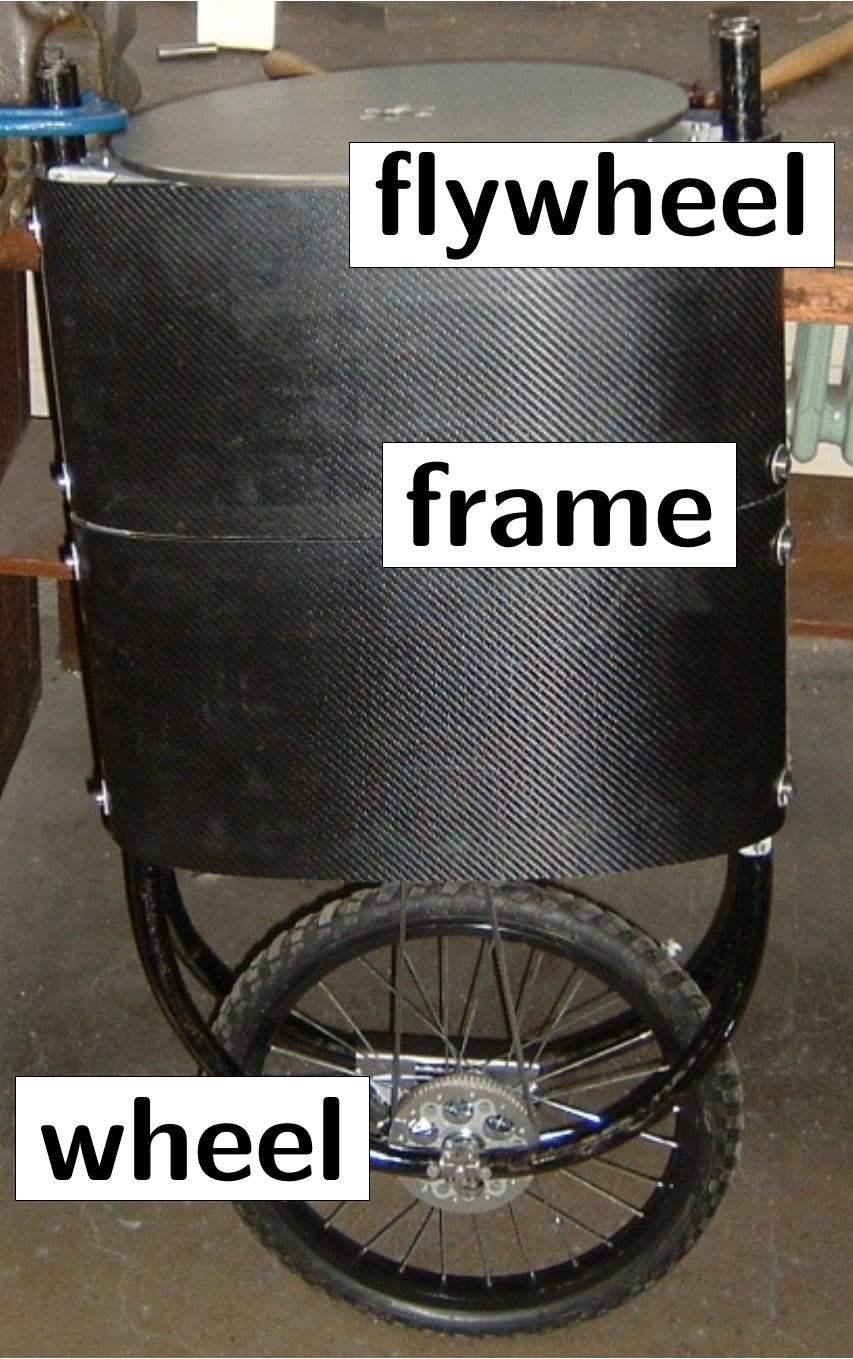}
\label{fig:unicycle}
}
\hfill
\subfigure[Histogram (after 1,000 test runs) of the distances of the
flywheel from being upright.]{
\includegraphics[height = 3.7cm]{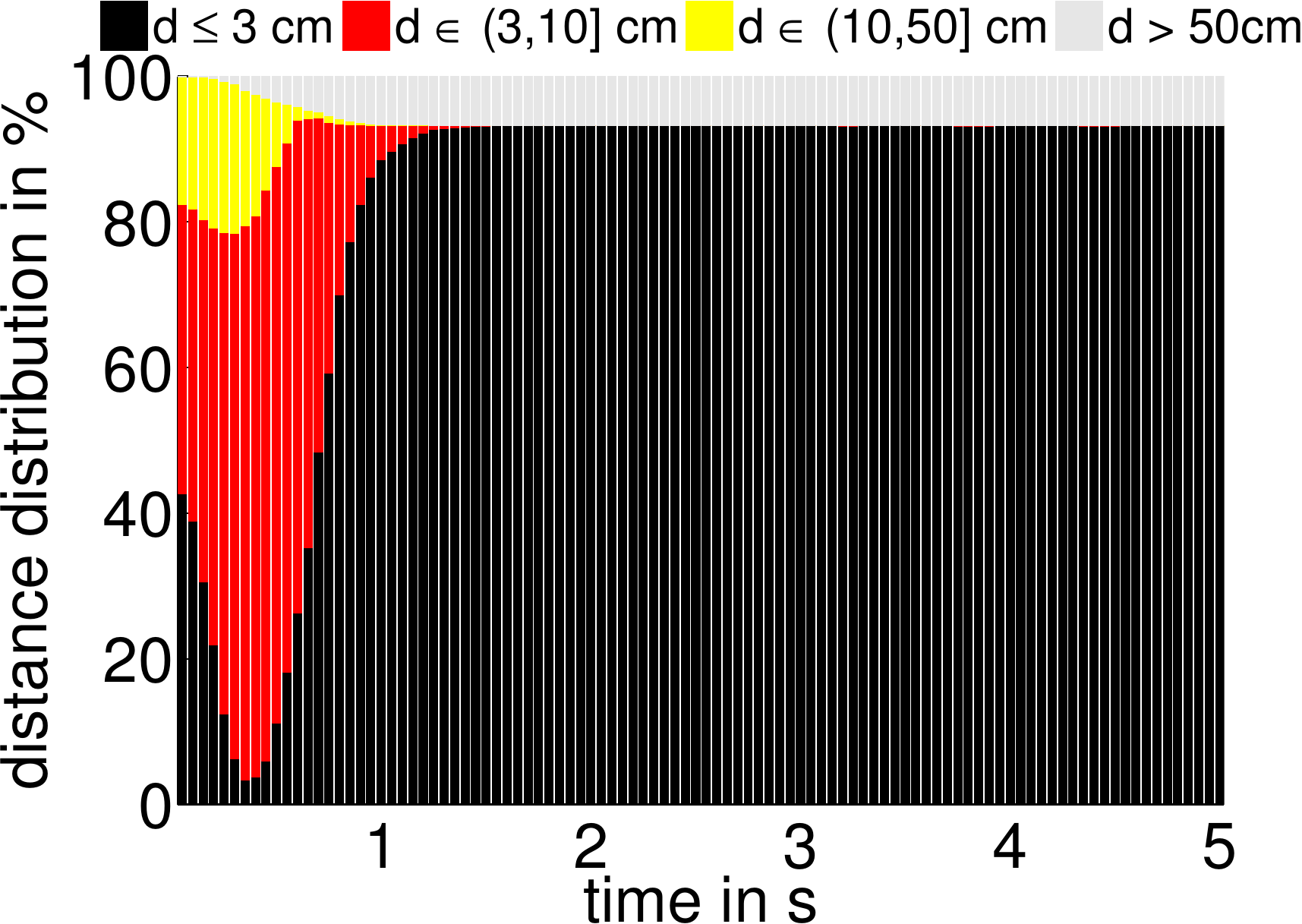}
\label{fig:costHist}
}
\caption{Robotic unicycle system and simulation results. The state
  space is $\R^{12}$, the control space $\R^2$.}
\figspace
\end{figure}

We applied \textsc{pilco} to learning to ride a 5-DoF unicycle with
$\vec x\in\R^{12}$ and $\vec u\in\R^2$ in a realistic simulation of
the one shown in \fig~\ref{fig:unicycle}.
%
The unicycle was $\unit[0.76]{m}$ high and consisted of a
$\unit[1]{kg}$ wheel, a $\unit[23.5]{kg}$ frame, and a $\unit[10]{kg}$
flywheel mounted perpendicularly to the frame. Two torques could be
applied to the unicycle: The first torque $|u_w|\leq\unit[10]{Nm}$ was
applied directly on the wheel to mimic a human rider using pedals. The
torque produced longitudinal and tilt accelerations. Lateral stability
of the wheel could be maintained by steering the wheel toward the
falling direction of the unicycle and by applying a torque
$|u_t|\leq\unit[50]{Nm}$ to the flywheel. The dynamics of the robotic
unicycle were described by 12 coupled first-order ODEs,
see~\cite{Forster2009}.

The objective was to learn a controller for riding the unicycle, i.e.,
to prevent it from falling.  To solve the balancing task, we used the
linear preliminary policy $\tilde\pi(\vec x,\vec\polpar) = \mat A\vec
x +\vec b$ with $\vec\polpar = \{\mat A,\vec b\}\in\R^{28}$. The
covariance $\mat\Sigma_0$ of the initial state was $0.25^2\mat I$
allowing each angle to be off by about $30^\circ$ (twice the standard
deviation).

\textsc{Pilco} differs from conventional controllers in that it learns
a single controller for all control dimensions \emph{jointly}. Thus,
\textsc{pilco} takes the correlation of all control and state
dimensions into account during planning and control. Learning separate
controllers for each control variable is often
unsuccessful~\cite{Naveh1999}.

\textsc{Pilco} required about 20 trials, corresponding to an overall
experience of about $\unit[30]{s}$, to learn a dynamics model and a
controller that keeps the unicycle upright. A trial was aborted when
the turntable hit the ground, which happened quickly during the five
random trials used for initialization. \Fig~\ref{fig:costHist} shows
empirical results after 1,000 test runs with the learned policy:
Differently-colored bars show the distance of the flywheel from a
fully upright position. Depending on the initial configuration of the
angles, the unicycle had a transient phase of about a second. After
$\unit[1.2]{s}$, either the unicycle had fallen or the learned
controller had managed to balance it very closely to the desired
upright position. The success rate was approximately $93\%$; bringing
the unicycle upright from extreme initial configurations was sometimes
impossible due to the torque constraints.

\subsection{Hardware Tasks}
In the following, we present results from~\cite{Deisenroth2011c,
  Deisenroth2011b}, where we successfully applied the \textsc{pilco}
policy search framework to challenging control and robotics tasks,
respectively. It is important to mention that no task-specific
modifications were necessary, besides choosing a controller
representation and defining an immediate cost function. In particular,
we used the same standard GP priors for learning the forward dynamics
models.

\subsubsection{Cart-Pole Swing-Up}
\begin{figure*}[tb]
\centering
\includegraphics[width=\sixfig]{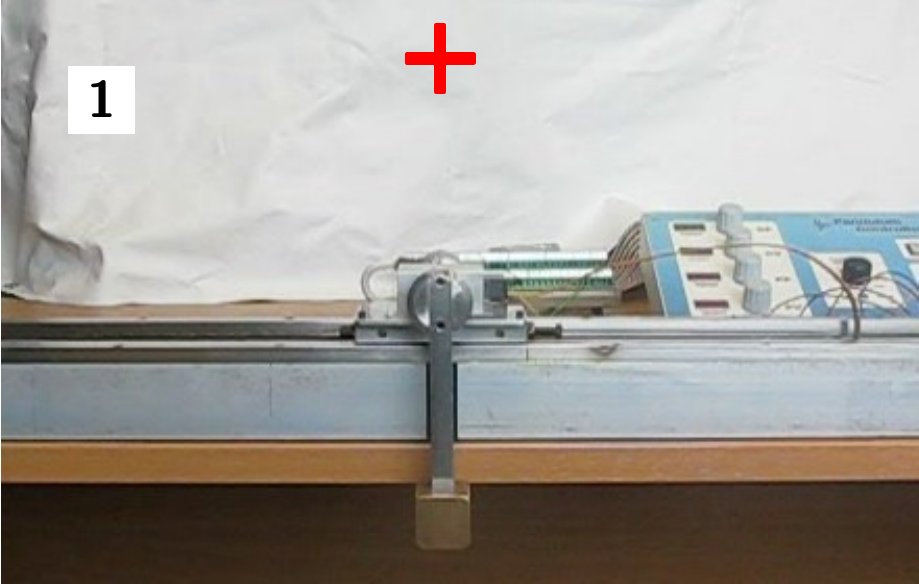}
\includegraphics[width=\sixfig]{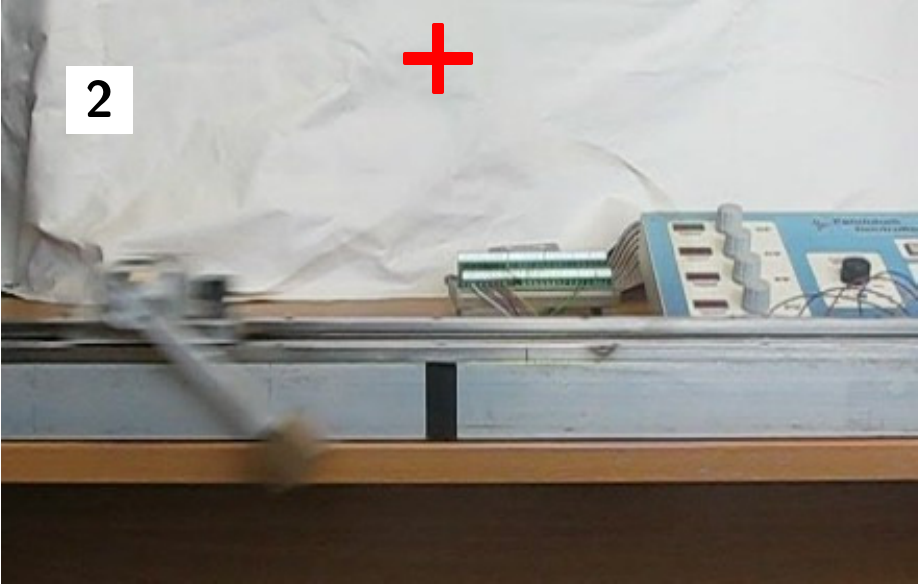}
\includegraphics[width=\sixfig]{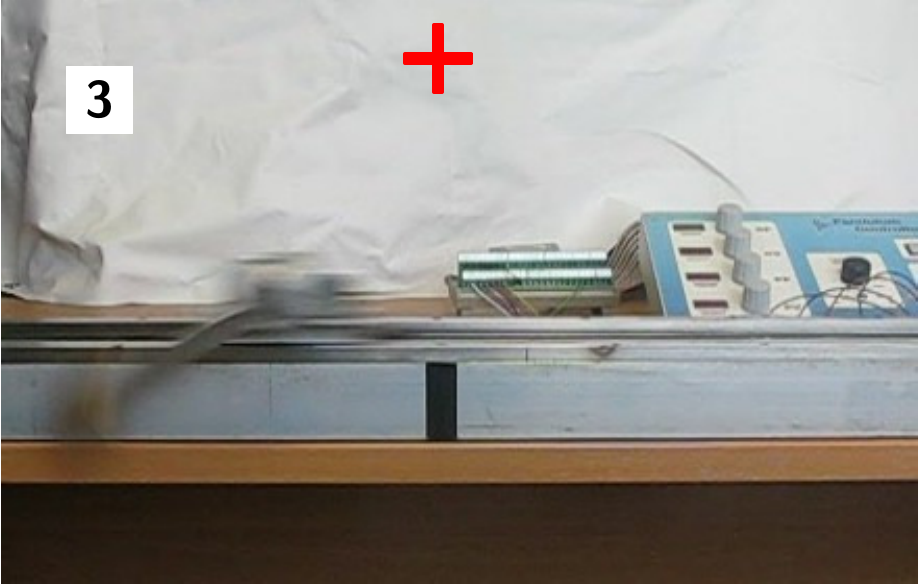}
\includegraphics[width=\sixfig]{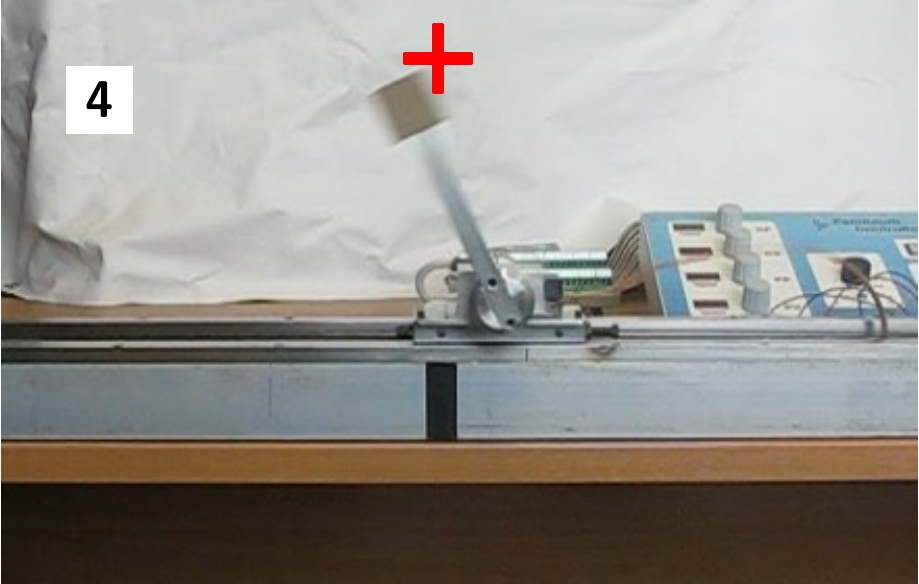}
\includegraphics[width=\sixfig]{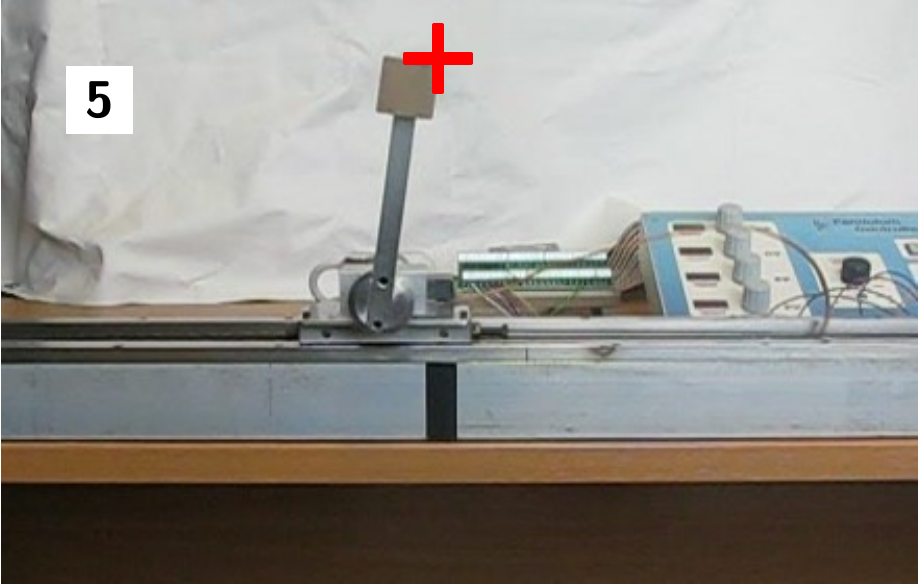}
\includegraphics[width=\sixfig]{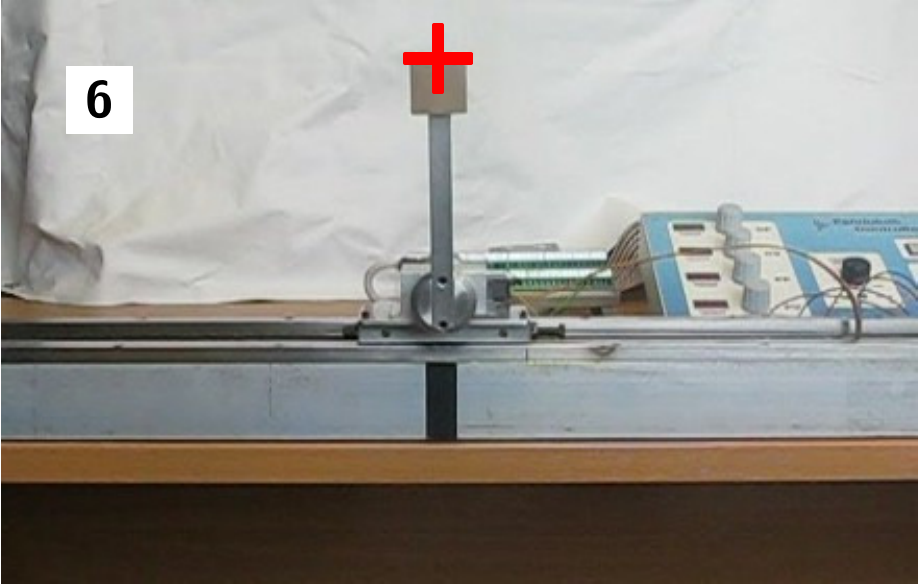}
\caption{Real cart-pole system~\cite{Deisenroth2011c}. Snapshots of a
  controlled trajectory of $\unit[20]{s}$ length after having learned
  the task. To solve the swing-up plus balancing, \textsc{pilco}
  required only $\unit[17.5]{s}$ of interaction with the physical
  system.}
\label{fig:cp-hw}
\figspace
\end{figure*}
As described in~\cite{Deisenroth2011c}, \textsc{pilco} was applied to
learning to control the \emph{real} cart-pole system, see
\fig~\ref{fig:cp-hw}, developed by~\cite{Jervis1992}. The masses of
the cart and pendulum were $\unit[0.7]{kg}$ and $\unit[0.325]{kg}$,
respectively.  A horizontal force $u\in[-10,10]\,\unit{N}$ could be
applied to the cart.

\textsc{Pilco} successfully learned a sufficiently good dynamics model
and a good controller fully automatically in only a handful of trials
and a total experience of $\unit[17.5]{s}$, which also confirms the
learning speed of the simulated cart-pole system in \fig~\ref{fig:cp
  comparison} despite the fact that the parameters of the system
dynamics (masses, pendulum length, friction, delays, stiction, etc.)
are different.  Snapshots of a $\unit[20]{s}$ test trajectory are
shown in \fig~\ref{fig:cp-hw}; a video of the entire learning process
is available at \mbox{\url{http://www.youtube.com/user/PilcoLearner}}.

\subsubsection{Controlling a Low-Cost Robotic Manipulator}
\label{sec:lynx results}
\begin{figure}
\vspace{-4mm}
  \centering 
\includegraphics[width = 0.6\hsize]{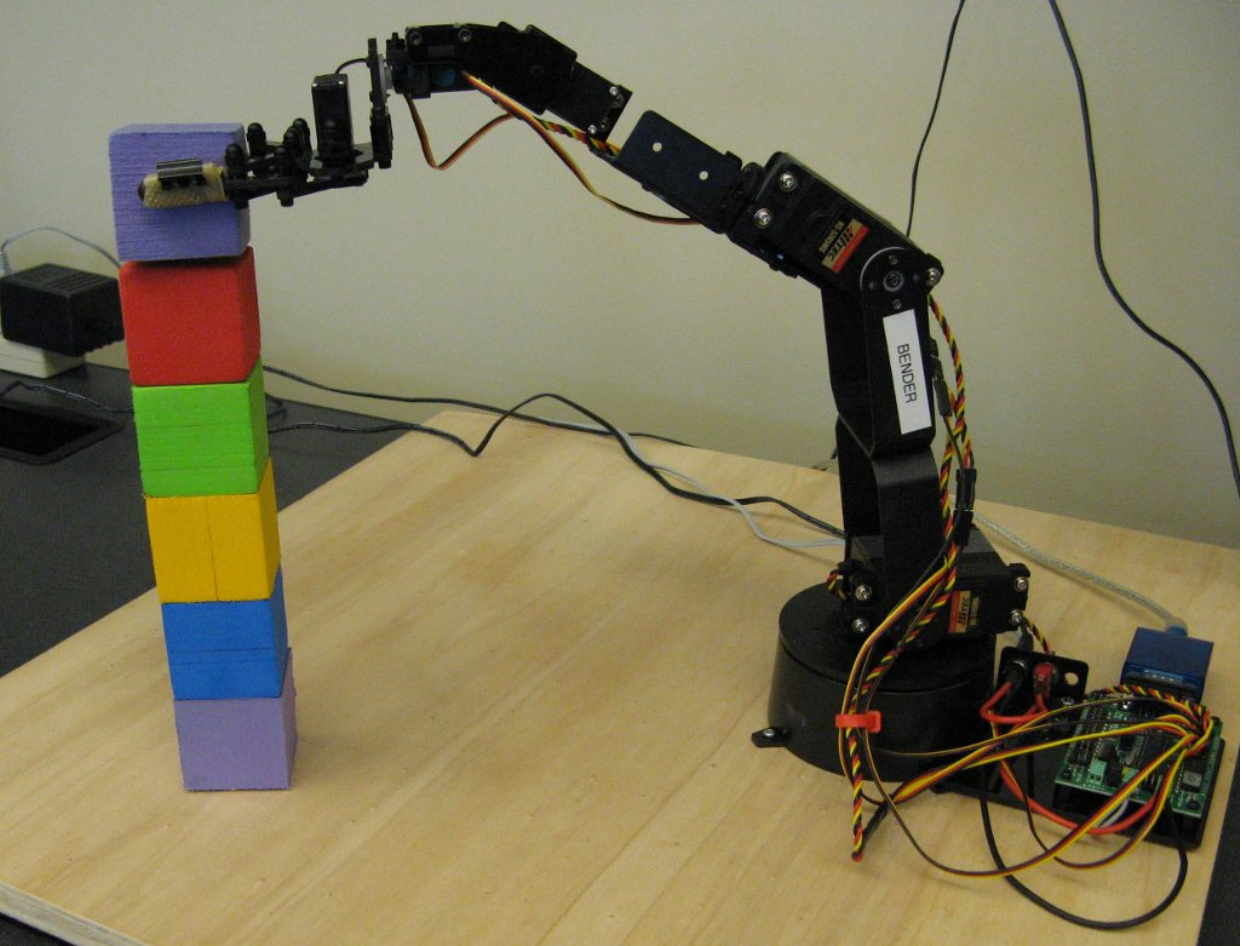}
\caption{Low-cost robotic arm by Lynxmotion~\cite{lynxmotion}. The
  manipulator does not provide any pose feedback. Hence,
  \textsc{pilco} learns a controller directly in the task space using
  visual feedback from a PrimeSense depth camera.}
\label{fig:robotic system}
\end{figure}
We applied \textsc{pilco} to make a low-precision robotic arm learn to
stack a tower of foam blocks---fully
autonomously~\cite{Deisenroth2011b}.  For this purpose, we used the
lightweight robotic manipulator by Lynxmotion \cite{lynxmotion} shown
in Fig.~\ref{fig:robotic system}. The arm costs approximately \$370
and possesses six controllable degrees of freedom: base rotate, three
joints, wrist rotate, and a gripper (open\slash close). The plastic
arm was controllable by commanding both a desired configuration of the
six servos via their pulse durations and the duration for executing
the command.  The arm was very noisy: Tapping on the base made the end
effector swing in a radius of about $\unit[2]{cm}$. The system noise
was particularly pronounced when moving the arm vertically (up\slash
down). Additionally, the servo motors had some play.

Knowledge about the joint configuration of the robot was not
available. We used a PrimeSense depth camera~\cite{primesense} as an
external sensor for visual tracking the block in the gripper of the
robot. The camera was identical to the Kinect sensor, providing a
synchronized depth image and a $640\times 480$ RGB image at
$\unit[30]{Hz}$. Using structured infrared light, the camera delivered
useful depth information of objects in a range of about
$\unit[0.5]{m}$--$\unit[5]{m}$. The depth resolution was approximately
$\unit[1]{cm}$ at a distance of $\unit[2]{m}$~\cite{primesense}.

Every $\unit[500]{ms}$, the robot used the 3D center of the block in
its gripper as the state $\vec x\in\R^3$ to compute a
continuous-valued control signal $\vec u\in\R^4$, which comprised the
commanded pulse widths for the first four servo motors. Wrist rotation
and gripper opening\slash closing were not learned.  For block
tracking we used real-time ($\unit[50]{Hz}$) color-based region
growing to estimate the extent and 3D center of the object, which was
used as the state $\vec x\in\R^3$ by \textsc{pilco}. 

As an initial state distribution, we chose $\prob(\vec
x_0)=\gaussx{\vec x_0}{\vec\mu_0}{\mat\Sigma_0}$ with $\vec\mu_0$
being a single noisy measurement of the 3D camera coordinates of the
block in the gripper when the robot was in its initial configuration.
The initial covariance $\mat\Sigma_0$ was diagonal, where the
95\%-confidence bounds were the edge length $b$ of the
block. Similarly, the target state was set based on a single noisy
measurement using the PrimeSense camera.
We used linear preliminary policies, i.e., $\tilde\pi(\vec x) = \vec u
= \mat A\vec x + \vec b$, and initialized the controller parameters
$\vec\polpar = \{\mat A,\vec b\}\in\R^{16}$ to zero.
The Euclidean distance $d$ of the end effector from the camera was
approximately $\unit[0.7]{m}$--$\unit[2.0]{m}$, depending on the
robot's configuration. 
The cost function in \eq~\eqref{eq:immediate cost} penalized the
Euclidean distance of the block in the gripper from its desired target
location on top of the current tower.
Both the frequency at which the controls were
changed and the time discretization were set to $\unit[2]{Hz}$; the
planning horizon $T$ was $\unit[5]{s}$.  After $\unit[5]{s}$, the
robot opened the gripper and released the block.

We split the task of building a tower into learning individual
controllers for each target block B2--B6 (bottom to top), see
Fig.~\ref{fig:robotic system}, starting from a configuration, in which
the robot arm was upright. 
All independently trained controllers shared the same initial trial.

The motion of the block in the end effector was modeled by GPs. The
inferred system noise standard deviations, which comprised
stochasticity of the robot arm, synchronization errors, delays, image
processing errors, etc., ranged from $\unit[0.5]{cm}$ to
$\unit[2.0]{cm}$. Here, the $y$-coordinate, which corresponded to the
height, suffered from larger noise than the other coordinates. The
reason for this is that the robot movement was particularly jerky in
the up\slash down movements.  These learned noise levels were in the
right ballpark since they were slightly larger than the expected
camera noise~\cite{primesense}. The signal-to-noise ratio in our
experiments ranged from 2 to 6.

\begin{figure*}[tb]
  \centering \subfigure[Average learning curve (block-stacking
  task). The horizontal axis shows the learning stage, the vertical
  axis the average distance to the target at the end of the episode.]{
\includegraphics[width = \fourfig]{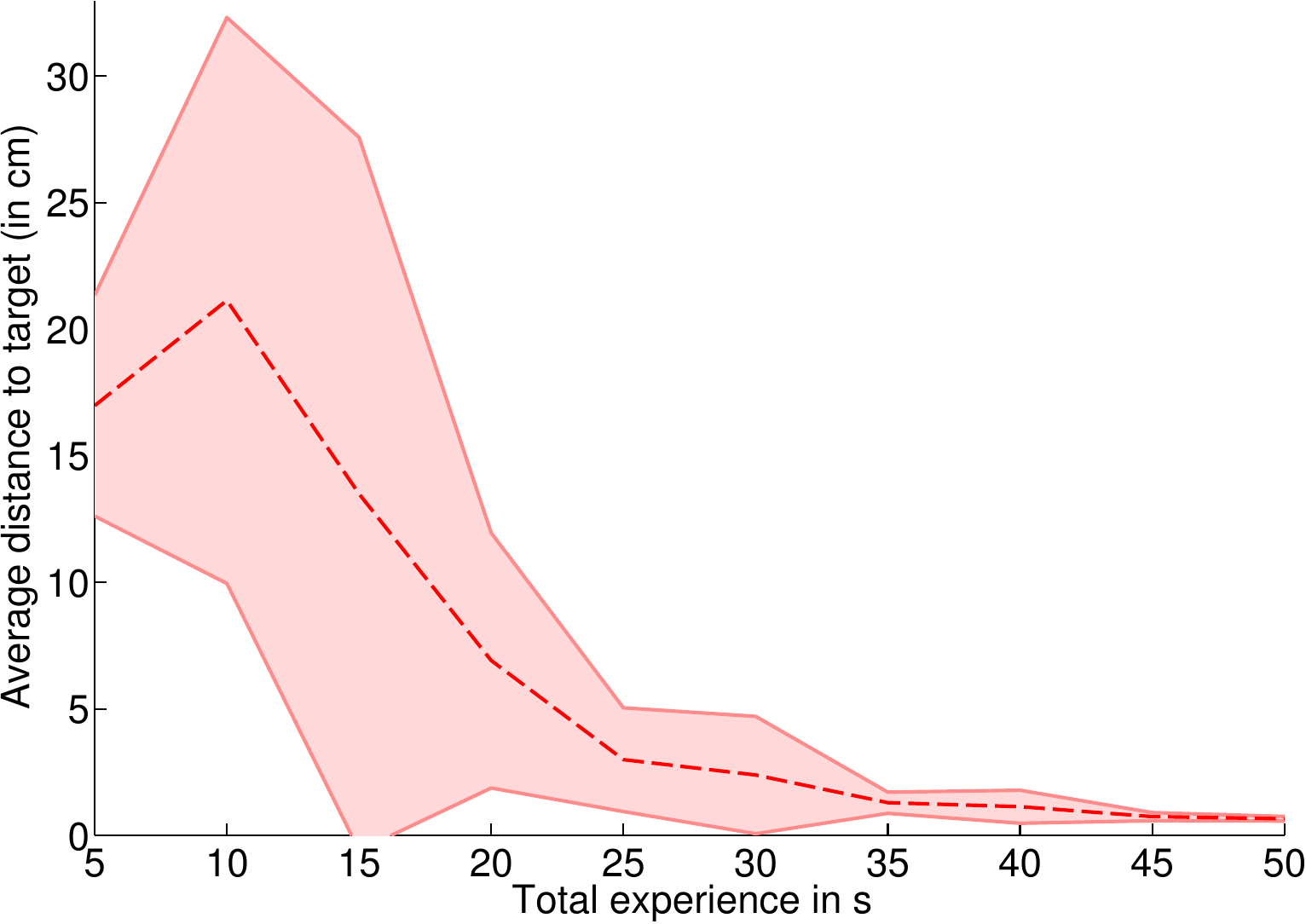}
\label{fig:learning curve}
}
\hfill
\subfigure[Marginal long-term predictive distributions and actually
  incurred trajectories. The red lines show the trajectories of the
  block in the end effector, the two dashed blue lines represent the
  95\% confidence intervals of the corresponding multi-step ahead
  predictions using moment matching. The target state is marked by the
  straight lines. All coordinates are measured in $\unit{cm}$.]{
\includegraphics[width = \fourfig]{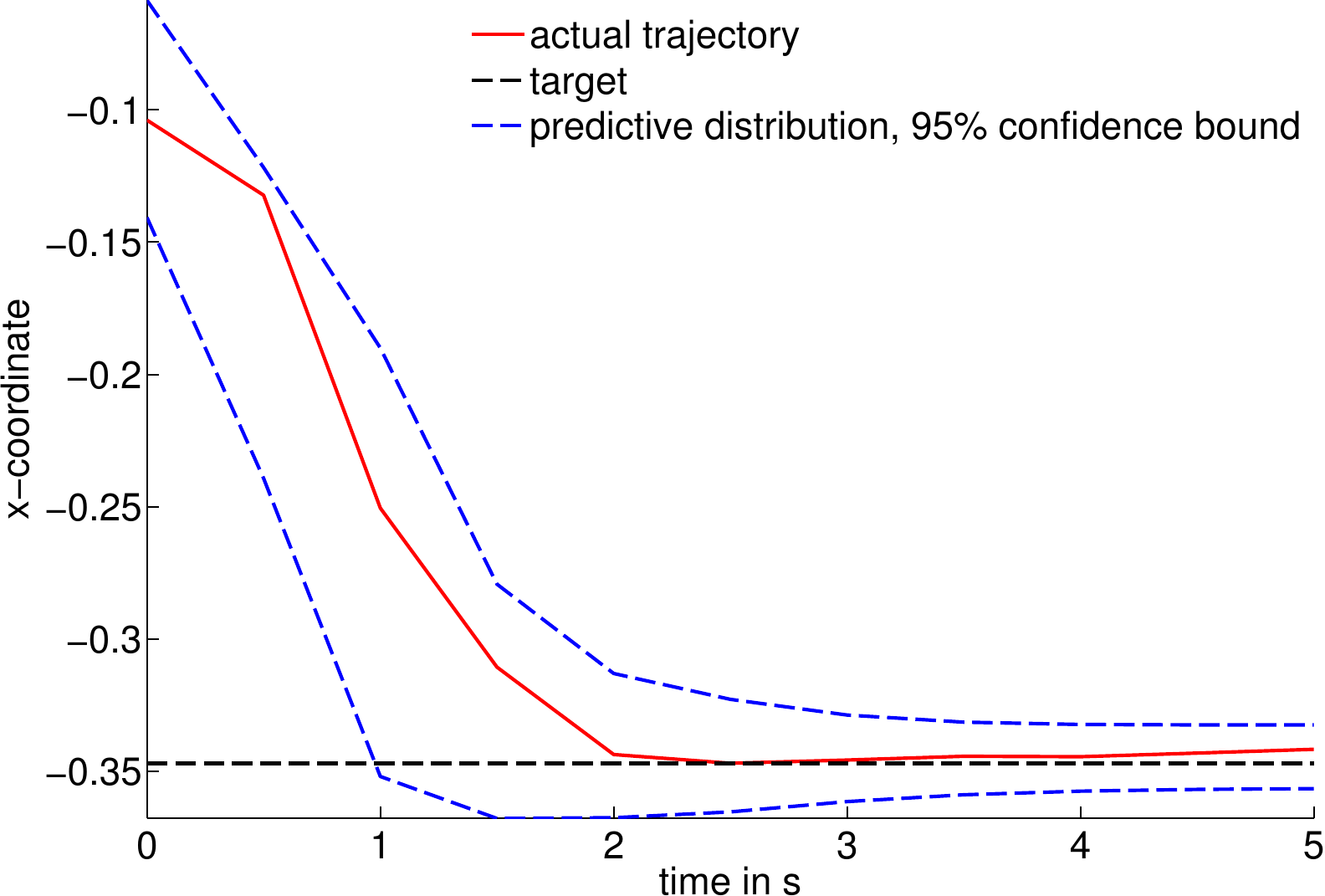}
\hfill
\includegraphics[width = \fourfig]{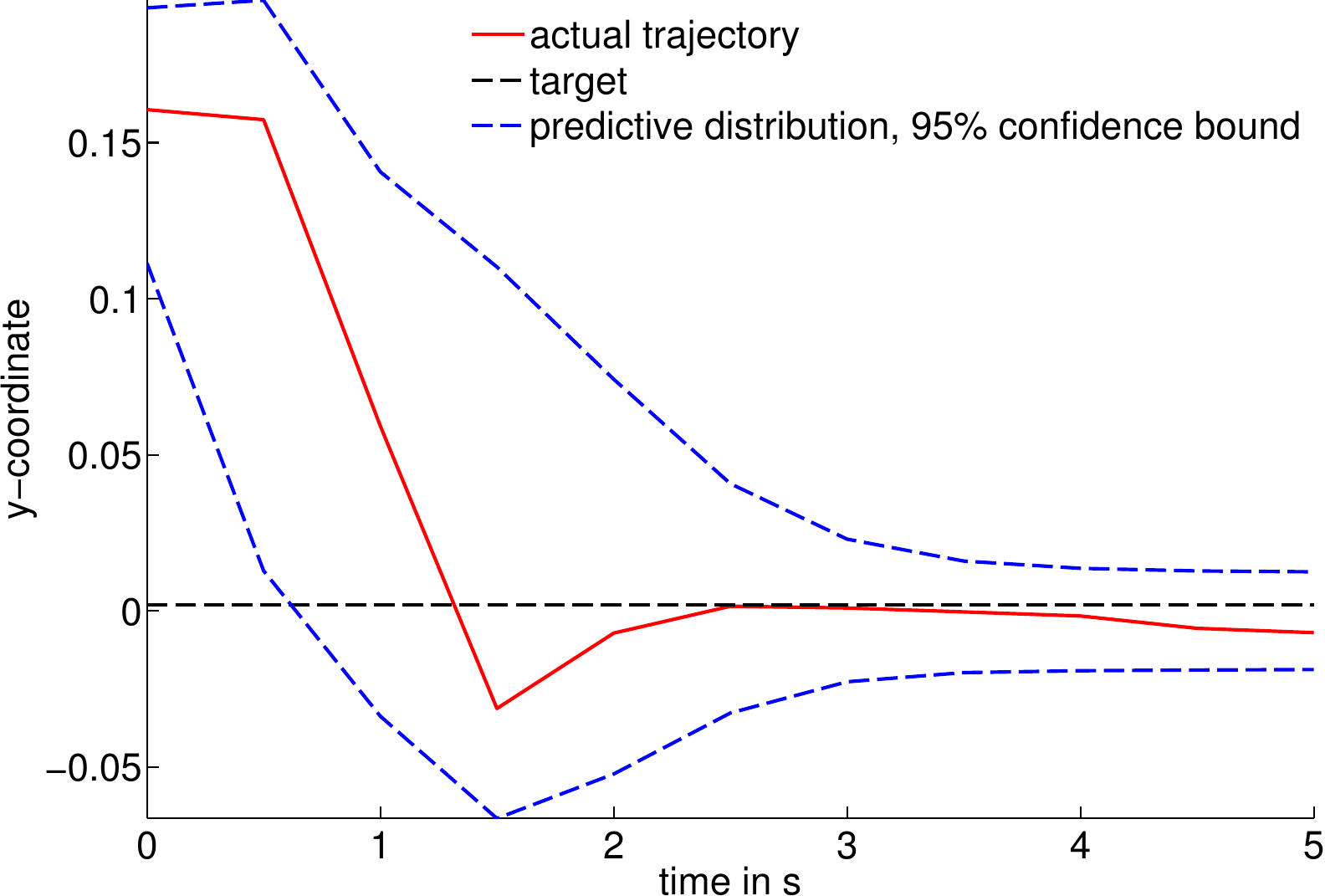}
\hfill
\includegraphics[width = \fourfig]{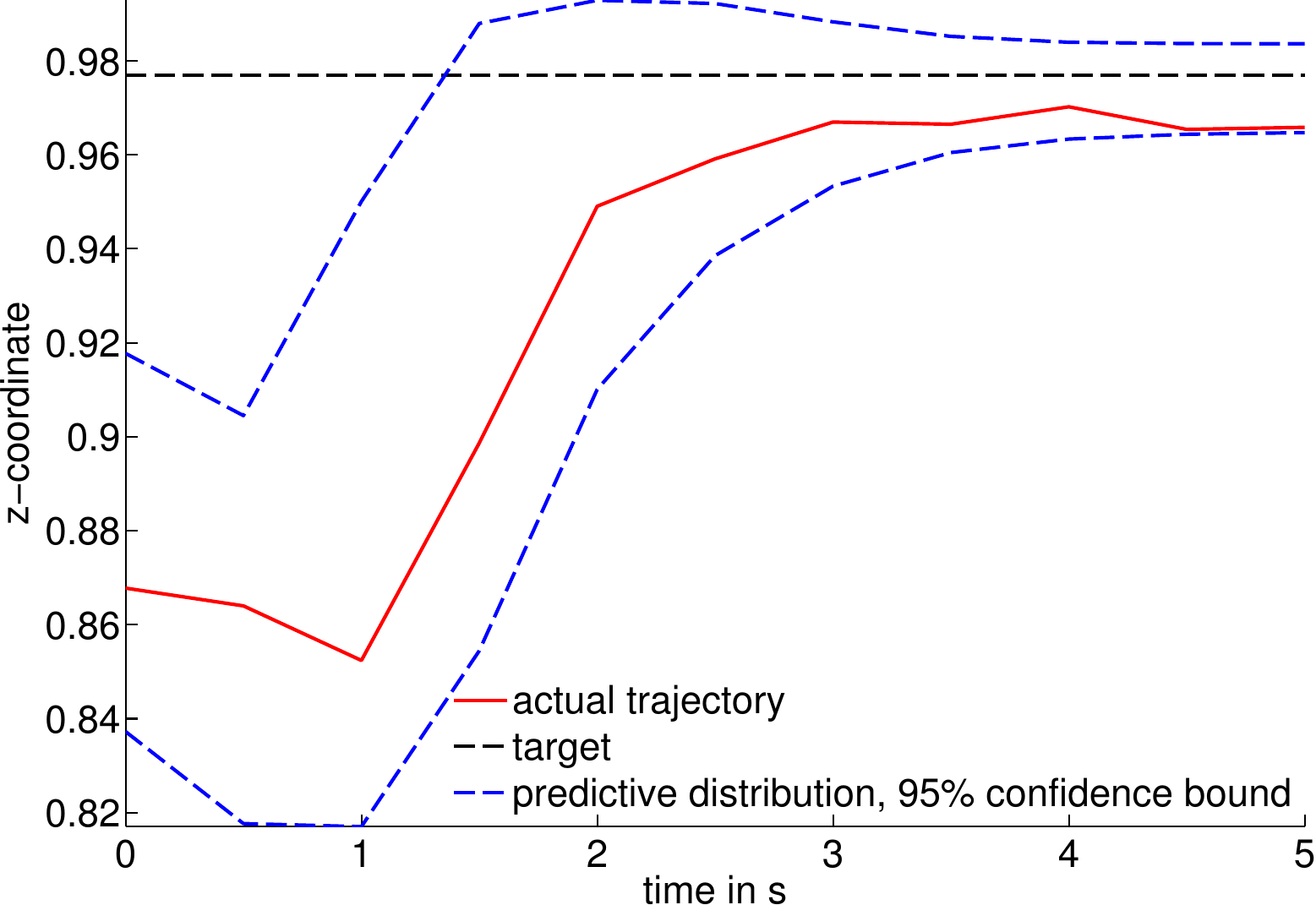}
\label{fig:predictive distributions}
}
\caption{Robot block stacking task: \subref{fig:learning curve}
  Average learning curve with 95\% standard error,
  \subref{fig:predictive distributions} Long-term predictions.}
\figspace
\end{figure*}

A total of ten learning-interacting iterations (including the random
initial trial) generally sufficed to learn both good forward models
and good controllers as shown in \fig~\ref{fig:learning curve}, which
displays the learning curve for a typical training session, averaged
over ten test runs after each learning stage and all blocks
B2--B6. The effects of learning became noticeable after about four
learning iterations. After 10 learning iterations, the block in the
gripper was expected to be very close (approximately at noise level)
to the target.
The required interaction time sums up to only $\unit[50]{s}$ per
controller and $\unit[230]{s}$ in total (the initial random trial is
counted only once). This speed of learning is difficult to achieve by
other RL methods that learn from scratch as shown in
\sec~\ref{sec:cartpole results}.

Fig.~\ref{fig:predictive distributions} gives some insights into the
quality of the learned forward model after 10 controlled trials. It
shows the marginal predictive distributions and the actual
trajectories of the block in the gripper.
The robot learned to pay attention to stabilizing the $y$-coordinate
quickly: Moving the arm up\slash down caused relatively large ``system
noise'' as the arm was quite jerky in this direction: In the
$y$-coordinate the predictive marginal distribution noticeably
increases between $\unit[0]{s}$ and $\unit[2]{s}$.  As soon as the
$y$-coordinate was stabilized, the predictive uncertainty in all three
coordinates collapsed.
Videos of the block-stacking robot
are available at
\mbox{\url{http://www.youtube.com/user/PilcoLearner}}.

\section{Discussion}
\label{sec:discussion}

We have shed some light on essential ingredients for successful and
efficient policy learning: (1) a probabilistic forward model with a
faithful representation of model uncertainty and (2) Bayesian
inference. We focused on very basic representations: GPs for the
probabilistic forward model and Gaussian distributions for the state
and control distributions. More expressive representations and
Bayesian inference methods are conceivable to account for
multi-modality, for instance. However, even with our current set-up,
\textsc{pilco} can already learn learn complex control and robotics
tasks.  In~\cite{Bischoff2013}, our framework was used in an
industrial application for throttle valve control in a combustion
engine.

\textsc{Pilco} is a model-based policy search method, which uses the
GP forward model to predict state sequences given the current
policy. These predictions are based on deterministic approximate
inference, e.g., moment matching.  Unlike all model-free policy search
methods, which are inherently based on sampling
trajectories~\cite{Deisenroth2013}, \textsc{pilco} exploits the
learned GP model to compute analytic
gradients of an approximation to the expected long-term cost $J^\pi$
for policy search.  Finite differences or more efficient
sampling-based approximations of the gradients require many function
evaluations, which limits the effective number of policy
parameters~\cite{Peters2006,Deisenroth2013}. Instead, \textsc{pilco}
computes the gradients analytically and, therefore, can learn
thousands of policy parameters~\cite{Deisenroth2011c}.

It is possible to exploit the learned GP model for sampling
trajectories using the PEGASUS algorithm~\cite{Ng2000}, for
instance. Sampling with GPs can be straightforwardly parallelized, and
was exploited in~\cite{Kupcsik2013} for learning meta controllers.
However, even with high parallelization, policy search methods based
on trajectory sampling do usually not rely on
gradients~\cite{Ng2004a,Bagnell2001,Ko2007,Kupcsik2013} and are
practically limited by a relatively small number of a few tens of
policy parameters they can manage~\cite{Ng2008}.\footnote{``Typically, PEGASUS policy search algorithms have been using [...]
  maybe on the order of ten parameters or tens of parameters; so, 30,
  40 parameters, but not thousands of parameters
  [...]'', A. Ng~\cite{Ng2008}.}

In \sec~\ref{sec:cost exploration}, we discussed \textsc{pilco}'s
natural exploration property as a result of Bayesian averaging. It is,
however, also possible to explicitly encourage additional exploration
in a UCB (upper confidence bounds) sense~\cite{Auer2002}: Instead of
summing up expected immediate costs, see \eq(\ref{eq:expected
  return}), we would add the sum of cost standard deviations, weighted
by a factor $\kappa\in\R$. Then, $J^\pi(\vec \theta) = \sum_t \big(\E[
c(\vec x_t)] + \kappa\sigma[\cost(\vec x_t)]\big)$. This type of
utility function is also often used in experimental
design~\cite{Chaloner1995} and Bayesian
optimization~\cite{Lizotte2008, Brochu2009, Osborne2009, Srinivas2010}
to avoid getting stuck in local minima. Since \textsc{pilco}'s
approximate state distributions $\prob(\vec x_t)$ are Gaussian, the
cost standard deviations $\sigma[\cost(\vec x_t)]$ can often be
computed analytically. For further details, we refer the reader
to~\cite{Deisenroth2010b}.

One of \textsc{pilco}'s key benefits is the reduction of model errors
by explicitly incorporating model uncertainty into planning and
control. \textsc{Pilco}, however, does not take temporal correlation
into account. Instead, model uncertainty is treated as noise, which
can result in an under-estimation of model
uncertainty~\cite{Schneider1997}. On the other hand, the
moment-matching approximation used for approximate inference is
typically a conservative approximation.

In this article, we focused on learning controllers in MDPs with
transition dynamics that suffer from \emph{system noise}, see
\eq(\ref{eq:system equation}). The case of \emph{measurement noise}
is more challenging: Learning the GP models is a real challenge since
we no longer have direct access to the state. However, approaches for
training GPs with noise on both the training inputs and training
targets yield initial promising results~\cite{McHutchon2011}. For a
more general POMDP set-up, Gaussian Process Dynamical Models
(GPDMs)~\cite{Wang2008,Ko2009a} could be used for learning both a
transition mapping and the observation mapping. However, GPDMs
typically need a good initialization~\cite{Turner2010} since the
learning problem is very high dimensional.


In~\cite{Hall2011}, the \textsc{pilco} framework was extended to allow
for learning reference tracking controllers instead of solely
controlling the system to a fixed target location.
In~\cite{Deisenroth2011b}, we used \textsc{pilco} for planning and
control in \emph{constrained environments}, i.e., environments with
obstacles. This learning set-up is important for practical robot
applications.  By discouraging obstacle collisions in the cost
function, \textsc{pilco} was able to find paths around obstacles
without ever colliding with them, not even during training. Initially,
when the model was uncertain, the policy was conservative to stay away
from obstacles.
The \textsc{pilco} framework has been applied in the context of
model-based imitation learning to learn controllers that minimize the
Kullback-Leibler divergence between a distribution of demonstrated
trajectories and the predictive distribution of robot
trajectories~\cite{Englert2013,Englert2013a}.
Recently, \textsc{pilco} has also been extended to a multi-task
set-up~\cite{Deisenroth2013d}.

\section{Conclusion}
\label{sec:conclusion}

We have introduced \textsc{pilco}, a practical model-based policy
search method using analytic gradients for policy
learning. \textsc{Pilco} advances state-of-the-art RL methods for
continuous state and control spaces in terms of learning speed by at
least an order of magnitude.  Key to \textsc{pilco}'s success is a
principled way of reducing the effect of model errors in model
learning, long-term planning, and policy learning. \textsc{Pilco} is
one of the few RL methods that has been directly applied to robotics
without human demonstrations or other kinds of informative
initializations or prior knowledge.

The \textsc{pilco} learning framework has demonstrated that Bayesian
inference and nonparametric models for learning controllers is not
only possible but also practicable. Hence, nonparametric Bayesian
models can play a fundamental role in classical control set-ups, while
avoiding the typically excessive reliance on explicit models.



\section*{Acknowledgments}
The research leading to these results has received funding from the
EC's Seventh Framework Programme (FP7/2007--2013) under grant
agreement \#270327, ONR MURI grant N00014-09-1-1052, and Intel Labs.



\begin{IEEEbiography}[{\includegraphics[width=1in,height=1.25in,clip,keepaspectratio]{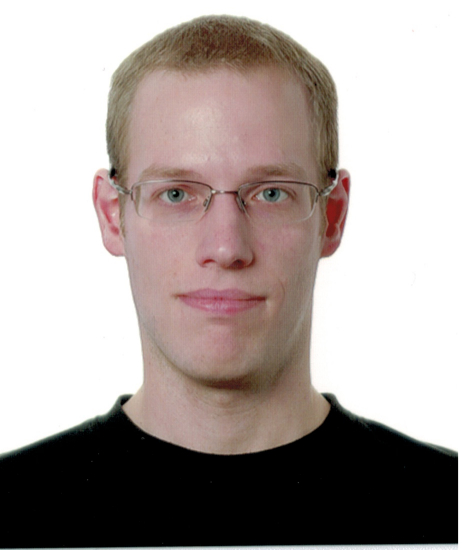}}]{Marc
    Peter Deisenroth} conducted his Ph.D. research at the Max Planck
  Institute for Biological Cybernetics (2006--2007) and at the
  University of Cambridge (2007--2009) and received his Ph.D. degree
  in 2009. He is a Research Fellow at the Department of Computing at
  Imperial College London. He is also adjunct researcher at the
  Computer Science Department at TU Darmstadt, where he has been Group
  Leader and Senior Researcher from December 2011 to August 2013. From
  February 2010 to December 2011, he has been a Research Associate at
  the University of Washington.  His research interests center
  around modern Bayesian machine learning and its application to
  autonomous control and robotic systems.
\end{IEEEbiography}

\begin{IEEEbiography}[{\includegraphics[width=1in,height=1.25in,clip,keepaspectratio]{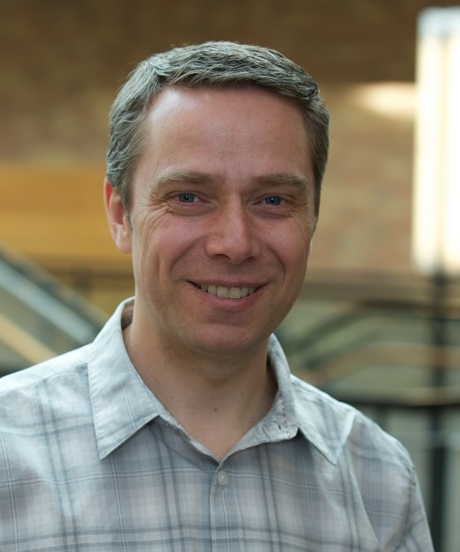}}]{Dieter
    Fox} received the Ph.D. degree from the University of Bonn,
  Germany.  He is Professor in the Department of Computer Science \&
  Engineering at the University of Washington, where he heads the UW
  Robotics and State Estimation Lab. From 2009 to 2011, he was also
  Director of the Intel Research Labs Seattle.  Before going to UW, he
  spent two years as a postdoctoral researcher at the CMU Robot
  Learning Lab. His research is in artificial intelligence, with a
  focus on state estimation applied to robotics and activity
  recognition. He has published over 150 technical papers and is
  coauthor of the text book \emph{Probabilistic Robotics}. Fox is an
  editor of the \emph{IEEE Transactions on Robotics}, was program
  co-chair of the 2008 AAAI Conference on Artificial Intelligence, and
  served as the program chair of the 2013 Robotics Science and Systems
  conference.  He is a fellow of AAAI and a senior member of IEEE.
\end{IEEEbiography}

\begin{IEEEbiography}[{\includegraphics[width=1in,height=1.25in,clip,keepaspectratio]{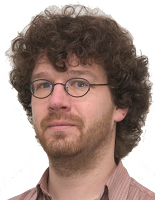}}]{Carl
    Edward Rasmussen} is Reader in Information Engineering at the
  Department of Engineering at the University of Cambridge. He was a
  Junior Research Group Leader at the Max Planck Institute for
  Biological Cybernetics in T\"ubingen, and a Senior Research Fellow
  at the Gatsby Computational Neuroscience Unit at UCL. He has wide
  interests in probabilistic methods in machine learning, including
  nonparametric Bayesian inference, and has co-authored the text book
  \emph{Gaussian Processes for Machine Learning}, the MIT Press 2006.
\end{IEEEbiography}


\appendices

\section{Trigonometric Integration}
\label{sec:integration}
This section gives exact integral equations for trigonometric
functions, which are required to implement the discussed
algorithms. The following expressions can be found in the
book by~\cite{Gradshteyn2000}, where $x\sim\mathcal N(x|\mu,\sigma^2)$ is
Gaussian distributed with mean $\mu$ and variance $\sigma^2$.
%
\begin{align*}
  \E_x[\sin(x)] &=\int\sin(x)\prob(x)\d x  = \exp(-\tfrac{\sigma^2}{2})\sin(\mu)\,,\\
  \E_x[\cos(x)] &=\int\cos(x)\prob(x)\d x  = \exp(-\tfrac{\sigma^2}{2})\cos(\mu)\,,\\
  \E_x[\sin(x)^2]&=\int\sin(x)^2\prob(x)\d x  \nonumber\\
&= \tfrac{1}{2}\big(1-\exp(-2\sigma^2)\cos(2\mu)\big)\,,\\
  \E_x[\cos(x)^2]&=\int\cos(x)^2\prob(x)\d x \\
&= \tfrac{1}{2}\big(1+\exp(-2\sigma^2)\cos(2\mu)\big)\,, \\
  \E_x[\sin(x)\cos(x)]&=\int \sin(x)\cos(x)\prob(x)\d x \\
&=
  \int\tfrac{1}{2}\sin(2x)\prob(x)\d x\\
  & = \tfrac{1}{2}\exp(-2\sigma^2)\sin(2\mu)\,.
\end{align*}

\section{Gradients}
\label{sec:derivatives}
In the beginning of this section, we will give a few derivative
identities that will become handy. After that we will detail
derivative computations in the context of the moment-matching
approximation.

\subsection{Identities}
Let us start with a set of basic derivative
identities~\cite{Petersen2008} that will prove useful in the
following:
\begin{align*}
&\frac{\partial |\mat K(\vec\theta)|}{\partial\vec\theta}=|\mat
K|\tr\left(\mat K\inv\frac{\partial\mat K}{\partial\vec\theta}\right)\,,\\
&\frac{\partial |\mat K|}{\partial\mat K}=|\mat K|(\mat K\inv)\T \,,\\
&\frac{\partial \mat K\inv(\vec\theta)}{\partial\vec\theta}=-\mat
K\inv\frac{\partial\mat K(\vec\theta)}{\partial\vec\theta}\mat K\inv \,,\\
&\frac{\partial\vec\theta\T\mat K\vec\theta}{\partial\vec\theta}=\vec\theta\T(\mat
K+\mat K\T) \,,\\
&\frac{\partial\tr(\mat A\mat K\mat B)}{\partial\mat K}=\mat A\T\mat
B\T \,,\\
&\frac{\partial |\mat A\mat K + \mat I|\inv}{\partial\mat K}= -  |\mat
A\mat K + \mat I|\inv \big((\mat A \mat K + \mat I)\inv\big)\T \,,\\
&\frac{\partial}{\partial B_{ij}} (\vec a - \vec b)\T (\mat A + \mat
B)\inv (\vec a - \vec b) \nonumber\\
&\qquad= -(\vec a - \vec b)\T\big[(\mat A + \mat
B)\inv_{(:,i)} (\mat A + \mat B)\inv_{(j,:)}\big](\vec a -
\vec b)\,.
\end{align*}
In in the last identity $\mat B(:,i)$ denotes the $i$th column of $\mat B$
and $\mat B(i,:)$ is the $i$th row of $\mat B$.

\subsection{Partial Derivatives of the Predictive Distribution with Respect to the Input Distribution}
For an input distribution $\tilde{\vec
  x}_{t-1}\sim\gaussx{\tilde{\vec
    x}_{t-1}}{\tilde{\vec\mu}_{t-1}}{\tilde{\mat\Sigma}_{t-1}}$, where
$\tilde{\vec x} = [\vec x\T \vec u\T]\T$ is the control-augmented
state, we detail the derivatives of the predictive mean
$\vec\mu_{\vec\Delta}$, the predictive covariance
$\mat\Sigma_{\vec\Delta}$, and the cross-covariance $\cov[\tilde{\vec
  x}_{t-1},\vec\Delta]$ (in the moment matching approximation) with
respect to the mean $\tilde{\vec\mu}_{t-1}$ and covariance
$\tilde{\mat\Sigma}_{t-1}$ of the input distribution.

\subsubsection{Derivatives of the Predictive Mean with Respect to the Input Distribution}
In the following, we compute the derivative of the predictive GP mean
$\vec\mu_{\vec\Delta}\in\R^E$ with respect to the mean and the covariance of
the input distribution $\gaussx{\vec
  x_{t-1}}{\vec\mu_{t-1}}{\mat\Sigma_{t-1}}$.  The function value of
the predictive mean is given as
\begin{align}
  \vec\mu_{\vec\Delta}^a &=  \sum_{i=1}^n \beta_{a_i}q_{a_i}\,,\\
  q_{a_i} &= \sigma_{f_a}^2 |\mat I
  +\mat\Lambda_a\inv\tilde{\mat\Sigma}_{t-1}|^{-\tfrac{1}{2}}\label{eq:appendix predictive mean function value}\\
&\quad\times \exp\big(-\tfrac{1}{2}
  (\tilde{\vec x}_i -\tilde{\vec\mu}_{t-1})\T(\mat\Lambda_a +
  \tilde{\mat\Sigma}_{t-1})\inv (\tilde{\vec x}_i
  -\tilde{\vec\mu}_{t-1})\big)\,.
\nonumber
\end{align}
%

\paragraph{Derivative with respect to the  Input Mean}
Let us start with the derivative of the predictive mean with respect
to the mean of the input distribution.  From the function value in
\eq~\eqref{eq:appendix predictive mean function value}, we obtain the
derivative
\begin{align}
  \frac{\partial\vec\mu_{\vec\Delta}^a}{\partial\tilde{\vec\mu}_{t-1}} & =
  \sum_{i=1}^n \beta_{a_i}\frac{\partial
    q_{a_i}}{\partial\tilde{\vec\mu}_{t-1}} \\
  &= \sum_{i=1}^n\beta_{a_i} q_{a_i}(\tilde{\vec x}_i -
  \tilde{\vec\mu}_{t-1})\T(\tilde{\mat\Sigma}_{t-1} +
  \mat\Lambda_a)\inv 
\label{eq:dq_a/dmu}
\end{align}
$\in\R^{1\times (D+F)} $ for the $a$th target dimension, where we used
\begin{align}
  \frac{\partial q_{a_i}}{\partial\tilde{\vec\mu}_{t-1}}
  =q_{a_i}(\tilde{\vec x}_i -
  \tilde{\vec\mu}_{t-1})\T(\tilde{\mat\Sigma}_{t-1} +
  \mat\Lambda_a)\inv\,.
\end{align}

\paragraph{Derivative with Respect to the Input Covariance Matrix}
For the derivative of the predictive mean with respect to the input
covariance matrix $\mat\Sigma_{t-1}$, we obtain
\begin{align}
  \frac{\partial\vec\mu_{\vec\Delta}^a}{\partial\tilde{\mat\Sigma}_{t-1}} &=
  \sum_{i=1}^n\beta_{a_i}\frac{\partial
    q_{a_i}}{\partial\tilde{\mat\Sigma}_{t-1}} \,.
\end{align}
By defining
\begin{align*}
&\eta(\tilde{\vec x}_i,
\tilde{\vec\mu}_{t-1},\tilde{\mat\Sigma}_{t-1})\nonumber\\
&\quad=\exp\big(-\tfrac{1}{2}
    (\tilde{\vec x}_i -\tilde{\vec\mu}_{t-1})\T(\mat\Lambda_a +
    \tilde{\mat\Sigma}_{t-1})\inv (\tilde{\vec x}_i
    -\tilde{\vec\mu}_{t-1})\big)
\end{align*}
we obtain
\begin{align}
  \frac{\partial q_{a_i}}{\partial\tilde{\mat\Sigma}_{t-1}} &=
  \sigma_{f_a}^2\Bigg(\frac{\partial|\mat I +
      \mat\Lambda_a\inv\tilde{\mat\Sigma}_{t-1}|^{-\tfrac{1}{2}}}{\partial\tilde{\mat\Sigma}_{t-1}}\eta(\tilde{\vec x}_i,
\tilde{\vec\mu}_{t-1},\tilde{\mat\Sigma}_{t-1})\Bigg.\nonumber\\
  &\Bigg.\quad + |\mat I
    +\mat\Lambda_a\inv\tilde{\mat\Sigma}_{t-1}|^{-\tfrac{1}{2}}\frac{\partial}{\partial\tilde{\mat\Sigma}_{t-1}}\eta(\tilde{\vec x}_i,
\tilde{\vec\mu}_{t-1},\tilde{\mat\Sigma}_{t-1}) \Bigg)\label{eq:dq_a/dSigma}
\nonumber
\end{align}
for $i=1,\dotsc,n$.  Here, we compute the two partial derivatives
\begin{align}
 &\frac{\partial|\mat I +
    \mat\Lambda_a\inv\tilde{\mat\Sigma}_{t-1}|^{-\tfrac{1}{2}}}
  {\partial\tilde{\mat\Sigma}_{t-1}} \\
&\quad\hspace{-2mm} =-\frac{1}{2}|\mat I +
  \mat\Lambda_a\inv\tilde{\mat\Sigma}_{t-1}|^{-\tfrac{3}{2}}\frac{\partial|\mat
    I + \mat\Lambda_a\inv\tilde{\mat\Sigma}_{t-1}|}
  {\partial\tilde{\mat\Sigma}_{t-1}}\\
  &\quad\hspace{-2mm} =-\frac{1}{2}|\mat I +
  \mat\Lambda_a\inv\tilde{\mat\Sigma}_{t-1}|^{-\tfrac{3}{2}}|\mat I +
  \mat\Lambda_a\inv\tilde{\mat\Sigma}_{t-1}|\nonumber\\
&\qquad\hspace{-2mm} \times\big((\mat I + \mat\Lambda_a\inv\tilde{\mat\Sigma}_{t-1})\inv\mat\Lambda_a\inv\big)\T\\
&\quad\hspace{-2mm} =-\frac{1}{2}|\mat I +
  \mat\Lambda_a\inv\tilde{\mat\Sigma}_{t-1}|^{-\tfrac{1}{2}}\big((\mat
  I +
  \mat\Lambda_a\inv\tilde{\mat\Sigma}_{t-1})\inv\mat\Lambda_a\inv\big)\T
\end{align}
and for $p,q=1,\dotsc, D+F$
\begin{align}
  &\frac{\partial}{\partial\tilde{\mat\Sigma}_{t-1}^\idx{pq}}
  (\mat\Lambda_a +
  \tilde{\mat\Sigma}_{t-1})\inv\\
&\quad =-\tfrac{1}{2}\Big((\mat\Lambda_a
    + \tilde{\mat\Sigma}_{t-1})\inv_{\idx{:,p}}(\mat\Lambda_a +
    \tilde{\mat\Sigma}_{t-1})\inv_{\idx{q,:}}\Big.\nonumber\\
&\quad\quad\Big. +(\mat\Lambda_a
    + \tilde{\mat\Sigma}_{t-1})\inv_{\idx{:,q}}(\mat\Lambda_a +
    \tilde{\mat\Sigma}_{t-1})\inv_{\idx{p,:}}\Big) \in\R^{(D+F)\times (D+F)}\,,
\nonumber
\end{align}
where we need to explicitly account for the symmetry of $\mat\Lambda_a
+ \tilde{\mat\Sigma}_{t-1}$. Then, we obtain
\begin{align}
  &\frac{\partial\vec\mu_{\vec\Delta}^a}{\partial\tilde{\mat\Sigma}_{t-1}} =
  \sum_{i=1}^n
  \beta_{a_i}q_{a_i}\Bigg(-\tfrac{1}{2}\big((\mat\Lambda_a\inv\tilde{\mat\Sigma}_{t-1}
    + \mat I)\inv\mat\Lambda_a\inv\big)\T \Bigg.\nonumber\\
  &\quad\Bigg.- \tfrac{1}{2}\underbrace{\underbrace{(\tilde{\vec x}_i
        -\tilde{\vec\mu}_{t-1})\T}_{1\times
        (D+F)}\!\!\!\!\!\!\!\!\!\underbrace{\frac{\partial(\mat\Lambda_a +
          \tilde{\mat\Sigma}_{t-1})\inv}{\partial\tilde{\mat\Sigma}_{t-1}}}_{(D+F)\times(D+F)\times(D+F)\times(D+F)}\!\!\!\!\!\!\!\!\!\underbrace{(\tilde{\vec 
          x}_i -\tilde{\vec\mu}_{t-1})}_{(D+F)\times
        1}}_{(D+F)\times(D+F)} \Bigg)\,,
\end{align}
where we used a tensor contraction in the last expression inside the
bracket when multiplying the difference vectors onto the matrix
derivative.

\subsubsection{Derivatives of the Predictive Covariance with Respect to the Input Distribution}

For target dimensions $a,b=1,\dotsc, E$, the entries of the predictive
covariance matrix $\mat\Sigma_{\vec\Delta}\in\R^{E\times E}$ are given as
\begin{align}
  \sigma_{\vec\Delta_{ab}}^2 &= \vec\beta_a\T\big(\mat Q - \vec q_a\vec
  q_b\T)\vec\beta_b \nonumber\\
&\quad + \delta_{ab} \big(\sigma_{f_a}^2 - \tr((\mat K_a +
  \sigma_{w_a}^2\mat I)\inv\mat Q)\big)
\end{align}
where $\delta_{ab}=1$ if $a=b$ and 0 otherwise.

The entries of $\mat Q\in\R^{n\times n}$ are given by
\begin{align}
  Q_{ij} &=\sigma_{f_a}^2\sigma_{f_b}^2|(\mat\Lambda_a\inv +
  \mat\Lambda_b\inv)\tilde{\mat\Sigma}_{t-1} +\mat
  I|^{-\frac{1}{2}}\nonumber\\
  &\quad\times\exp\big(-\tfrac{1}{2}(\tilde{\vec x}_i-\tilde{\vec
  x}_j)\T(\mat\Lambda_a+\mat\Lambda_b)\inv(\tilde{\vec x}_i-\tilde{\vec
  x}_j)\big)\nonumber\\
  &\quad \times\exp\Big(-\tfrac{1}{2}(\hat{\vec z}_{ij} - \tilde{\vec
    \mu}_{t-1})\T\nonumber\\
&\quad \times \big((\mat\Lambda_a\inv + \mat\Lambda_b\inv)\inv +
  \tilde{\mat\Sigma}_{t-1}\big)\inv(\hat{\vec
    z}_{ij}-\tilde{\vec\mu}_{t-1})\Big)\,,
\label{eq:Q-matrix entries for derivatives}\\
  \hat{\vec z}_{ij}&\coloneqq
  \mat\Lambda_b(\mat\Lambda_a+\mat\Lambda_b)\inv\tilde{\vec x}_i +
  \mat\Lambda_a(\mat\Lambda_a + \mat\Lambda_b)\inv\tilde{\vec x}_j\,,
\end{align}
where $i,j=1,\dotsc,n$.

\paragraph{Derivative with Respect to the Input Mean}
For the derivative of the entries of the predictive covariance matrix
with respect to the predictive mean, we obtain
\begin{align}
  \frac{\partial\sigma_{\vec\Delta_{ab}}^2}{\partial\tilde{\vec\mu}_{t-1}}
  &= \vec\beta_a\T\left(\frac{\partial\mat
      Q}{\partial\tilde{\vec\mu}_{t-1}} - \frac{\partial\vec
      q_a}{\partial\tilde{\vec\mu}_{t-1}}\vec q_b\T - \vec q_a
    \frac{\partial\vec
      q_b\T}{\partial\tilde{\vec\mu}_{t-1}}\right)\vec\beta_b\nonumber\\
&\quad +
  \delta_{ab}\left(-(\mat K_a + \sigma_{w_a}^2\mat
    I)\inv\frac{\partial\mat
      Q}{\partial\tilde{\vec\mu}_{t-1}}\right)\,,
\end{align}
where the derivative of $Q_{ij}$ with respect to the input mean is
given as
\begin{align}
  \frac{\partial
    Q_{ij}}{\partial\tilde{\vec\mu}_{t-1}}&=Q_{ij}(\hat{\vec
    z}_{ij}-\tilde{\vec\mu}_{t-1})\T((\mat\Lambda_a\inv
  + \mat\Lambda_b\inv)\inv + \tilde{\mat\Sigma}_{t-1})\inv\,.
\end{align}

\paragraph{Derivative with Respect to the Input Covariance Matrix}
The derivative of the entries of the predictive covariance matrix with
respect to the \emph{covariance matrix of the input distribution} is
\begin{align}
  \frac{\partial\sigma_{\vec\Delta_{ab}}^2}{\partial\tilde{\mat\Sigma}_{t-1}}
  &= \vec\beta_a\T\left(\frac{\partial\mat
      Q}{\partial\tilde{\mat\Sigma}_{t-1}} - \frac{\partial\vec
      q_a}{\partial\tilde{\mat\Sigma}_{t-1}}\vec q_b\T - \vec q_a
    \frac{\partial\vec
      q_b\T}{\partial\tilde{\mat\Sigma}_{t-1}}\right)\vec\beta_b\nonumber\\
&\quad +
  \delta_{ab}\left(-(\mat K_a + \sigma_{w_a}^2\mat
    I)\inv\frac{\partial\mat
      Q}{\partial\tilde{\mat\Sigma}_{t-1}}\right)\,.
\label{eq:dsigmadSigma}
\end{align}
Since the partial derivatives $\partial\vec
q_a/\partial\tilde{\mat\Sigma}_{t-1}$ and $\partial\vec
q_b/\partial\tilde{\mat\Sigma}_{t-1}$ are known from
Eq.~\eqref{eq:dq_a/dSigma}, it remains to compute $\partial \mat
Q/\partial\tilde{\mat\Sigma}_{t-1}$. The entries $Q_{ij}$,
$i,j=1,\dotsc,n$ are given in Eq.~\eqref{eq:Q-matrix entries for
  derivatives}. By defining
\begin{align*}
  c&\coloneqq
  \sigma_{f_a}^2\sigma_{f_b}^2\exp\left(-\tfrac{1}{2}(\tilde{\vec
      x}_i-\tilde{\vec
      x}_j)\T(\mat\Lambda_a\inv+\mat\Lambda_b\inv)\inv(\tilde{\vec
      x}_i-\tilde{\vec x}_j))\right)\\
  e_2&\coloneqq\exp\Big(-\tfrac{1}{2}(\hat{\vec z}_{ij} -
  \tilde{\vec\mu}_{t-1})\T
  \big((\mat\Lambda_a\inv+\mat\Lambda_b\inv)\inv +
  \tilde{\mat\Sigma}_{t-1}\big)\inv\nonumber\\
  &\qquad\qquad\times(\hat{\vec z}_{ij} - \tilde{\vec\mu}_{t-1})\Big)
\end{align*}
we obtain the desired derivative
\begin{align}
  \frac{\partial Q_{ij}}{\partial\tilde{\mat\Sigma}_{t-1}}&= c\Bigg[
  -\tfrac{1}{2}|(\mat\Lambda_a\inv+\mat\Lambda_b)\inv
  \tilde{\mat\Sigma}_{t-1}+\mat
  I|^{-\tfrac{3}{2}}\nonumber\\
  &\quad\times\frac{\partial|(\mat\Lambda_a\inv +
    \mat\Lambda_b)\inv\tilde{\mat\Sigma}_{t-1}+\mat
    I|}{\partial\tilde{\mat\Sigma}_{t-1}} e_2
  \nonumber\\
  &\quad + |(\mat\Lambda_a\inv +
  \mat\Lambda_b\inv)\tilde{\mat\Sigma}_{t-1} + \mat I|^{-\tfrac{1}{2}}
  \frac{\partial e_2}{\partial\tilde{\mat\Sigma}_{t-1}} \Bigg]\,.
\end{align}
Using the partial derivative
\begin{align}
  &\frac{\partial|(\mat\Lambda_a\inv +
    \mat\Lambda_b)\inv\tilde{\mat\Sigma}_{t-1}+\mat
    I|}{\partial\tilde{\mat\Sigma}_{t-1}}\nonumber\\
  &= |(\mat\Lambda_a\inv+\mat\Lambda_b)\inv
  \tilde{\mat\Sigma}_{t-1}+\mat I|\nonumber\\
  &\quad\times\left(\big( (\mat\Lambda_a\inv +
    \mat\Lambda_b\inv)\tilde{\mat\Sigma}_{t-1} + \mat I \big)\inv
    (\mat\Lambda_a\inv + \mat\Lambda_b\inv)\right)\T\\
  &= |(\mat\Lambda_a\inv+\mat\Lambda_b)\inv
  \tilde{\mat\Sigma}_{t-1}+\mat
  I|\\
  &\quad\times\tr\left(\big((\mat\Lambda_a\inv+\mat\Lambda_b\inv)\tilde{\mat\Sigma}_{t-1}+
    \mat I\big)\inv(\mat\Lambda_a\inv +
    \mat\Lambda_b\inv)\frac{\partial
      \tilde{\mat\Sigma}_{t-1}}{\partial\tilde{\mat\Sigma}_{t-1}}\right)
\nonumber
\end{align}
the partial derivative of $Q_{ij}$ with respect to the covariance matrix
$\tilde{\mat\Sigma}_{t-1}$ is given as
\begin{align}
  &\frac{\partial Q_{ij}}{\partial\tilde{\mat\Sigma}_{t-1}}\nonumber\\
  &= c\Bigg[ -\tfrac{1}{2}|(\mat\Lambda_a\inv+\mat\Lambda_b)\inv
  \tilde{\mat\Sigma}_{t-1}+\mat
  I|^{-\tfrac{3}{2}}\nonumber\\
  &\quad\times|(\mat\Lambda_a\inv+\mat\Lambda_b)\inv
  \tilde{\mat\Sigma}_{t-1}+\mat I|e_2
  \nonumber\\
  &\quad\times
  \tr\left(\big((\mat\Lambda_a\inv+\mat\Lambda_b\inv)\tilde{\mat\Sigma}_{t-1}+
    \mat I\big)\inv(\mat\Lambda_a\inv +
    \mat\Lambda_b\inv)\frac{\partial\tilde{\mat\Sigma}_{t-1}}{\partial\tilde{\mat\Sigma}_{t-1}}\right)
  \nonumber\\
  &\quad+|(\mat\Lambda_a\inv+\mat\Lambda_b)\inv
  \tilde{\mat\Sigma}_{t-1}+\mat I|^{-\tfrac{1}{2}}\frac{\partial
    e_2}{\partial\tilde{\mat\Sigma}_{t-1}}
  \Bigg]\\
  &= c|(\mat\Lambda_a\inv+\mat\Lambda_b)\inv
  \tilde{\mat\Sigma}_{t-1}+\mat I|^{-\tfrac{1}{2}}\\
  &\quad\times \Bigg[ -\tfrac{1}{2}\left(\big( (\mat\Lambda_a\inv +
    \mat\Lambda_b\inv)\tilde{\mat\Sigma}_{t-1} + \mat I \big)\inv
    (\mat\Lambda_a\inv + \mat\Lambda_b\inv)\right)\T\! e_2\nonumber\\
  &\quad + \frac{\partial e_2}{\partial\tilde{\mat\Sigma}_{t-1}}
  \Bigg]\,,
\end{align}
where the partial derivative of $e_2$ with respect to the entries
$\Sigma_{t-1}^{\idx{p,q}}$ is given as
\begin{align}
  \frac{\partial e_2}{\partial\tilde{\Sigma}_{t-1}^{(p,q)}}&=
  -\frac{1}{2}(\hat{\vec z}_{ij}-\tilde{\vec\mu}_{t-1})\T
  \frac{\partial \big((\mat\Lambda_a\inv + \mat\Lambda_b\inv)\inv +
    \tilde{\mat\Sigma}_{t-1}\big)\inv}{\partial\tilde{\Sigma}_{t-1}^\idx{p,q}}\nonumber\\
&\quad\times(\hat{\vec 
    z}_{ij}-\tilde{\vec\mu}_{t-1})e_2\,.
\label{eq:de2dSigma}
\end{align}
The missing partial derivative in \eq~\eqref{eq:de2dSigma} is given by
\begin{align}
  &\frac{\partial \big((\mat\Lambda_a\inv + \mat\Lambda_b\inv)\inv +
    \tilde{\mat\Sigma}_{t-1}\big)\inv}{\partial\tilde{\Sigma}_{t-1}^\idx{p,q}}
  = -\mat\Xi_{(pq)}\,,
\end{align}
where we define 
\begin{align}
  \mat\Xi_{(pq)} &= \tfrac{1}{2}(\mat \Phi_{(pq)} +
  \mat\Phi_{(qp)})\in\R^{(D+F)\times (D+F)}\,,
\end{align}
$p,q=1,\dotsc,D+F$ with
\begin{align}
  \mat\Phi_{(pq)}&=\Bigg(\big((\mat\Lambda_a\inv +
  \mat\Lambda_b\inv)\inv +
  \tilde{\mat\Sigma}_{t-1}\big)_{(:,p)}\inv\nonumber\\
  &\qquad\times\big((\mat\Lambda_a\inv + \mat\Lambda_b\inv)\inv +
  \tilde{\mat\Sigma}_{t-1}\big)_{(q,:)}\inv\Bigg)\,.
\end{align}
This finally yields
\begin{align}
  \frac{\partial Q_{ij}}{\partial\tilde{\mat\Sigma}_{t-1}} &=
 ce_2|(\mat\Lambda_a\inv+\mat\Lambda_b)\inv
  \tilde{\mat\Sigma}_{t-1}+\mat I|^{-\tfrac{1}{2}} \nonumber\\
  &\quad\times\Bigg[ \left(\big( (\mat\Lambda_a\inv +
      \mat\Lambda_b\inv)\tilde{\mat\Sigma}_{t-1} + \mat I \big)\inv
      (\mat\Lambda_a\inv + \mat\Lambda_b\inv)\right)\T\nonumber\\
&\qquad-(\hat{\vec
      z}_{ij}-\tilde{\vec\mu}_{t-1})\T \mat\Xi (\hat{\vec
      z}_{ij}-\tilde{\vec\mu}_{t-1})
  \Bigg]\\
  &=-\tfrac{1}{2}Q_{ij}\nonumber\\
&\quad\times \Bigg[ \left(\big( (\mat\Lambda_a\inv +
      \mat\Lambda_b\inv)\tilde{\mat\Sigma}_{t-1} + \mat I \big)\inv
      (\mat\Lambda_a\inv + \mat\Lambda_b\inv)\right)\T\nonumber\\
&\qquad - (\hat{\vec
      z}_{ij}-\tilde{\vec\mu}_{t-1})\T \mat\Xi (\hat{\vec
      z}_{ij}-\tilde{\vec\mu}_{t-1}) \Bigg]\,,
\end{align}
 which concludes the
computations for the partial derivative in
\eq~\eqref{eq:dsigmadSigma}.

\subsubsection{Derivative of the Cross-Covariance with Respect to the Input Distribution}
For the cross-covariance 
\begin{align*}
  \cov_{f,\tilde{\vec x}_{t-1}}[\tilde{\vec
  x}_{t-1},\Delta_t^a]&=\tilde{\mat\Sigma}_{t-1}\mat R\inv \sum_{i =
    1}^n \beta_{a_i} q_{a_i}(\tilde{\vec x}_{i}
  -\tilde{\vec\mu}_{t-1})\,,\\
\mat R &\coloneqq \tilde{\mat\Sigma}_{t-1} + \mat\Lambda_a\,,
\end{align*}
we obtain
\begin{align}
  &\frac{\partial\cov_{f,\tilde{\vec x}_{t-1}}[\vec\Delta_t, \tilde{\vec
    x}_{t-1}]}{\partial\tilde{\vec\mu}_{t-1}}\nonumber\\
&\quad=
  \tilde{\mat\Sigma}_{t-1}\mat R\inv \sum_{i=1}^n \beta_i
  \left((\tilde{\vec x}_{i} -\tilde{\vec\mu}_{t-1})\frac{\partial
      q_i}{\partial \tilde{\vec\mu}_{t-1}} + q_i \mat I\right)
\end{align}
$\in\R^{(D+F)\times (D+F)}$ for all target dimensions $a=1,\dotsc, E$. 

The corresponding derivative with respect to the covariance matrix
$\tilde{\mat\Sigma}_{t-1}$ is given as
\begin{align}
  &\frac{\partial\cov_{f,\tilde{\vec x}_{t-1}}[\vec\Delta_t, \tilde{\vec
    x}_{t-1}]}{\partial\tilde{\mat\Sigma}_{t-1}} \nonumber\\
  &\quad=
  \left(\frac{\partial\tilde{\mat\Sigma}_{t-1}}{\partial\tilde{\mat\Sigma}_{t-1}}\mat
    R\inv + \tilde{\mat\Sigma}_{t-1}\frac{\partial\mat R\inv
    }{\partial\tilde{\mat\Sigma}_{t-1}}\right)\sum_{i = 1}^n
  \beta_{a_i} q_{a_i}(\tilde{\vec x}_{i} -\tilde{\vec\mu}_{t-1})\nonumber\\
  &\qquad + \tilde{\mat\Sigma}_{t-1}\mat R\inv \sum_{i = 1}^n
  \beta_{a_i} (\tilde{\vec x}_{i}
  -\tilde{\vec\mu}_{t-1})\frac{\partial
    q_{a_i}}{\partial\tilde{\mat\Sigma}_{t-1}}\,.
\end{align}
%

\ifCLASSOPTIONcaptionsoff
  \newpage
\fi



%

\end{document}